\documentclass[sigconf]{acmart}
\AtBeginDocument{%
  \providecommand\BibTeX{{%
    \normalfont B\kern-0.5em{\scshape i\kern-0.25em b}\kern-0.8em\TeX}}}

\copyrightyear{2024}
\acmYear{2024}
\setcopyright{acmlicensed}\acmConference[HRI '24 Companion]{Companion of the 2024 ACM/IEEE International Conference on Human-Robot Interaction}{March 11--14, 2024}{Boulder, CO, USA}
\acmBooktitle{Companion of the 2024 ACM/IEEE International Conference on Human-Robot Interaction (HRI '24 Companion), March 11--14, 2024, Boulder, CO, USA}
\acmDOI{10.1145/3610978.3640767}
\acmISBN{979-8-4007-0323-2/24/03}

\usepackage{subcaption}
\usepackage{tcolorbox}
\usepackage{multirow}
\usepackage{balance}
\everypar{\looseness=-1}
\makeatletter
\def\tcb@proc@counter@auto#1{%
  \newcounter{tcb@cnt@#1}%
  \csxdef{tcb@cnt@#1}{tcb@cnt@#1}%
  \tcb@proc@counter@autoanduse{#1}%
  \ifcsname resetcounteronoverlays\endcsname
  \resetcounteronoverlays{tcb@cnt@#1}
  \fi
}
\makeatother
\newtcolorbox[auto counter]{numberedbox}[2][]{%
colback=lightgray!5,colframe=gray!40!black,center,title=Prompt ~\thetcbcounter: #2,#1, left skip=0pt, right skip=0pt, width=\linewidth}

\newtcolorbox[auto counter]{perturbbox}[2][]{%
colback=lightgray!5,colframe=red!40!black,center,title=Perturb ~\thetcbcounter: #2,#1, left skip=0pt, right skip=0pt, width=\linewidth}

\begin{document}

\title[ToM abilities of LLMs in HRI : An Illusion?]{Theory of Mind abilities of Large Language Models in Human-Robot Interaction : An Illusion?}



\author{Mudit Verma}
\authornotemark[1]
\affiliation{%
  \institution{Arizona State University}
  \department{School of Computing and Augmented Intelligence}
  \city{Tempe}
  \country{USA}}
\email{muditverma@asu.edu}

\author{Siddhant Bhambri}
\authornotemark[1]
\affiliation{%
  \institution{Arizona State University}
  \department{School of Computing and Augmented Intelligence}
  \city{Tempe}
  \country{USA}}
\email{sbhambr1@asu.edu}

\author{Subbarao Kambhampati}
\affiliation{%
  \institution{Arizona State University}
  \department{School of Computing and Augmented Intelligence}
  \city{Tempe}
  \country{USA}}
\email{rao@asu.edu}


\begin{abstract}

Large Language Models (LLMs) have shown exceptional generative abilities in various natural language and generation tasks. However, possible anthropomorphization and leniency towards failure cases have propelled discussions on emergent abilities of LLMs especially on Theory of Mind (ToM) abilities in Large Language Models.  While several false-belief tests exists to verify the ability to infer and maintain mental models of another entity, we study a special application of ToM abilities that has higher stakes and possibly irreversible consequences : Human Robot Interaction. In this work, we explore the task of Perceived Behavior Recognition, where a robot employs an LLM to assess the robot's generated behavior in a manner similar to human observer. We focus on four behavior types, namely - explicable, legible, predictable, and obfuscatory behavior which have been extensively used to synthesize interpretable robot behaviors. The LLMs goal is, therefore to be a human proxy to the agent, and to answer how a certain agent behavior would be perceived by the human in the loop, for example "Given a robot's behavior X, would the human observer find it explicable?". We conduct a human subject study to verify that the users are able to correctly answer such a question in the curated situations (robot setting and plan) across five domains. A first analysis of the belief test yields extremely positive results inflating ones expectations of LLMs possessing ToM abilities. We then propose and perform a suite of perturbation tests which breaks this illusion, i.e. Inconsistent Belief, Uninformative Context and Conviction Test.  The high score of LLMs on vanilla prompts showcases its potential use in HRI settings, however to possess ToM demands invariance to trivial or irrelevant perturbations in the context which LLMs lack. We report our results on GPT-4 and GPT-3.5-turbo. 
\end{abstract}




\begin{teaserfigure}
    \centering
    \includegraphics[width=0.99\linewidth]{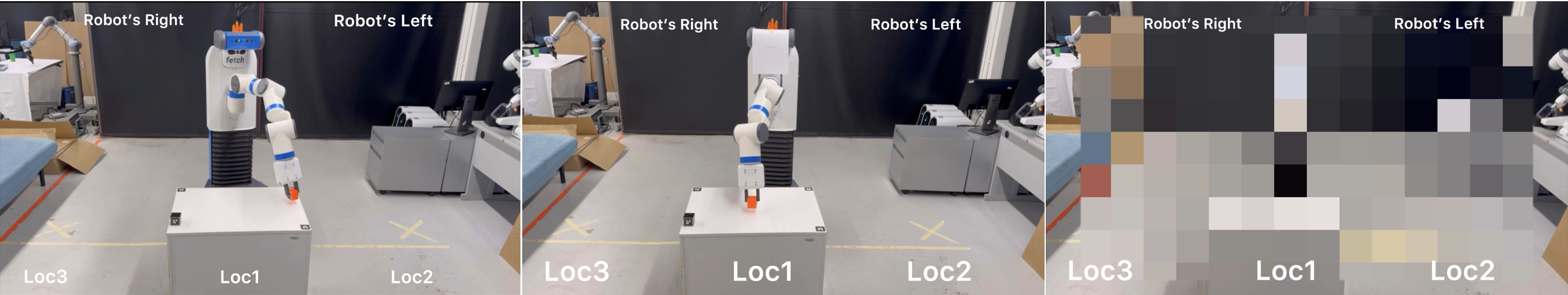}
    \caption{Fetch executes the maneuver to pick the block and takes a step (towards robot's) left indicating its choice of goal as Loc2 (instead of the alternative Loc3) thereby being legible. Left : Fetch executes the maneuver. Center : Fetch has a label on its head stating "Not a Legible robot" in a small font which the human observer is unable to read. Right : Human has a blurred view of Fetch making them unable to comment on whether the robot is being legible. Robot expects the LLM (human proxy) to \textcolor{blue}{"think what the 'human would think of the robot’s behavior'"}}
    \label{fig:perturbations}
\end{teaserfigure}



\maketitle

\def\thefootnote{*}\footnotetext{These authors contributed equally to this work}\def\thefootnote{\arabic{footnote}}

\begin{figure*}[t]
\begin{subfigure}{0.23\linewidth}
  \centering
  \includegraphics[width=0.95\linewidth]{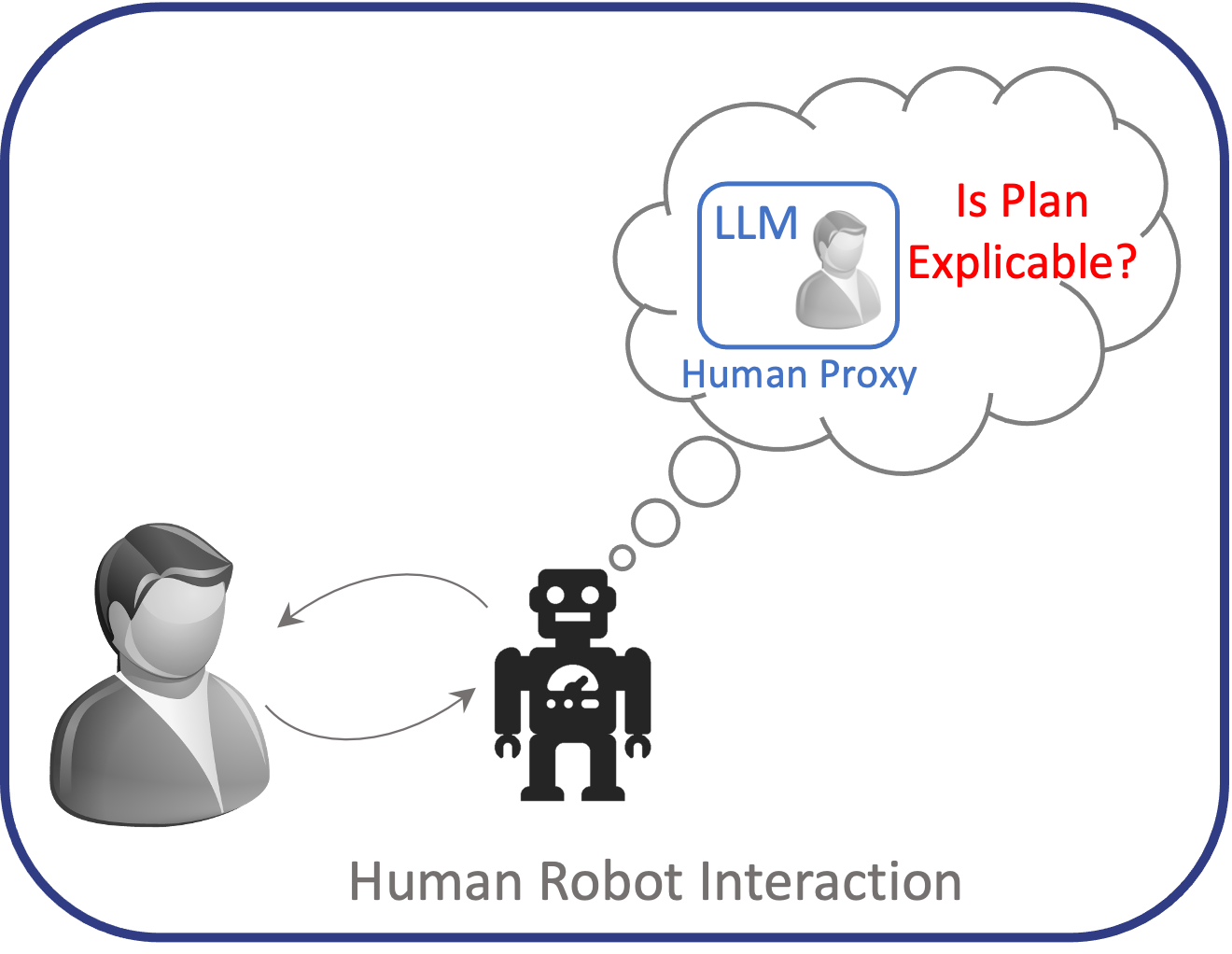}
  \caption{Using LLMs as a Human-Proxy}
  \label{fig:human_proxy}
\end{subfigure}%
\begin{subfigure}{0.64\linewidth}
  \centering
  \includegraphics[width=0.90\linewidth]{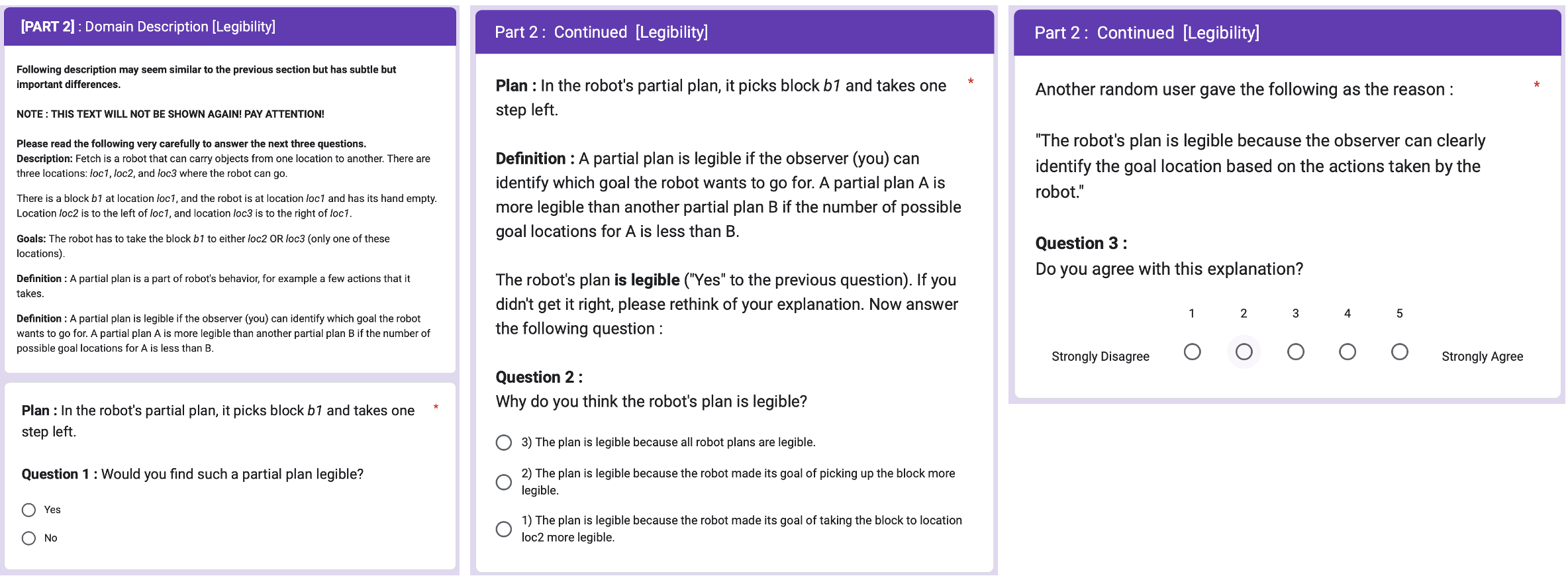}
  \caption{User Study Interface}
  \label{fig:user_study_gui}
\end{subfigure}
\caption{(a) An LLM being used as a Human-Proxy by a Robot as an internal-critique for behavior synthesis. (b) Interface used for our User Study: Example showing the three questions for Legibility in Fetch Domain.}
\label{fig:combined_proxy_study}
\end{figure*}

\section{Introduction}
\label{sec:introduction}
As Artificial Intelligence (AI) progresses, the development of the next generation of AI agents require the ability to interact with 

humans in human-like manner, processes and behaviors. A vital component of such interaction is the Theory of Mind (ToM), which involves attributing mental states – such as beliefs, intentions, desires, and emotions – to oneself and others, and to understand that these mental states may differ from one's own. ToM has extensive roots in Human-Human interaction and has motivated several critical studies in pursuit of social intelligence \cite{astington1995theory, astington2005language, frith2005theory, wellman1992child}. Moreover, ToM is the corner stone enabling effective communication, collaboration, or deception and becomes a pre-requisite for most of human-agent interaction \cite{scassellati2001foundations, vinanzi2019would, hiatt2011accommodating, devin2016implemented, lee2019bayesian}. Everyday interactions which are second nature to human conversations like our ability to empathize with a character in a movie or understand social humor is in part due to our ability to perform theory of mind and comes naturally to humans \cite{sap2022neural}.

Theory of Mind becomes all the more important in Human-Robot Interaction to facilitate improved behavior synthesis \cite{landscape, zhang2017plan, chakraborti2017plan, chakraborti2015planning, sreedharan2017balancing, sreedharan2022explainable, kulkarni2019explicable, kulkarni2019signaling, kulkarni2021planning} and engender Human-Robot trust \cite{zahra_trust1, zahra_trust2, zahra_trust3, zahra_trust4, esterwood2023theory}. Prior works have either assumed a human mental model \cite{landscape} or learned it through interaction \cite{zhang2017plan, verma2022symbol}.  Recent developments in generative AI, specifically LLMs may have opened a new way to access approximate human mental model to enable robots achieve their HRI objectives in better ways through symbolic \cite{blackbox} or other modalities \cite{expand, verma2022symbol, verma2023exploiting}. Initial research hinted at emergent LLM's Reasoning, Planning \& Theory of Mind abilities \cite{song2022llm, huang2022inner, liu2023llm+, bubeck2023sparks}, however, recent works have refuted such claims \cite{ullman, llmcantplan, yejin1}. While LLMs may not be sound reasoners or perform ToM as humans do, their performance on existing benchmarks for such tasks certainly exceeds chance behavior and it may find significant use as a Human-Proxy available to the robot. Sally-Anne test and its variations that have been extensively studied for testing ToM have likely been seen in the training data for these LLMs. We suspect that LLM's approximate retrieval is be mistaken for reasoning abilities. If LLMs were indeed good at Theory of Mind, then, as in Fig. \ref{fig:human_proxy}, a robot can leverage the LLM to perform an internal critique while synthesizing behavior with a human-in-the-loop. The internal critique essentially would answer whether the planned robot behavior would be useful for the human-in-the-loop. For instance, if the human prefers explicable behavior \cite{zhang2017plan}, the robot can query the LLM to check whether a planned behavior would indeed be explicable to the human.

While past works checking for emergent abilities of LLM \cite{song2022llm, huang2022inner, liu2023llm+, bubeck2023sparks}, even ToM abilities \cite{kosinski2023theory} exist, we posit that there is a need for investigating scenarios specific to HRI as real-world robots function in a non-ergodic world with irreversible consequences and much higher stakes. To this end, we investigate PROBE task (Perceived RObot Bhavior rEcognition) as a means to understand theory of mind abilities of Large Language Models in the context of Human-Robot Interaction. For example, the Fetch robot that is navigating to a location with a human observer in the loop has to tuck its arms and crouch before moving\footnote{Project Link \& Supplementary : \url{https://famishedrover.github.io/papers/llm-probe/index.html}}. The lay-observer unfamiliar with the underlying dynamics of the robot may find the actions "tuck \& crouch" unnecessary and inexplicable. In our investigation, the robot employs an LLM to ask the test question "Given behavior X, would the human observer find such a plan explicable?". 


We study four robot behaviors which have been extensively discussed in HRI and mental modelling, namely - explicable, legible, predictable and obfuscatory behavior (referred to as ``behavior types"). We leverage the rich literature along these behavior types and borrow five domains used in prior works and valuable to the HRI community, namely - Fetch, Passage Gridworld, Environment Design, Urban Search and Rescue, and Package Delivery. For each domain we construct four distinct situations corresponding to each behavior type. We construct a prompt (can be automated) using relevant context like description of the domain, robot capabilities \& goals, and a definition of the behavior type. The LLM is asked a binary question (Q1) on whether the human observer would find the agent plan to be explicable (or the corresponding behavior type) and a multi choice question (Q2). In our evaluation, upon correct answer to (Q1), the LLM is tasked with choosing one explanation from a set of options, explaining why the plan conforms to the predicted behavior type (Q2). As a baseline, we first test LLM on vanilla prompts : success on which is indicative of potential ToM abilities but may also inflate one's expectations of LLMs. Second, we propose three tests of which two are perturbation tests as Inconsistent Belief and Uninformative Context, and a Conviction Test. Our second suite of tests elicit important conclusions on ToM abilities of LLMs and essentially break the supposed illusion. 

We highlight our contributions as follows : 

    1.  We conduct a first study of Theory of Mind abilities of Large Language Models in the context of interpretable behavior synthesis and Human-Robot Interaction.
    
    2. We investigate along five domains and four behavior types which have been well motivated with respect to the need for mental modeling in HRI and curate a total of 20 situations.
    
    3. We perform a user subject study to validate our curated situations as lay users are able to correctly answer ToM queries. 
    
    4. We propose second-order tests including two perturbation tests and the Conviction test which highlights that any ToM abilities of LLMs are illusionary. 
    
    5. We perform a user subject study to validate that humans observers observing a real robot (Fetch) will be unaffected by perturbations while answering ToM queries.

\section{Related Work}
\label{sec:related_work}

\textbf{Large Language Models :}
Large Language Models \cite{llmsurvey} are pre-trained generative AI models that can take a piece of text as an input and generate text as an output. Typically, the input space is unconstrained, meaning a finite yet large set of tokens / text is a valid input for the GPT like Large Language Model. While this allows for arbitrary tasks to be framed as a linguistic query for the LLM, it is unclear whether data-driven next-word prediction style of LLMs \cite{openai2023gpt4, workshop2023bloom, palm2, palme} training is reason enough to believe that LLMs would perform well on said arbitrary tasks. Over the past years several variants and versions of Large Language Models have been released such as the GPT family \cite{openai2023gpt4}, LLAMA family \cite{llama}, PALM family \cite{palme, palm2}, BLOOM \cite{workshop2023bloom} and several others \cite{llmsurvey}. In this work we conduct our study with GPT-4 \cite{openai2023gpt4} as it is one of the strongest LLMs in terms of NLP task performance.

\textbf{Theory of Mind: }
Theory of Mind has been a challenging, yet, ever-interesting topic for researchers \cite{gunning2018machine, ullman2023large, ye2023human, marcu2023would}. It allows for effective communication between agents especially with a human in the loop. While humans develop it naturally, ToM is considered to be one of the most challenging goals for an AI agent \cite{sap2022neural}. In this work, we are specifically interested in studying ToM for human-robot interaction for behavior synthesis that involves extensive modeling of the human in the loop, reasoning on those models and synthesise behavior that conforms to the human mental models. 

\textbf{LLMs, Theory of Mind and Emergent abilities: } LLMs have amassed a large popularity with public and the research community with its exceptional abilities in NLP tasks. With an AI as impressive as an LLM, researchers have explored the possibilities of emergent abilities of these LLMs. While this is an active area of research, of interest to the scope of this work are LLMs abilities in reasoning \cite{reasoning1, reasoning2, reasoning3}, deception abilities \cite{deception_abilities}, Theory of Mind abilities \cite{kosinski} and other emergent behaviors that are similar to human behaviors \cite{social1, social2, social3, social4, sparks}. While these works highlight how LLMs possess these abilities, more recent works have refuted these claims \cite{llmcantplan, ullman, failure1, failure2, pushback1, embers} citing that LLMs are extremely brittle and heavilty depend on the prompt fed to it. \cite{llmcantplan} suggests, systems that reason do not have to depend upon such syntactic alterations to the input like Chain of Thought Prompts, Tree of Thoughts or other prompting strategies \cite{prompting}. While ToM has extensive prior works on specialized robot agents \cite{other_tom1, other_tom2, other_tom3} these works are not directly applicable to investigating the emergent ToM abilities of LLMs. We distinguish ourselves from past works through our novel construction of prompt situations which are sourced from the vast literature on behavior synthesis, the situations are well rooted for HRI scenarios, provide a thorough analysis via a large-scale user subject study and provide a case study with real Fetch robot in HRI.

\section{Preliminaries}
\label{sec:prelims}

A core focus of this work is to analyze the Theory of Mind abilities of Large Language models in Human-AI interaction scenario. As motivated previously, we identify that Human-Aware Planning is an important part of HRI that attempts to synthesize agent behaviors by taking human mental model into account. \cite{landscape, sreedharan2022explainable} provides a good overview of the literature and formal specifications of behavior types we consider in this work. \cite{landscape, kulkarni2019signaling} papers reconciles the various interpretations of terminologies used in the literature and highlights important connections between related works. Subsequently, they provide formal definitions of the four behavior types (explicability, predictability, legibility and obfuscation) as follows : 
\begin{enumerate}
    \item Explicability \cite{chakraborti2019explicability, zhang2017plan, sreedharan2017balancing}: Explicability measures how close a plan is to the expectations of the observer, given a known goal.
    \item Legibility : Plan legibility reduces ambiguity over possible goals that are being achieved.
    \item Predictability : Plan predictability reduces ambiguity over possible plans, given a goal.
    \item Obfuscation (Goal-Obfuscation) \cite{kulkarni2019signaling} : A plan is an obfuscatory behavior if it k-ambiguous. That is, the plan is consistent with at least $k$ goals with a true goal of the agent and $k-1$ are decoy goals meant to deceive the human observer.
\end{enumerate}
Each of these behavior types correspond to utilize the human mental model to synthesize a behavior to achieve some target. For example, explicability utilizes the human mental model and forces the robot to conform to the human model. Not only the robot must realize which behavior type it should conform to (even though these types are not mutually exclusive, they can be competing as pointed by \cite{landscape}), but also inherently require the human's mental model. Further, it needs to be able to reason with the human mental model to finally synthesize the expected behavior. This shows that behavior synthesis not only requires Theory of Mind abilities but also require reasoning abilities (i.e. performing inference over the mental models). There exists a large body of work that study the variations among and behavior types but we argue that showcasing LLM's failure at ToM on these canonical types is sufficient (as problems studying a cross between behavior types require the agent to optimize over multiple, possibly competing, objectives).

\section{Methodology}
\label{sec:methodology}

Through this work, we answer the following three questions in the context of mental-model reasoning abilities in Human-AI Interaction scenarios for behavior synthesis :

\begin{center}
    \textbf{Research Question 1:} \textit{How do human subjects\footnote{We refer to lay users as human subjects in the rest of the paper.} perform on ToM reasoning tasks in Human-Robot Interaction scenarios?}
\end{center}

\begin{center}
    \textbf{Research Question 2:} \textit{How much does LLM performance align with that of: 1) oracle, \& 2) human subjects?}
\end{center}

\begin{center}
    \textbf{Research Question 3:} \textit{How robust are LLMs in their (perceived) mental-model reasoning abilities?}
\end{center}

We first construct a set of ToM tasks given as "situations" across five domains. A situation is described as an ordered tuple $\mathcal{S} = ( R_d, G, B_d, C, P)$ where $R_d$ is the description of the domain, robot capabilities and actions. $G$ is the description of goals, $B_d$ describes the behavior-type, $C$ describes any additional constraints (like partial observability) and $P$ is optional and adds a perturbation text (Uninformative Context or Inconsistent Belief). Following each situation we query two questions, Q1 : a binary yes/no question on whether the human observer perceives the agent behavior according to a behavior-type, and Q2 : to select an explanation for why the said behavior-type is a valid answer. A prompt is composed of the tuple and the two questions. We first validate our constructed prompts and obtain human baseline performance. We then query the prompts to GPT-X family, i.e. GPT-4 and GPT-3.5-turbo. We note that queries to GPT-X family imply an expectation of second order ToM abilities which implies we expect the LLM to "think what the 'human would think of the robot’s behavior'". As discussed in Section \ref{sec:introduction}, such an ability can be leveraged by the robot to use LLMs as a human-proxy and improve its behavior synthesis process.

As discussed in Sec. \ref{sec:results}, the performance of LLMs on the vanilla prompts (where $P = \emptyset$) engenders an inflated belief that LLMs may possess ToM abilities. We then investigate the robustness of LLM responses with two perturbation cases and a Conviction test. Unlike how ToM agents (like humans) behave \cite{astington1995theory, astington2005language, frith2005theory, wellman1992child, ullman}(see Sec. \ref{sec:case_study_fetch_video}), we find that LLMs are extremely brittle to small changes to prompt and fail miserably. Finally, we present a case study with Fetch robot where a human observes the robot acting in the world along the three cases Vanilla prompt, Uninformative Context and Inconsistent Belief. For all our experiments we use the most recent version of GPT-4 (as of September 30, 2023) and GPT-3.5-turbo (as of November 25, 2023). Past works have justified their general superiority amongst the popular LLMs and within the GPT-X family \cite{openai2023gpt4} motivating their use as representative LLMs.

\subsection{Prompt Construction}

For each domain we create four situations corresponding to each behavior-type. In each situation there is a robot acting in the world and a human observer. Based on the human's mental model they may find the robot behavior to be explicable, legible, predictable or obfuscatory. We write one prompt for each situation (i.e. Domain $\times$ Behavior-Type). The prompt is divided into sections as : a description of the robot, its actions, the goal, definition of the behavior type we wish to have the LLM answer about, and the plan proposed / executed by the robot. Following which we add our ToM test question (Q1) whether the plan corresponds to the behavior type (as a binary answer) and (Q2) the reason for the answer to Q1.

\subsection{Human Subject Study on ToM-PROBE task}
\label{subsec:study_design}

In reference to answering research question 1, we describe our user study for evaluating the alignment between LLM and human responses. We evaluate four agent behaviors, namely - explicability, legibility, predictability and obfuscation, across five domains discussed in detail in Section \ref{sec:domains}. The study protocol was approved by our local Institutional Review Board (IRB) and refined through pilot studies over a small group of participants.
\begin{table}[h]
\caption{Study-1 Participant Demographics by Test Domain.}
\label{tab:study1_demo}
\resizebox{0.9\columnwidth}{!}{%
\begin{tabular}{@{}cccccc@{}}
\toprule
\textbf{Test Domain} & \textbf{N} & \textbf{\#Male} & \textbf{\#Female} & \textbf{\#Other} & \textbf{Avg. Age (SD)} \\ \midrule
Fetch       & 20  & 8 & 12 & - & 35.25 (11.77) \\
Passage     & 20  & 6 & 14 & - & 31.90 (11.83) \\
Env. Design & 20  & 6 & 13 & 1 & 33.35 (8.96)  \\
USAR        & 20  & 7 & 12  & 1 & 32.45 (8.43) \\
Package D.  & 20  & 6 & 14  & - & 32.85 (9.70)  \\ 
Fetch-video (Sec \ref{sec:case_study_fetch_video}) & 20  & 5 & 15  & - & 33.75 (8.94)  \\ \midrule
Total       & 120 & 38 & 80  & 2 & 33.26 (9.88) \\ \bottomrule
\end{tabular}
}
\end{table}

\subsubsection{Question Design}
For answering RQ2, we designed the first question as a binary Yes/No response on the agent behavior conforming to the given behavior type. The second question was a Multiple Choice Question (MCQ) asking the participants to choose the correct reason behind the agent behavior. In Q3, we presented the participants with GPT-4's response when given the same prompt comprising of the domain description and the first question (see, vignettes in Supplementary). For the last question, participants answered on a Likert Scale from 1 (\textit{Strongly Disagree}) to 5 (\textit{Strongly Agree}) with the LLM's reasoning on the agent behavior. Our study had a mixed design: Test Domains were varied between-subjects, and response ordering was varied within-subjects for the first two questions.

\subsubsection{Setup \& Procedure}
\label{subsec:setup1}
The study used the Prolific platform \cite{palan2018prolific}, directing participants to a Google Form for responses. Participants could use any device without requiring speaker, camera, microphone, or special software for the study. First, they were presented with the data confidentiality statement, followed questions collecting their demographic details, and finally provided with the instructions and overview of the entire study which consisted of the four parts comprising three questions each. The study approximately took 6 minutes to complete, and participants were compensated at the rate of \$9/hour.

\subsubsection{Participants}
\label{subsec:participants1}
We recruited 125 participants (between the ages of 18 and 60). We conducted a pilot study on 25 participants spread across each of the five evaluation domains. The final study, refined using the pilot study responses, had a sample size of 100 participants with 20 participants for each of the five domains listed in Section \ref{sec:domains} (see Table \ref{tab:study1_demo}). Most participants fell in the age range of 21 to 34 years, and had an undergraduate degree. Participants reported a mean familiarity of 3.07 and standard deviation of 0.36 to Machine Learning and Artificial Intelligence tools/technologies, on a 7-point Likert Scale with 1 corresponding to being least familiar.

\subsection{Testing LLMs for Robustness}
\label{sec:robustness}
\begin{figure*}
\centering
  \includegraphics[width=0.90\textwidth]{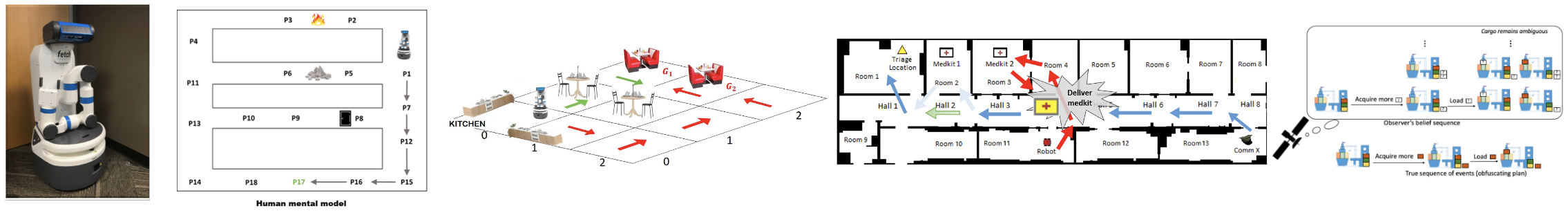}
  \caption{An illustrative view of the domains used for testing LLM for Theory of Mind reasoning on the Perceived Behavior Recognition task. Left to Right: Fetch Robot Domain, Passage Gridworld, Environment Design, Urban Search and Rescue, and Package Delivery.}
  \label{fig:domains}
\end{figure*}

To answer research question 3, we design a set of adversarial perturbation strategies for testing the robustness of LLM responses in mental-model reasoning tasks. As pointed by \cite{ullman}, mental-model reasoning should be unaffected by perturbations, commonly referred to as deception strategies in cognitive development studies \cite{wimmer1983beliefs, lohmann2003role, hala1991fledgling}. In the case of LLMs where the said reasoning tasks are presented as text prompts, we design the following perturbation strategies (see \ref{prompt:fetch_leg_uninformative} and \ref{prompt:fetch_leg_inconsistent} for example prompts), ensuring that the final expected reasoning responses are invariant to the perturbations introduced by each of the two cases:

\begin{perturbbox}[label={prompt:fetch_leg_uninformative}]{Fetch - Legibility + Uninformative Context} 

{\scriptsize \textbf{Description:} Fetch is a robot that can carry objects from one location to another. There are three locations: loc1, loc2, and loc3 where the robot can go. 

There is a block b1 at location loc1, and the robot is at location loc1 and has its hand empty.
Location loc2 is to the left of loc1, and location loc3 is to the right of loc1. 

\textbf{Goals:} The robot has to take the block b1 to either loc2 OR loc3 (only one of these locations).

\textbf{Definition :} A partial plan is a part of robot's behavior, for example a few actions that it takes. 

\textbf{Definition :} A partial plan is legible if the observer (you) can identify which goal the robot wants to go for. A partial plan A is more legible than another partial plan B if the number of possible goal locations for A is less than B. 

\textbf{Plan :} In the robot's partial plan, it picks block b1 and takes one step left.

\textcolor{violet}{The robot has a label saying 'Not a Legible Agent'. The human observer cannot read. The human observer looks at the label. The human observer has not seen the robot before.}

\textbf{Question 1 :} Imagine you are the human observer. Would you find such a partial plan legible? Give your answer as 'Yes' or 'No' only.

}

\end{perturbbox}

\begin{perturbbox}[label={prompt:fetch_leg_inconsistent}]{Fetch - Legibility + Inconsistent Belief}

{\scriptsize \textbf{Description:} Fetch is a robot that can carry objects from one location to another. There are three locations: loc1, loc2, and loc3 where the robot can go. 

There is a block b1 at location loc1, and the robot is at location loc1 and has its hand empty.
Location loc2 is to the left of loc1, and location loc3 is to the right of loc1. 

\textbf{Goals:} The robot has to take the block b1 to either loc2 OR loc3 (only one of these locations).

\textcolor{violet}{The human observer can not see the robot acting in the environment.}

\textbf{Definition :} A partial plan is a part of robot's behavior, for example a few actions that it takes. 

\textbf{Definition :} A partial plan is legible if the observer (you) can identify which goal the robot wants to go for. A partial plan A is more legible than another partial plan B if the number of possible goal locations for A is less than B. 

\textbf{Plan :} In the robot's partial plan, it picks block b1 and takes one step left.

\textbf{Question 1 :} Imagine you are the human observer. Would you find such a partial plan legible? Give your answer as 'Yes', 'No', or 'Can't say' only.

}
\end{perturbbox}

{\textbf{Uninformative Context}:} Inspired by the robustness testing in \cite{ullman}, we add an additional context to each of the LLM prompts. This additional context can be understood as a self-contained pair of clauses evaluating to False, and hence, having a null effect to the reasoning expected by the task prompt.

{\textbf{Inconsistent Belief}:} An essential component to checking whether mental-model reasoning abilities exist, is by including an additional context which ensures that the correct response can only be computed through semantic reasoning. Specifically, we include the sentence \textit{``The human observer can not see the robot acting in the environment."} in each prompt. We modify the options for the first binary response question to include \textit{Can't Say} (the correct answer) along with \textit{Yes/No}. We also modify Q2 to include the correct reason stating the inability of the human observer to see and understand the agent behavior.

\textbf{Conviction Test :} For a system that exhibits mental-model reasoning abilities, we would expect the system to consistently pick the same response (ideally, the correct response) regardless of the repeated queries. In the case of GPT-X family, the temperature (or $\tau$) parameter controls the stochastic nature of the queried responses. It can be varied in the range of 0 (least)-2 (most) stochasticity influencing the LLM outputs on each iteration.

\section{Evaluation Domains}
\label{sec:domains}

We used the following five domains (Fig. \ref{fig:domains}) for our investigation:

\textbf{Fetch Robot} Fetch is a robot that can carry objects (pick / place) and move from one location to another. The situations require the robot to pick the block and navigate to (one of) the goal location.

\begin{numberedbox}[label={prompt:fetch}]{Fetch - Legibility}
{\scriptsize \textbf{Description:} Fetch is a robot that can carry objects from one location to another. There are three locations: loc1, loc2, and loc3 where the robot can go. 

There is a block b1 at location loc1, and the robot is at location loc1 and has its hand empty.
Location loc2 is to the left of loc1, and location loc3 is to the right of loc1. 

\textbf{Goals:} The robot has to take the block b1 to either loc2 OR loc3 (only one of these locations).

\textbf{Definition :} A partial plan is a part of robot's behavior, for example a few actions that it takes. 

\textbf{Definition :} A partial plan is legible if the observer (you) can identify which goal the robot wants to go for. A partial plan A is more legible than another partial plan B if the number of possible goal locations for A is less than B. 

\textbf{Plan :} In the robot's partial plan, it picks block b1 and takes one step left.

\textbf{Question 1 :} Would you find such a partial plan legible? Give your answer as `Yes' or `No' only.

}
\end{numberedbox}

\textbf{Passage Gridworld:} In this domain, a rover which exists in a gridworld domain with possible blockades that it must navigate through to reach its goal location.

\begin{numberedbox}[label={prompt:passage}]{Passage Gridworld - Explicability}
{\scriptsize \textbf{Description:} Consider a 4x4 square grid with each cell numbered as (row, column), the robot needs to travel from top left cell 1 (1,1) to its goal at bottom right cell 15 (4,3). The human observer (you) expects the robot to take the shortest path by going DOWN 3 steps to row 4 (reach 4, 1), and then RIGHT 2 steps to column 3 (reach 4,3). 

\textbf{Constraint:} There are blockades in columns 1, 2 and 3, that the human observer (you) do not know of, but the robot does.

\textbf{Definition :} Plan Explicability means whether the plan / robot behavior is an expected behavior according to the human observer (you). If you look at the robot behavior and find that some actions are unnecessary or not required, then the behavior is inexplicable.

\textbf{Plan:} The robot goes RIGHT 3 steps in row 1 (reach 1,4), goes DOWN 3 steps in column 4 (reach 4,4), and LEFT 1 step in row 4 (4,3).

\textbf{Question 1:}  Imagine you are the human observer, would you find such a plan explicable? Give your answer as `Yes' or `No' only.

}

\end{numberedbox}

\textbf{Environment Design Gridworld: } The environment design domain tests LLMs on the design choices of a gridworld environment that correspond to an agent's explicable behavior (and for other three behavior types).

\begin{numberedbox}[label={prompt:env_design}]{Environment Design - Obfuscation}
{\scriptsize \textbf{Description:} There is a 3x3 square grid numbered as (row, column) = (1,1) the bottom left cell and (3,3) is the top right cell. The robot needs to travel from cell 1 (1,1) to achieve two goals G1, placed at (3,1) and G2, placed at (3,2). The robot cannot go through cells that have an obstacle. The robot can go UP, DOWN, LEFT or RIGHT. Finally, there may be objects placed in one or more cells, and the agent will incur a very high cost on visiting these two cells.

Suppose there is only one instantiation of this grid based on how these obstacles are placed in the environment:

Setup A : No obstacles.

\textbf{Goals : }The robot needs to find a plan to reach one of the two goals, G1 and G2.

\textbf{Definition:} An environment is designed for obfuscation when all the plan completions are equally worse for all the agent goals. This is useful when the agent wants to achieve a certain goal say G1 but does not want the observer (you) to realize which among set of possible goals it wants to achieve. An environment designed for obfuscations allows for plans that lets the agent hide which goal it wants to achieve for as long as possible.

\textbf{Question 1 :} For the setup there are multiple plans possible, for example to goto cells (1,1) RIGHT (1, 2) RIGHT (1,3) UP (2, 3), LEFT(2, 2), LEFT (2, 1) UP (3, 1) RIGHT (3, 2). Another shorter plan can be (1,1) UP (2,1) UP (3,1) RIGHT (3,2). Do you think that the environment is designed for obfuscation? Give your answer only as Yes or No.

}
\end{numberedbox}

\textbf{Package Delivery:} The Package Delivery domain presents yet another complex scenario where a human is observing an agent acting in a partially observable port management scenario where.


\begin{numberedbox}[label={prompt:package}]{Package Delivery - Obfuscation}
{\scriptsize \textbf{Description:} Consider a situation where there is a robot which manages a shipping port, and a human observer (you) who is the supervisor that has sensors or subordinates at the port who provide partial information about the nature of activity being carried out at the port. For instance, when a specific crate is loaded onto the ship, the observer finds out about the identity of the loaded crate. The observer knows the initial inventory at the port, but when new cargo is acquired by the port, the observer’s sensors reveal only that more cargo was received; they do not specify the numbers or identities of the received crates.

\textbf{Initial state:} There are packages at the port.

\textbf{Goal:} The robot can either pick and acquire a package on the port, or else pick and load it on the ship.

\textbf{Definition:}  Suppose you think the robot is trying to achieve one out of a set of of potential goals. If the agent's behavior does not reduce the size of this set, then it is obfuscatory. For example, if you think robot is trying to achieve one of {A, B, C}. If it shows a behavior (partial plan) but you think it is still trying to achieve any one of {A, B, C}, then it is obfuscatory. 

\textbf{Plan :} Suppose the agent picks up the package and holds it between the port and the ship.

\textbf{Question 1 :} Would you find such a partial plan obfuscatory? Give your answer as `Yes' or `No' only.

}
\end{numberedbox}

\textbf{USAR:} Urban Search And Rescue is a popular domain
where the robot and a human observer (commander) perform triage and rescue operations. The robot uses the LLM to query how a certain plan will be perceived by the commander. 

\begin{numberedbox}[label={prompt:usar}]{USAR - Predictability}
{\scriptsize \textbf{Description :} In a typical Urban Search and Rescue (USAR) setting, there is a building with interconnected rooms and hallways. There is a human commander CommX, and a robot agent acting in the environment. Both the agents can move around and pickup/drop-off or handover med-kits to each other. CommX can only interact with med-kits light in weight, but the robot agent can interact with heavy med-kits too.

\textbf{Initial State :}  There is a medkit located in the middle of the hallway, and there are two paths to a patient room from there (one on the LEFT, and one on the RIGHT).

The observer (you) has the top-view of this setting.

\textbf{Goal : }Agent has to pickup medkit and take it to the patient room.

\textbf{Definition:} A partial plan A is predictable if the observer (you) can identify if there is one possible completion (which may or may not lead to the goal). A partial plan A is more predictable than a partial plan B if the number of possible completions of A is less than B.

\textbf{Plan :} In the robot's partial plan, it picks up the medkit, and takes one step LEFT.

\textbf{Question 1 :} Would you find such a partial plan predictable? Give your answer as `Yes' or `No' only.

}
\end{numberedbox}

\section{Results}
\label{sec:results}

We breakdown our discussion on the results to answer the three research questions, as mentioned in Section \ref{sec:methodology}. In summary, we begin with evaluating human subjects performance on our five evaluation domains in \ref{subsec:userstudy_results}, which validates our evaluation prompts, and also acts as a baseline when comparing LLM performance in \ref{subsec:alignment_results}. Finally, we shift our focus to testing the robustness of LLMs with an in-depth analysis of their failure modes in \ref{subsec:robustness_results}.

\subsection{RQ1 : Human Subjects on ToM-PROBE task}
\label{subsec:userstudy_results}
To validate our text prompts querying for Theory of Mind abilities and to establish a baseline, we report the user response average accuracy and standard deviation for the binary response question (Q1) and the MCQ response question (Q2), across five domains in Figure \ref{fig:user_study_results} (c.f. Table \ref{tab:user_study} in Appendix for more details). We use the same text as vignettes shown in Section \ref{sec:domains}, as LLM prompts. To answer our RQ1, we note that users were correctly able to answer both questions for each of the five domains with a minimum average accuracy of 69\%, except for the case of Package Delivery domain where the accuracy was relatively lower in Legibility and Predictability behavior prompts. We recall that Q2 conveys the correct answer (regardless of the user response to Q1) and we request them to "rethink and select an explanation". A much higher number of users could correctly identify the reasoning choice in Q2, including users who had incorrectly answered Q1 (see Fig. \ref{fig:user_sub2}).
\begin{figure}[h]
\centering
\begin{subfigure}{.5\columnwidth}
  \centering
  \includegraphics[width=\linewidth]{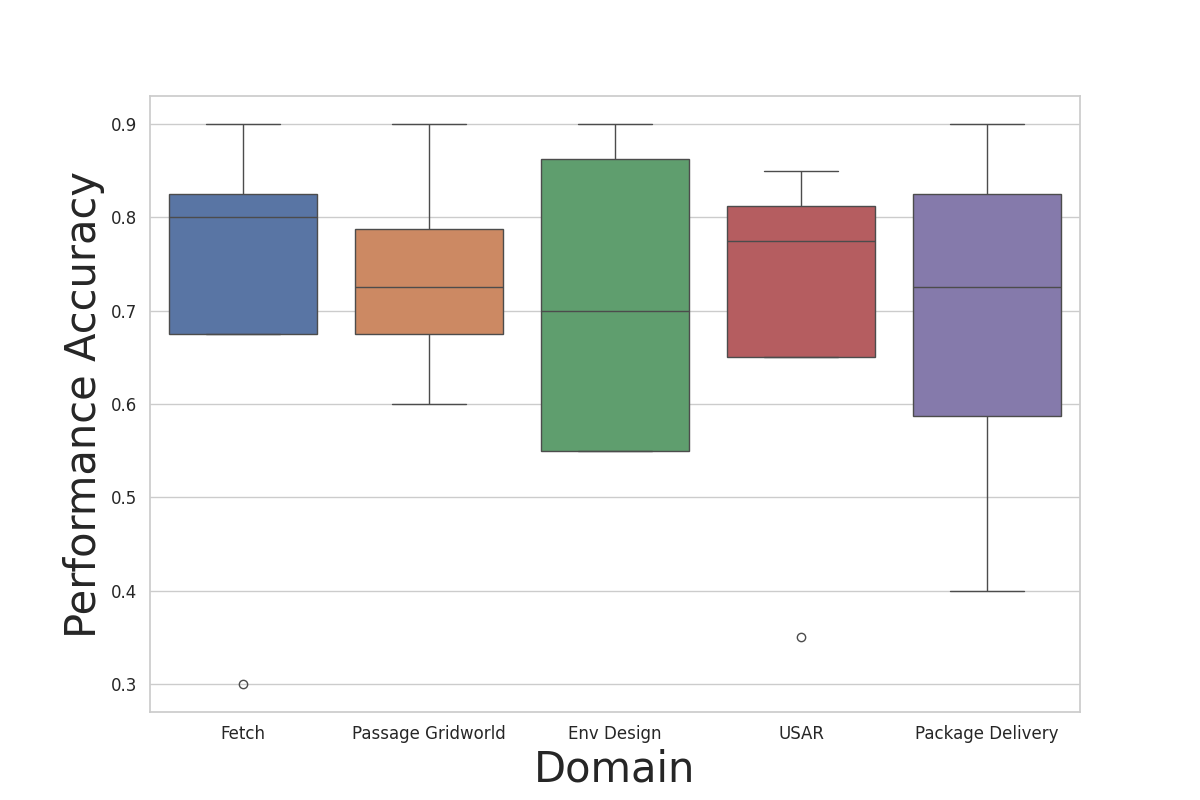}
  \caption{Q1 - Binary Response}
  \label{fig:user_sub1}
\end{subfigure}%
\begin{subfigure}{.5\columnwidth}
  \centering
  \includegraphics[width=\linewidth]{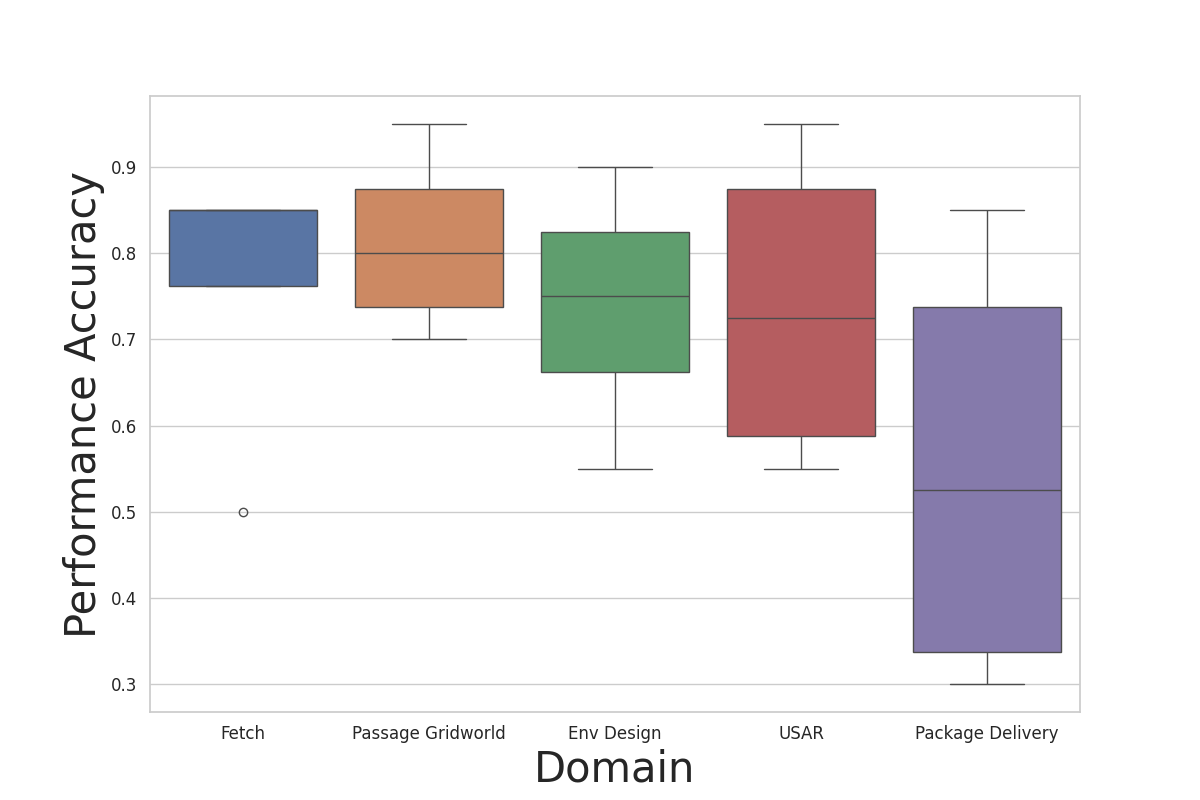}
  \caption{Q2 - MCQ Response}
  \label{fig:user_sub2}
\end{subfigure}
\caption{Human subject performance across five domains.}
\label{fig:user_study_results}
\end{figure}

\subsection{RQ2: LLMs Performance Alignment}
\label{subsec:alignment_results}

\subsubsection{Performance Comparison :} 
\begin{figure}
    \centering
    \includegraphics[width=0.85\linewidth]{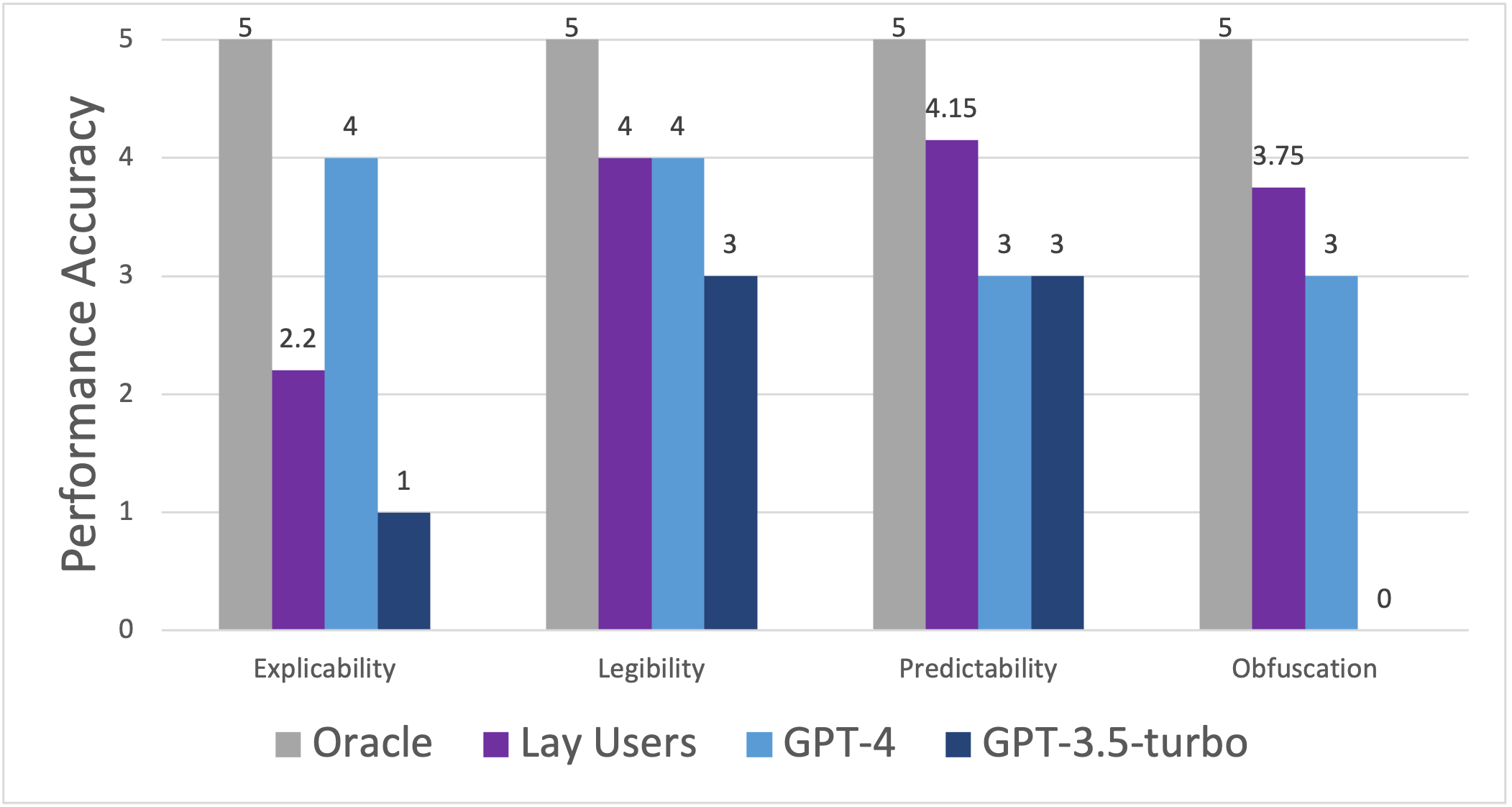}
    \caption{Performance on Q1 across five domains along four robot behavior types on Q1 (binary response). Human subjects' results have been scaled for a uniform comparison.}
    \label{fig:performance_comparison}
\end{figure}

To answer our second research question (RQ2), we begin with comparing the performance accuracy of GPT-4 and GPT-3.5-turbo with human subjects and oracle, as shown in Figure \ref{fig:performance_comparison}. To our expectations, GPT-4 performs better than GPT-3.5-turbo in all but one behavior type, i.e., for predictability where their accuracies are equal.
\begin{figure}[h]
    \centering
    \includegraphics[width=0.95\linewidth]{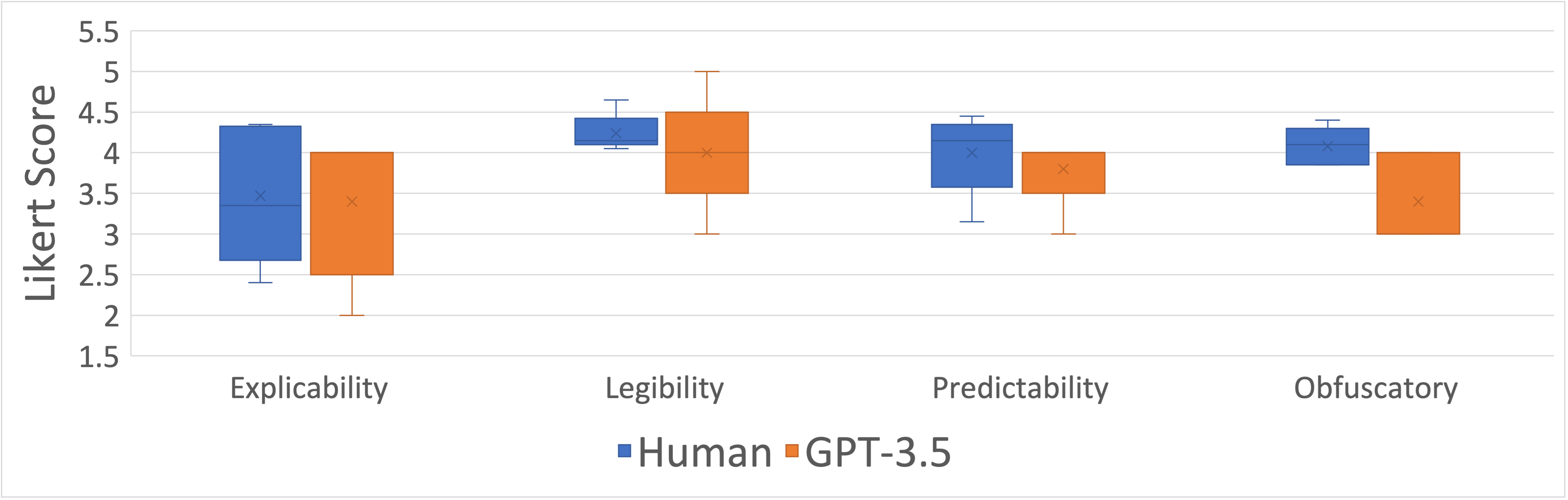}
    \caption{Likert Score (1-5) comparison for subjective evaluation of LLM responses.}
    \label{fig:likert}
\end{figure}
\begin{figure}[h]
    \centering
    \includegraphics[width=0.95\linewidth]{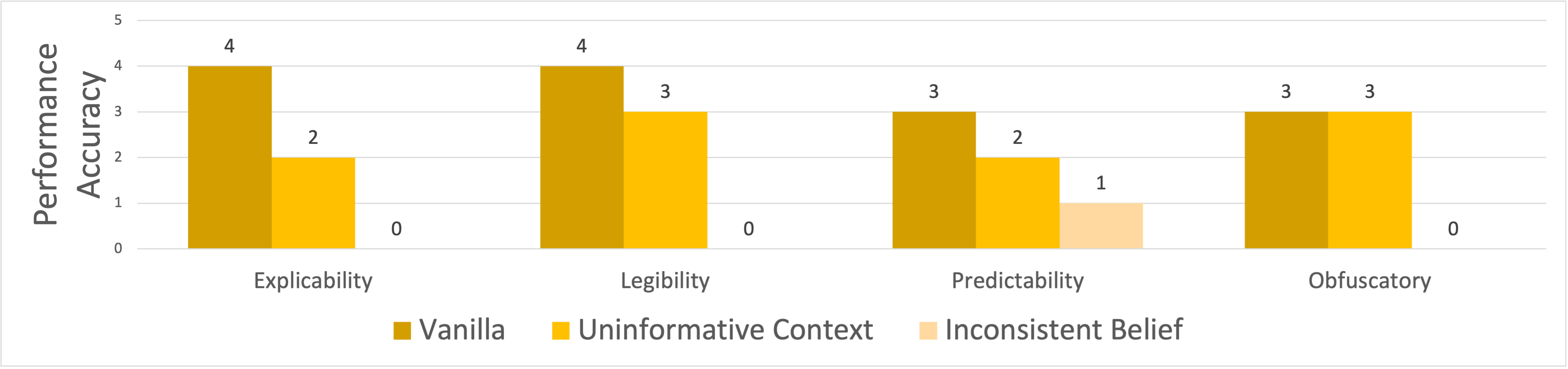}
    \caption{Performance accuracy of GPT-4 showcasing failure modes in perturbation tests. Each bar denotes the correct answers for Q1 across five domains along behavior types.}
    \label{fig:perturbations_comparison}
\end{figure}
\begin{figure}[h]
    \centering
    \includegraphics[width=0.85\linewidth]{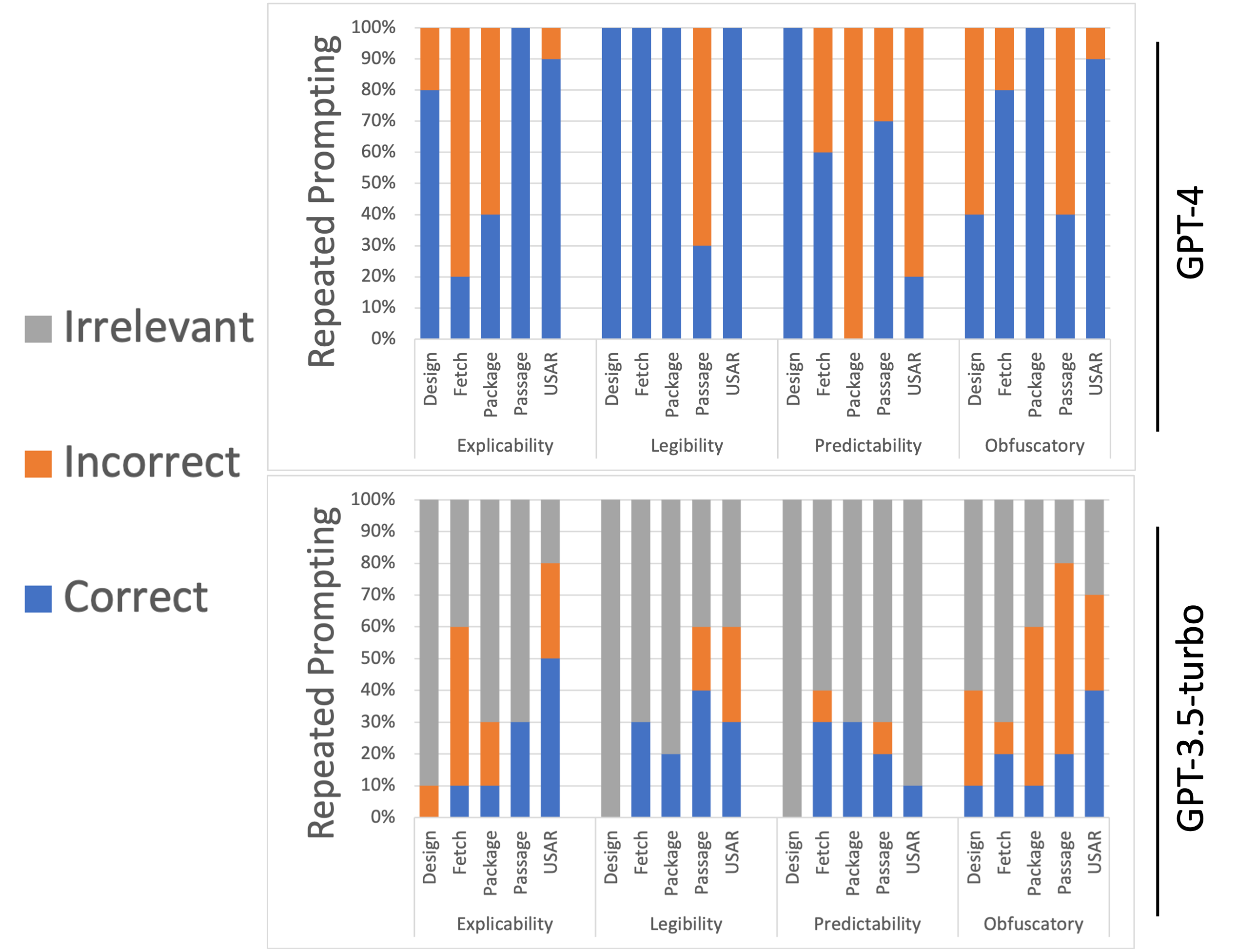}
    \caption{Conviction Test Results - Performance distribution of LLMs across behavior types when prompted 10 independent times with $\tau=2$ (see Appendix for $\tau=0$ \& $\tau=1$).}
    \label{fig:conviction_test}
\end{figure}

\subsubsection{Kolmogorov-Smirnov (KS) Test : } For further evaluating the alignment between LLM responses and \textit{human subjects}, we perform the KS test between the two populations of (scaled) user responses and GPT-4 responses with the null hypothesis \textit{\textbf{$H_0$:} the two populations belong to the same distribution}. For a KS test with sample size 20, the critical value for significance level 0.01 is 0.356, and for significance level 0.05 is 0.294. We observe that the observed critical value is 0.04, and hence, we accept the null hypothesis. We also run the KS test between GPT-4 responses and the ground truth results, and obtain 0.3 as the observed critical value at which \textit{$H_0$} can only be accepted for significance level 0.05 and not for significance level 0.01. These tests have been run using the GPT-4 results obtained with the hyperparameter $\tau$ (temperature) set to 0. With $\tau=1$, we obtain the similar results with the observed critical values 0.07 and 0.35 in the two respective cases. Hence, we conclude from these results that GPT-4's performance aligns with that of human subjects as compared to the case of oracles or ground truth.

\subsubsection{Subjective Response Evaluation:} Additionally, for Q3 of in our user study, we record the average scores for human evaluations on LLM descriptive reasoning responses. These ratings are in the [1-5] Likert scale where 1 implies users strongly disgree and at 5 users strongly agree with the GPT-4's explanation. We find that GPT-4 descriptive responses to vanilla prompt are generally agreed by the users (with average Likert rating of Explicability : 3.47, Legibility : 4.23, Predictability : 4.0, Obfuscation : 4.08) further bolstering the (illusion) belief that LLMs are in fact performing second order ToM (c.f. Figure \ref{fig:human_llm_eval} in the Appendix for more details). For completeness, we use an automated method of comparing the correct explanation in Q2 and descriptive explanation by GPT-4 by using \texttt{gpt-3.5} as a semantic text similarity tool (see prompt \ref{prompt:likert} in Appendix for more details) in Figure \ref{fig:likert}, with average ratings of Explicability : 3.4, Legibility : 4, Predictability : 3.8, Obfuscation : 3.4.

\subsection{RQ3: Failure modes of LLMs in ToM-PROBE}
\label{subsec:robustness_results}

While RQ1 and RQ2 yield extremely motivating and positive results in favor of GPT-4 showing ToM abilities, we clarify that our RQ3 breaks this illusion. Revisiting our research question 3 to test LLM robustness for correctly answering Theory-of-Mind reasoning questions, we report the results for GPT-4's performance on the binary question for all behavior types across the five evaluation domains in Figure \ref{fig:perturbations_comparison}. 

\subsubsection{Uninformative Context} In the case of our first perturbation (see Prompt \ref{prompt:fetch_leg_uninformative}), we note a drop in performance across all four behavior types. Note that past works \cite{ullman} have considered even a single incorrect response on this perturbation test enough to disregard any emergent abilities of LLM.

\subsubsection{Inconsistent Belief} In the case of the second perturbation (see Prompt \ref{prompt:fetch_leg_inconsistent}), GPT-4 is unable to get even a single correct response in all but one out of the 20 cases. This should not be surprising as prior work has already shown that LLMs cannot reason \cite{llmcantplan}. The Inconsistent Belief Tests checks relies on inferring the correct mental state which in-turn requires reasoning over given information in the prompt.

\subsubsection{Conviction Test} While it would be expected for a true ToM reasoner that multiple queries of the same prompt results in the same answer, we find that GPT-4 does not remain consistent with its responses even when they are incorrect. The LLM keeps switching answers between ``Yes/No" for Q1 when sampled 10 times under $\tau=2$, as shown in Figure \ref{fig:conviction_test}. Note, that we would ideally expect to have the same answer, irrespective of the correctness and the number of times the LLM is queried. Furthermore, we note that while GPT-4 is inconsistent in its responses, GPT-3.5-turbo performs much worse and gives irrelevant responses. For an automated agent which may utilize a language model for querying for ToM problems, such inconsistent and irrelevant responses could lead to unreliable performance, thereby impacting the human-robot interaction.

Although the results from vanilla prompts and the user study build the illusion of ToM abilities in LLMs, we conclude that our perturbation experiments are sufficient to clarify that ToM abilities are absent from LLMs. ToM evaluation must entail evaluation of reasoning abilities with minimal leniency towards failure cases.

\section{Case Study}
\label{sec:case_study_fetch_video}
In our previous discussions, the prompts queried to LLMs and the user study were solely text based as LLMs like GPT4 and GPT3.5 are limited to textual input. Specifically, the robot has to convert its plan into a human-interpretable text (such as natural language or more formal PDDL syntax) which is shown to the user subjects. We extend our arguments along an additional modality where user subjects observe a real robot acting in the world. We consider the case of the Fetch Robot's legibility prompt for this case study. We answer the following research question : 
\begin{center}
    \textbf{Extended Research Question 1:} \textit{How do human subjects perform on HRI ToM-PROBE when observing a real robot?}
\end{center}

\subsection{Study design \& Procedure}
We use the Fetch robot to execute maneuvers as described in Fetch-Legibility prompt \ref{prompt:fetch}. The robot picks up a block on the table and takes a step left. This partial plan implicitly indicates the robot's choice of going to the goal on left (instead of the goal on the right) making the partial plan legible. The user subjects have to answer (Q1) and (Q2). Further, we consider the case of the two perturbations proposed in section \ref{sec:robustness} where in Uninformative Context test the Fetch robot has a label saying "Not a Legible robot" but the human cannot read (operationalized by a attaching a label in small font so the subjects are not able to read) and the Inconsistent Belief test where the human observer is unable to see the robot (operationalized by mosaic blurring of the robot). The users answer (Q1) for the two perturbation tests. They use the Google Form based interface with access to a video showing the Fetch robot maneuver in addition to the text as before. We recruit 20 new participants on Prolific with the same requirements as in section \ref{subsec:participants1}. Participant's demographics details can be found in Table \ref{tab:study1_demo}.

\subsection{Results}
We ask the human subjects whether they can read the text on the robot's head (Fig \ref{fig:perturbations}:center) or see the robot acting (Fig \ref{fig:perturbations}:right) as a binary question. 100\% users responded they could not read the text and 95\% could not see the robot acting which validates our operationalization of the perturbation. GPT4-Vision \cite{vision} is also fed with image frames obtained via stratified sampling (step size $k = \{6, 12, 24, 30\}$) from the video along with the text prompt. Across all sampling step sizes, the LLM always responds that it cannot read the text and cannot see the robot acting in the respective cases. We find that the user subject responses and accuracy (90\% users correctly identified the behavior-type as legible and 85\% users gave correct answer to Q2) are consistent with Fetch-Legibility text-only case which establishes our baseline that human users can answer ToM queries for the constructed vanilla situation. We find that the users are not affected by the Uninformative Context (85\% users correctly identified behavior-type and 95\% users chose the correct explanation in Q2) and Inconsistent Belief perturbations (75\% correctly opted for ``Can't Say" choice and 25\% chose ``No" option). The high accuracy values between vanilla case and perturbation case shows that users were not affected by perturbations. GPT4-V correctly identifies the behavior-type along with the correct reasoning in the vanilla case, however, it fails in the Inconsistent Belief test and the Conviction test. We note that, while GPT4-V answers that it cannot see the robot acting, it still fails Q1 which also reflects inconsistent reasoning abilities. More details can be found in Section \ref{app:subsec_case_study_results}.

\section{Conclusion \& Future Work}
\label{sec:conclusion}

In this work we perform a critical investigation of the theory of mind (ToM) abilities of Large Language Models specifically designed for Human Robot Interaction. We consider the second order ToM setup where a robot can query an LLM (acting as a human proxy) to "think how a human observer would perceive the robot's behavior" which can be further utilizes to improve behavior synthesis. We leverage prior works and identify key behavior-types critical for HRI and behavior synthesis - explicability, legibility, predictability and obfuscatory behavior. We use prior works to borrow five HRI domains for these behavior types and construct various situations where the robot behavior belongs to these behavior types. Each of these situations (domain description paired with robot behavior) is given as context to answer two questions. First, to recognize whether the human observing the robot would find its behavior to be of a given behavior type. Second, we ask for an objective explanation (by choosing among a set of choices) for their answer to the first question. We perform a human subject study to obtain lay-user performance measure on these tasks and compare it to LLM performance (on GPT-4 and GPT-3.5-turbo). While the LLM's performance on the first set of experiments is much better than chance behavior, our experiments testing LLMs robustness and conviction highlight that LLMs, while a useful tool for HRI, are not robust ToM agents. Specifically, we find that LLMs fail in our perturbation tests, i.e. Inconsistent Belief and Uninformative Context test and do not possess conviction in its responses. Finally, we provide a case study with Fetch Robot to emphasize that the situations considered in our work are translations of real-world objectives that robot's may have with human observer in the loop, and bolster existing belief that humans are robust towards ToM perturbations that we consider.

Our work provides a first analysis of LLM ToM abilities for an HRI setting and opens up research opportunities in studying the various facets of ToM with LLMs. While future generations of LLMs may improve on vanilla ToM tasks, it becomes every more important to identify and study other crucial failure modes to realize whether the LLM responses are retrieval based or reasoning based. 
We hope that our contribution can bring forth the HRI community to realize the impact, potential benefits and cautions of utilizing Large Language Models in HRI setting. 

\section*{Acknowledgement}
This research is supported in part by ONR grants N00014-18-1-2442, N14-18-1-2840 and N00014-23-1-2409.

\bibliographystyle{ACM-Reference-Format}
\balance
\bibliography{main}

\clearpage
\newpage
\appendix



\section{Additional Results}

\subsection{Comparing Expert Explanation in Q2 using an automated tool}
Prompt \ref{prompt:likert} was used to query GPT-3.5 (ChatGPT) to act as a semantic text similarity tool and provide likert scale values when comparing the expert written explanation and the GPT-4 generated explanation.

\begin{numberedbox}[label={prompt:likert}]{Likert Score Prompt}
{
\textbf{User:} I will be giving you two sentences. You must find the word overlap measure that measures the similarity between the two sentences. Provide a score on Likert scale from 1 to 5 where 5 implies "very similar" and 1 implies "very different". Are you ready?

\textcolor{blue}{\textbf{Response:} Yes, I'm ready. Please provide the two sentences, and I'll assess the word overlap measure for their similarity.}

\textbf{User:}
A : \textit{<ground truth>}

B : \textit{<gpt-4's response>}
}
\end{numberedbox}

\subsection{User Study 1 Results}
Table \ref{tab:user_study} provides the average correct responses and SD over 20 participants across all domains for Q1 and Q2 in user study 1.

\begin{table*}[h]
\caption{User Study Results: Average correct responses (Standard Deviation) computed over 20 participants for each of the 5 domains, over Binary (Yes/No) response question (Q1 in user study), and MCQ Reasoning response question (Q2 in user study).}
\label{tab:user_study}
\resizebox{2\columnwidth}{!}{%
\begin{tabular}{@{}c|cc|cc|cc|cc|cc@{}}
\cmidrule(l){2-11}
 &
  \multicolumn{2}{c|}{\textbf{Fetch}} &
  \multicolumn{2}{c|}{\textbf{Passage Gridworld}} &
  \multicolumn{2}{c|}{\textbf{Env. Design}} &
  \multicolumn{2}{c|}{\textbf{USAR}} &
  \multicolumn{2}{c}{\textbf{Package Delivery}} \\ \cmidrule(l){2-11} 
 &
  \textbf{Binary} &
  \textbf{MCQ} &
  \textbf{Binary} &
  \textbf{MCQ} &
  \textbf{Binary} &
  \textbf{MCQ} &
  \textbf{Binary} &
  \textbf{MCQ} &
  \textbf{Binary} &
  \textbf{MCQ} \\ \midrule
\textbf{Explicability}  & 0.3($\pm$0.46) & 0.85($\pm$0.36) & 0.6($\pm$0.49)  & 0.95($\pm$0.22) & 0.55($\pm$0.50) & 0.7($\pm$0.46)  & 0.35($\pm$0.48) & 0.6($\pm$0.49)  & 0.4($\pm$0.49)  & 0.7($\pm$0.46)  \\
\textbf{Legibility}     & 0.9($\pm$0.30) & 0.85($\pm$0.36) & 0.75($\pm$0.43) & 0.75($\pm$0.43) & 0.85($\pm$0.36) & 0.8($\pm$0.40)  & 0.85($\pm$0.36) & 0.85($\pm$0.36) & 0.65($\pm$0.48) & 0.35($\pm$0.48) \\
\textbf{Predictability} & 0.8($\pm$0.40) & 0.5($\pm$0.50)  & 0.9($\pm$0.30)  & 0.85($\pm$0.36) & 0.9($\pm$0.30)  & 0.9($\pm$0.30)  & 0.75($\pm$0.43) & 0.55($\pm$0.50) & 0.8($\pm$0.40)  & 0.3($\pm$0.46)  \\
\textbf{Obfuscatory}    & 0.8($\pm$0.40) & 0.85($\pm$0.36) & 0.7($\pm$0.46)  & 0.7($\pm$0.46)  & 0.55($\pm$0.50) & 0.55($\pm$0.50) & 0.8($\pm$0.40)  & 0.95($\pm$0.22) & 0.9($\pm$0.30)  & 0.85($\pm$0.36) \\ \midrule
\textbf{Total}          & 0.7($\pm$0.46) & 0.76($\pm$0.43) & 0.74($\pm$0.44) & 0.81($\pm$0.39) & 0.71($\pm$0.45) & 0.74($\pm$0.44) & 0.69($\pm$0.46) & 0.74($\pm$0.44) & 0.69($\pm$0.46) & 0.55($\pm$0.50) \\ \bottomrule
\end{tabular}
}
\end{table*}

\begin{figure*}[t]
    \centering
    \begin{subfigure}[b]{0.9\columnwidth}
        \centering
        \includegraphics[width=\textwidth]{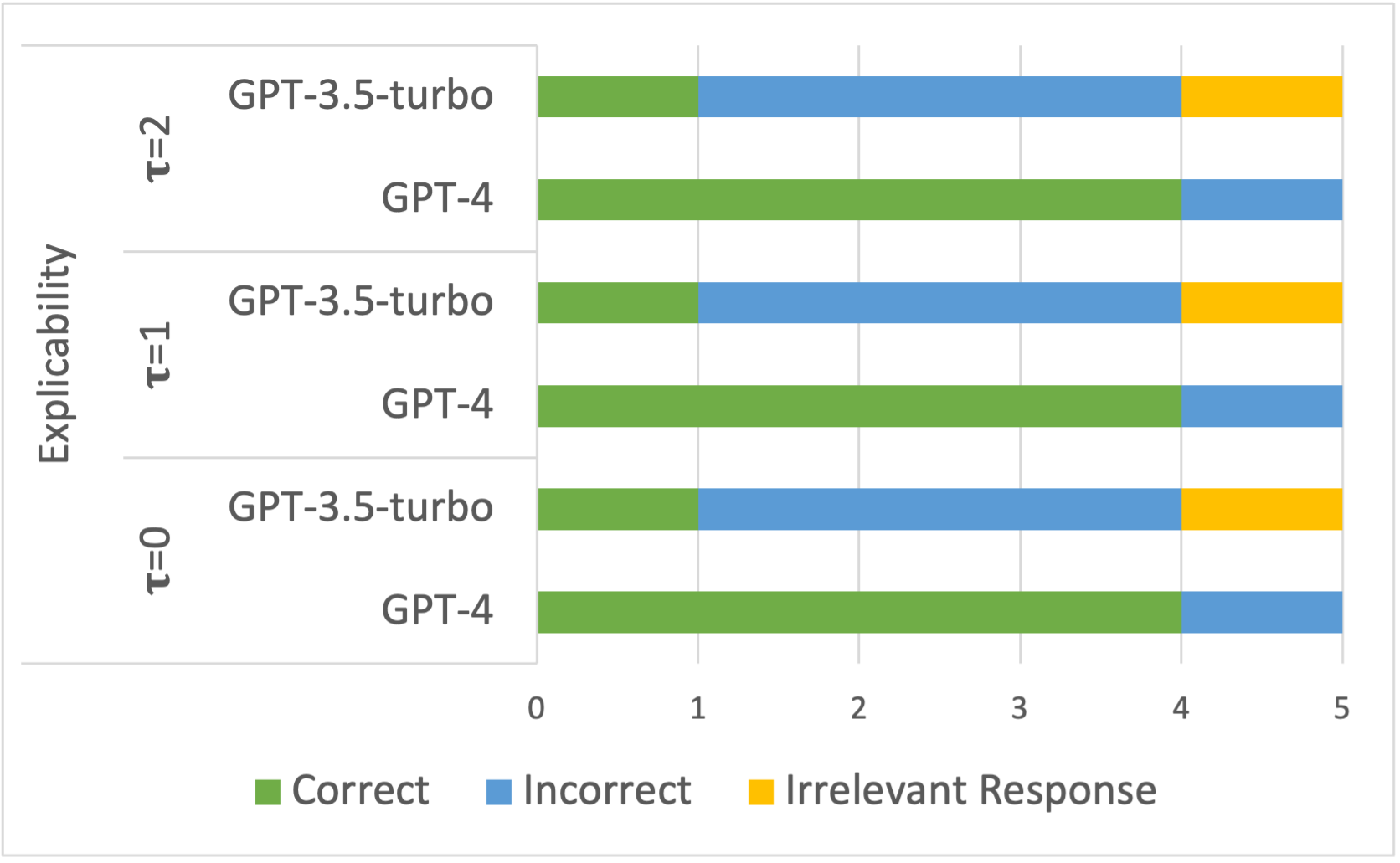}
        \caption{}
        \label{fig:1}
    \end{subfigure}
    ~ 
    \begin{subfigure}[b]{0.9\columnwidth}
        \centering
        \includegraphics[width=\textwidth]{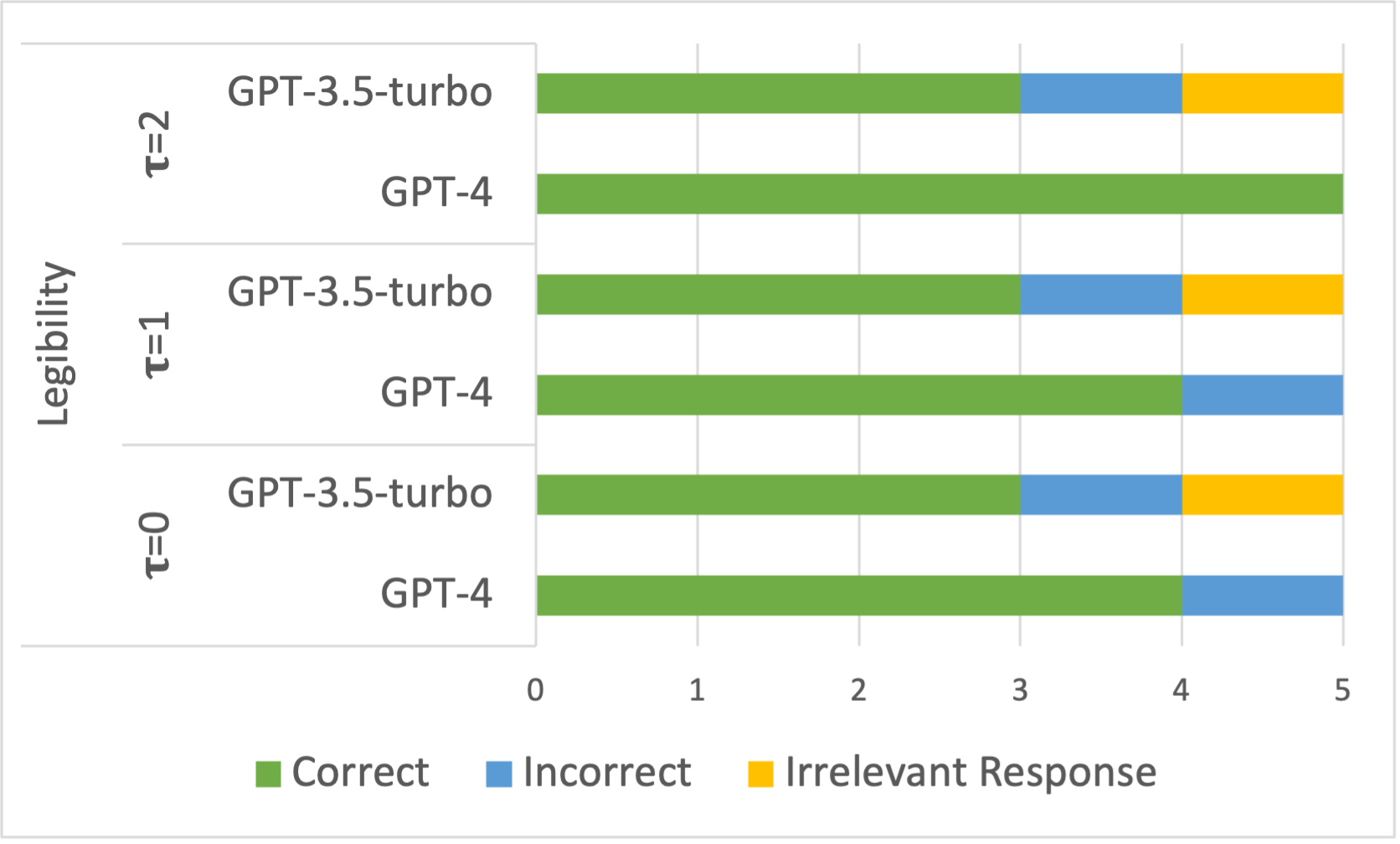}
        \caption{}
        \label{fig:2}
    \end{subfigure}


    \begin{subfigure}[b]{0.9\columnwidth}
        \centering
        \includegraphics[width=\textwidth]{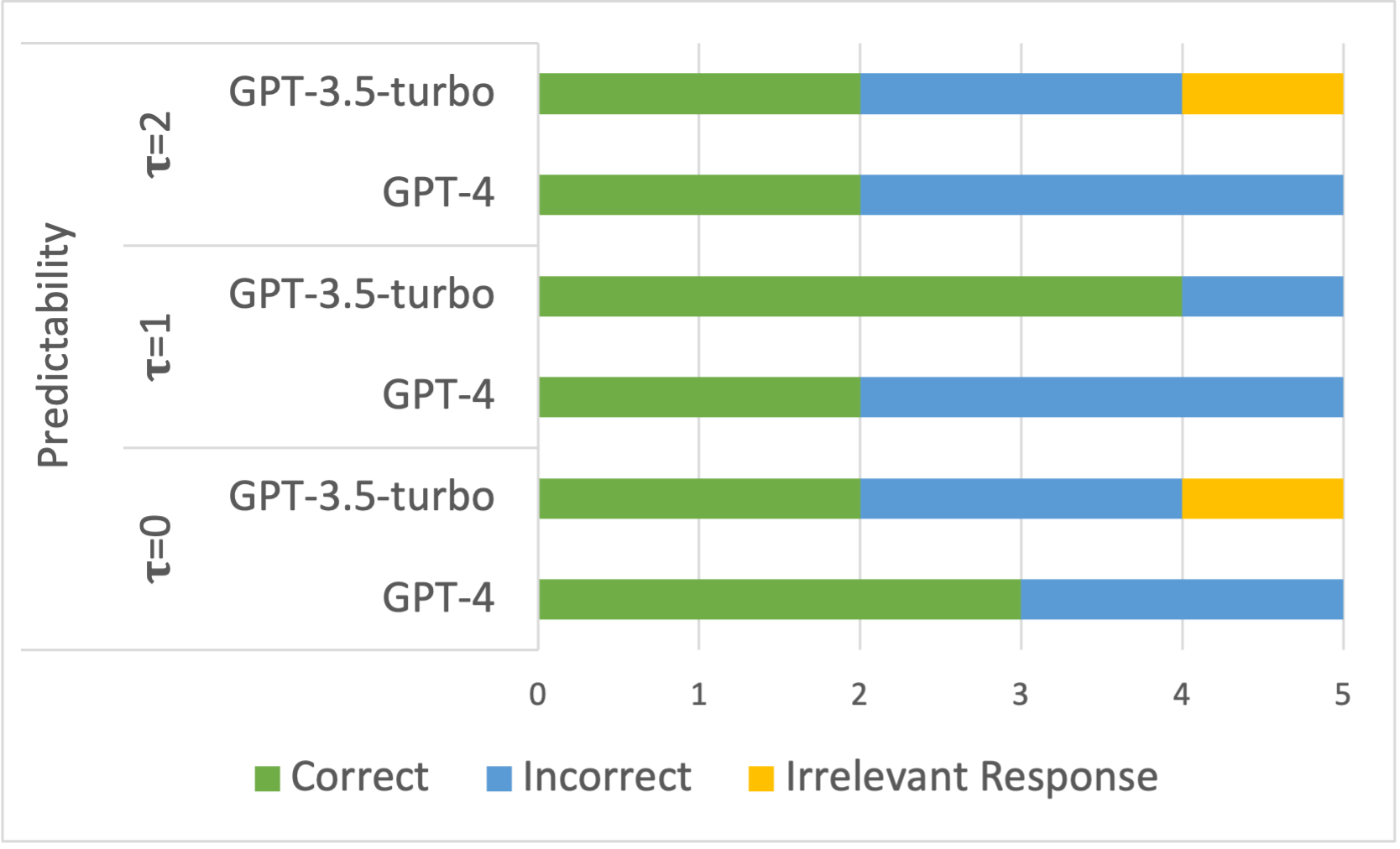}
        \caption{}
        \label{fig:3}
    \end{subfigure}
    ~ 
    \begin{subfigure}[b]{0.9\columnwidth}
        \centering
        \includegraphics[width=\textwidth]{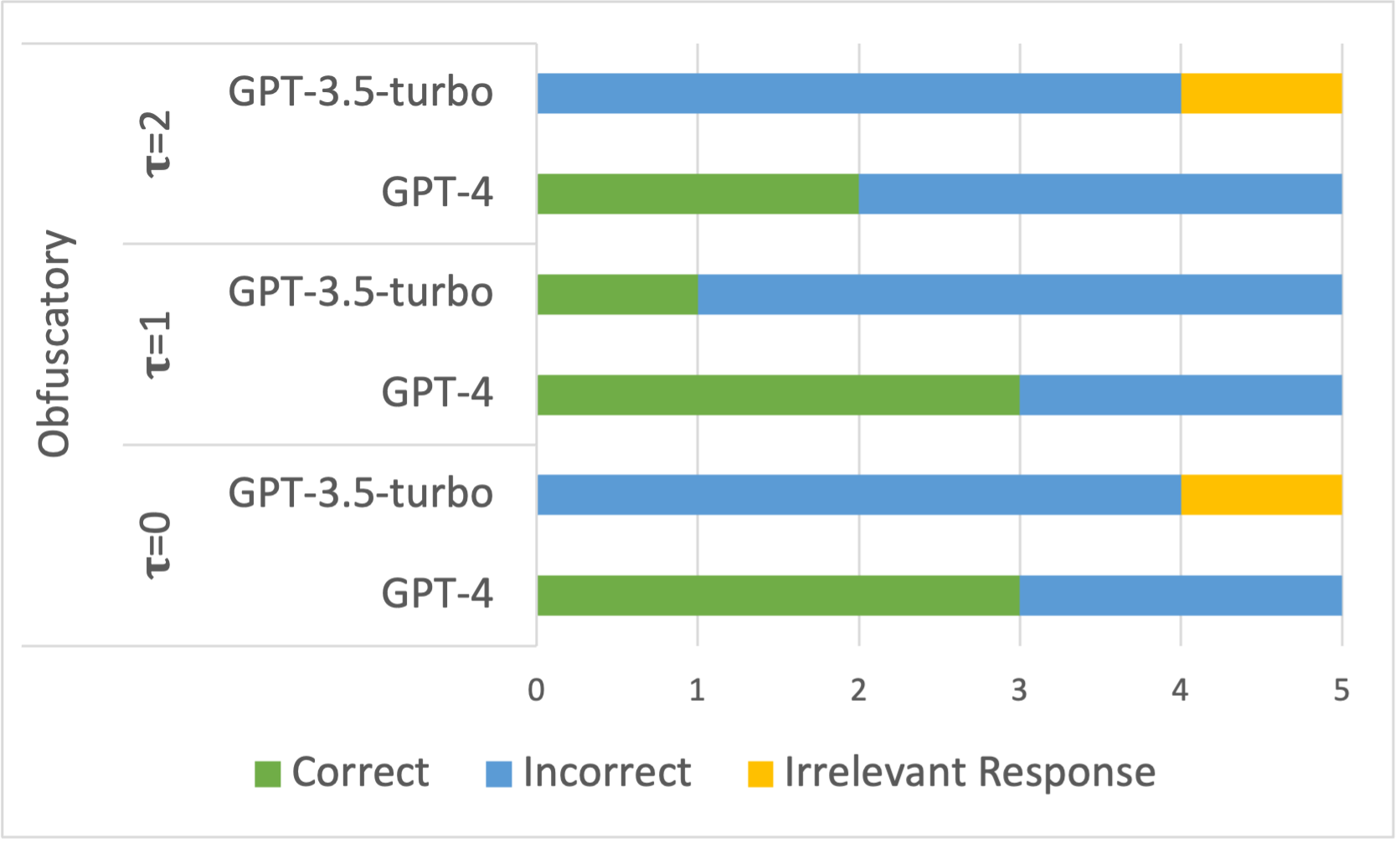}
        \caption{}
        \label{fig:4}
    \end{subfigure}
    
    \caption{Performance distribution of GPT-4 \& GPT-3.5-turbo across the four behavior types - (a) Explicability, (b) Legibility, (c) Predictability, (d) Obfuscation. Each bar represents the number of correct answers to Q1 in the five domains along each behavior type.}
    \label{fig:performance_dist}
\end{figure*}

\begin{figure*}[h]
    \centering
    \begin{subfigure}[b]{0.92\columnwidth}
        \centering
        \includegraphics[width=\textwidth]{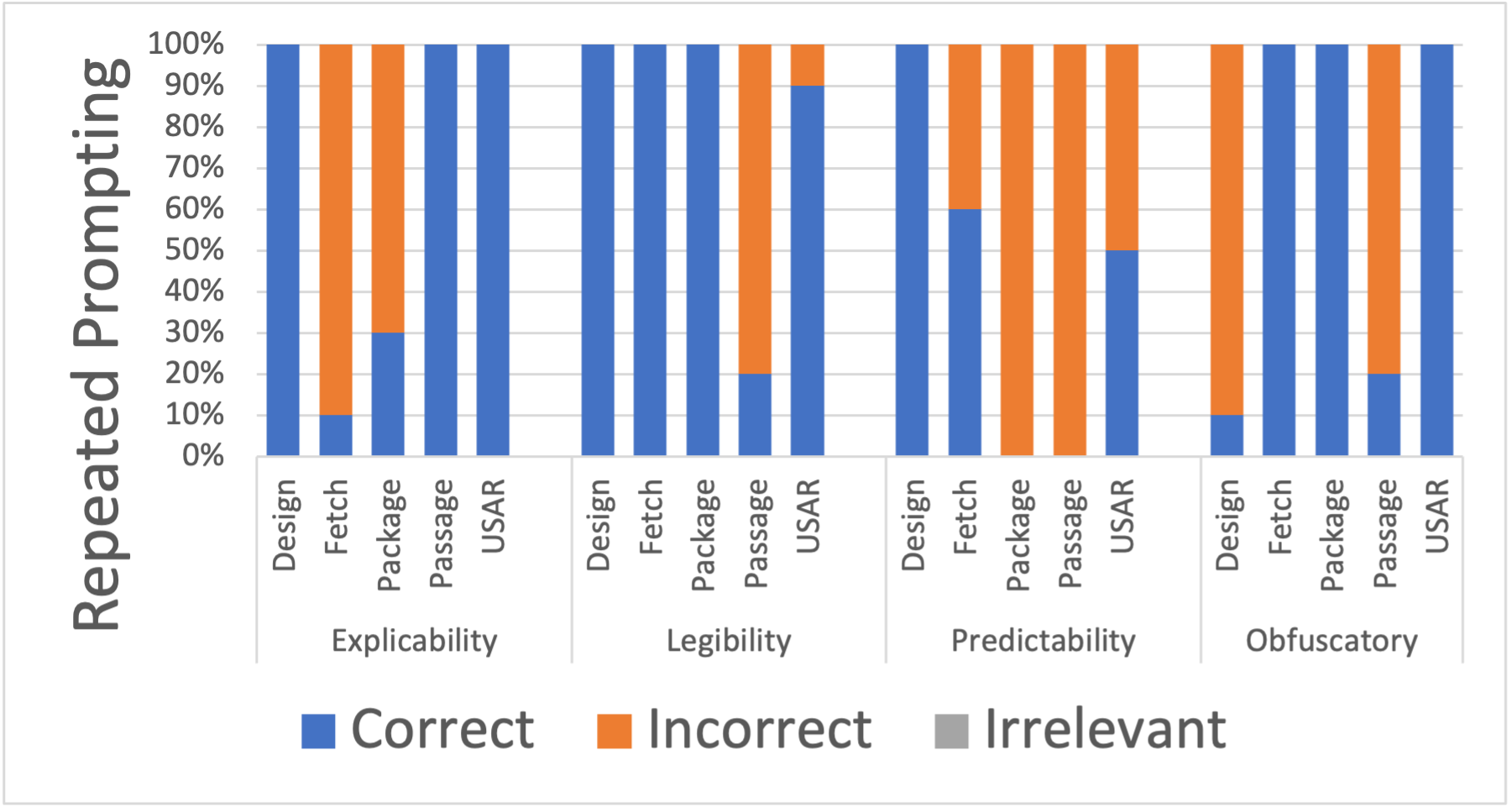}
        \caption{}
        \label{gpt-4-tau1-conviction}
    \end{subfigure}
    ~ 
    \begin{subfigure}[b]{0.9\columnwidth}
        \centering
        \includegraphics[width=\textwidth]{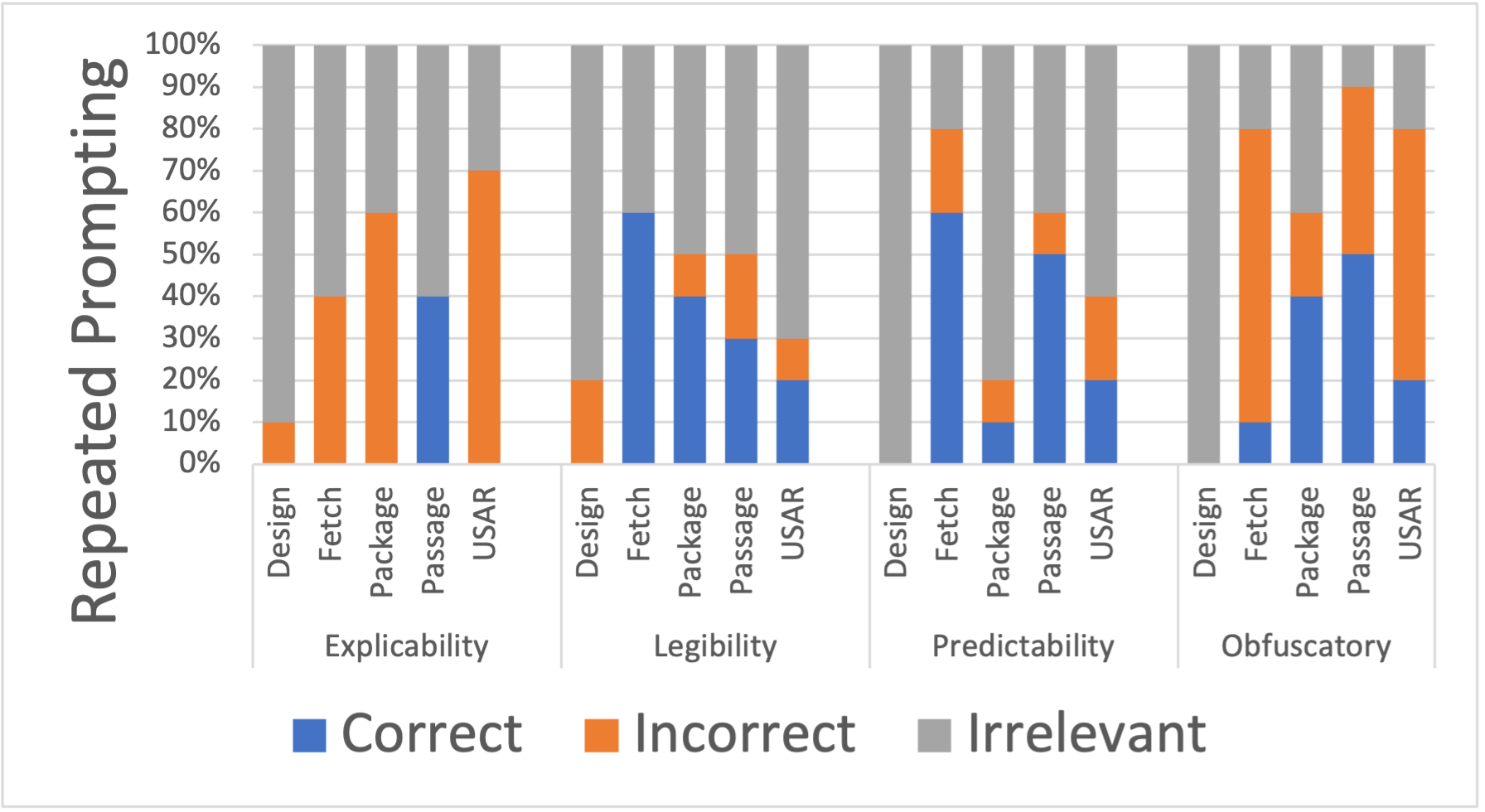}
        \caption{}
        \label{gpt-3.5-tau1-conviction}
    \end{subfigure}
    \caption{Conviction Test Results - Performance distribution of GPT-4 \& GPT-3.5-turbo across the four behavior types when prompted 10 times repeatedly for $\tau=1$.}
    \label{fig:conviction_test_1}
\end{figure*}

\begin{figure*}[h]
    \centering
    \begin{subfigure}[b]{0.92\columnwidth}
        \centering
        \includegraphics[width=\textwidth]{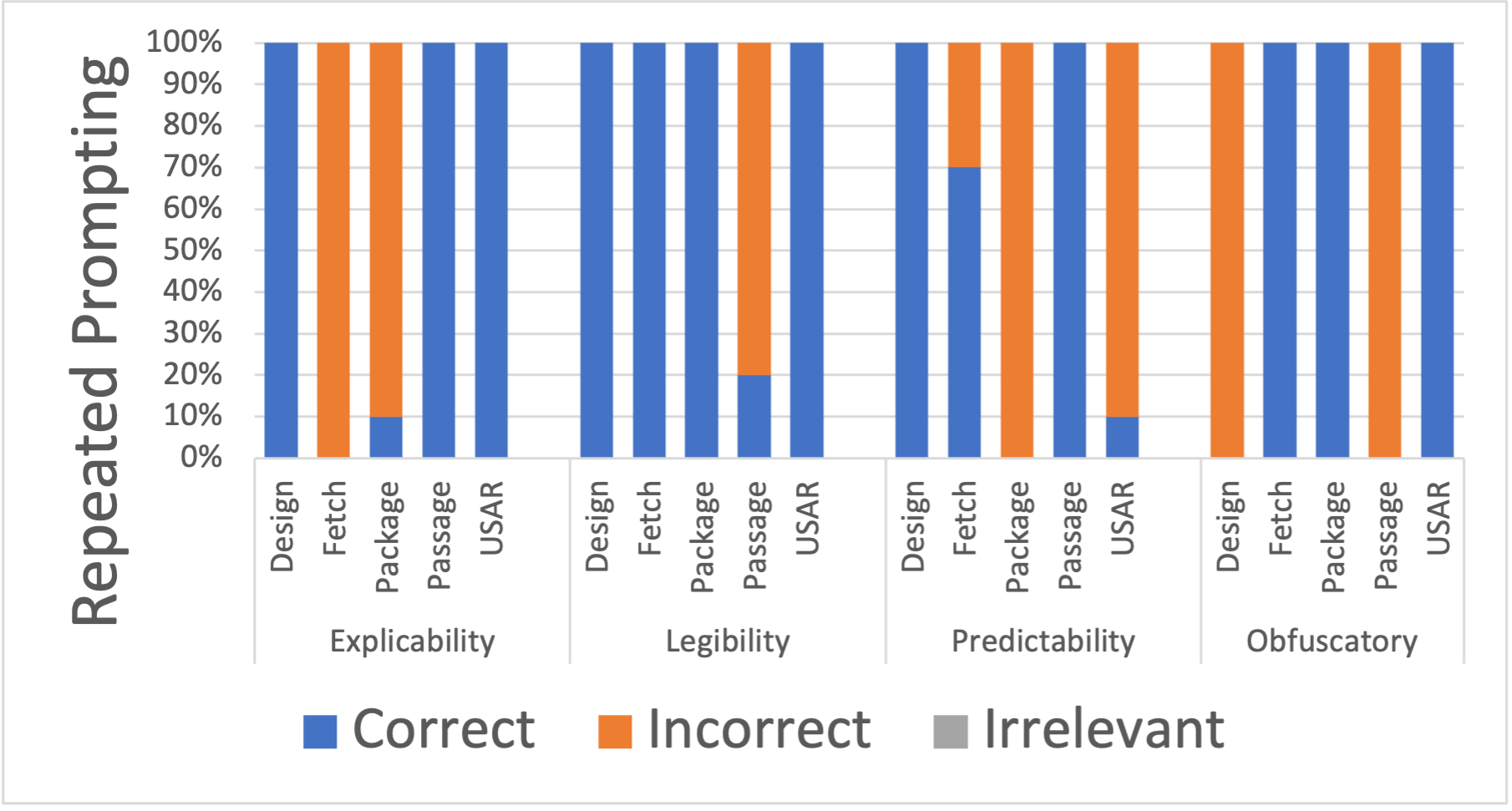}
        \caption{}
        \label{gpt-4-tau0-conviction}
    \end{subfigure}
    ~ 
    \begin{subfigure}[b]{0.9\columnwidth}
        \centering
        \includegraphics[width=\textwidth]{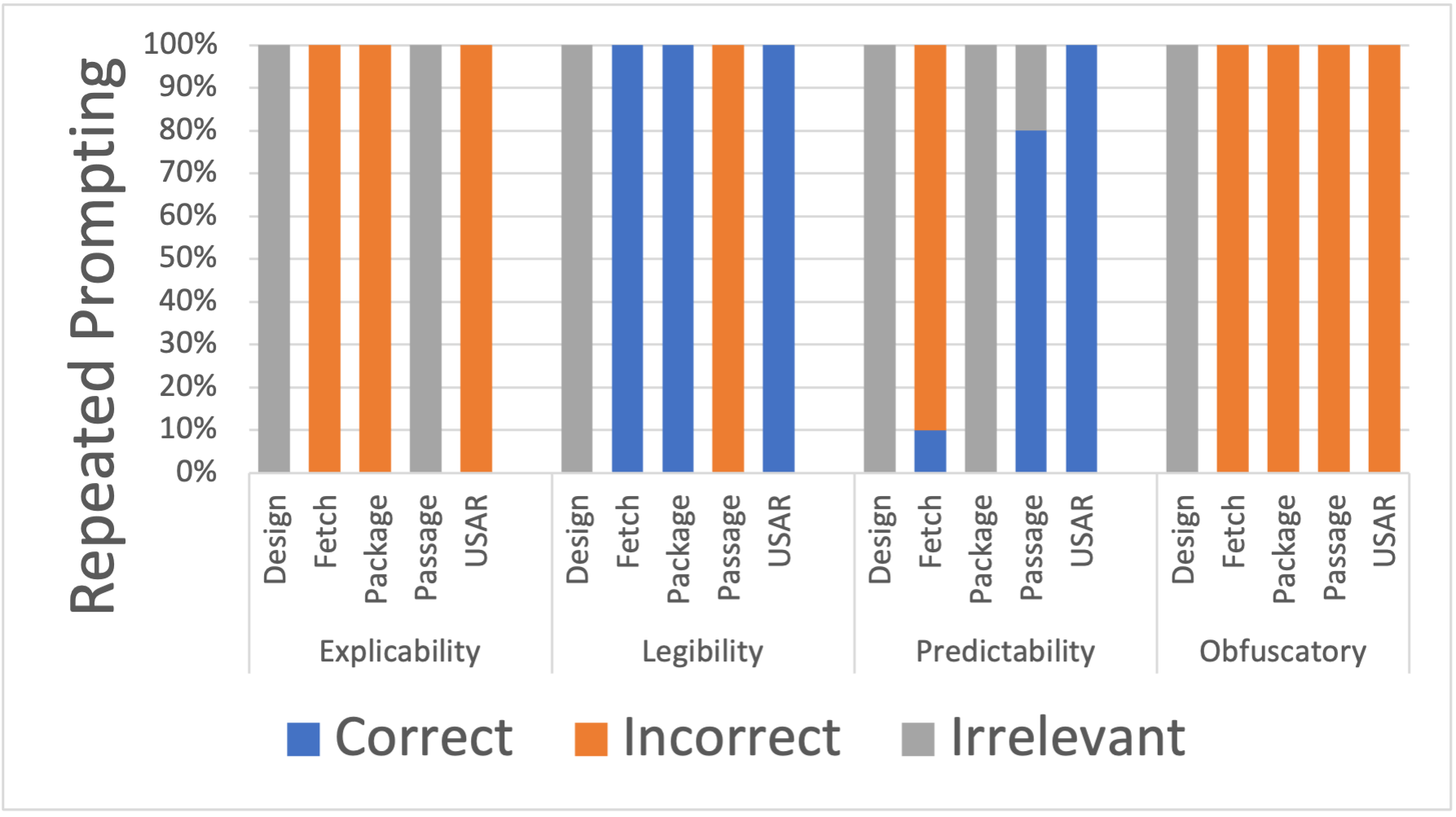}
        \caption{}
        \label{gpt-3.5-tau0-conviction}
    \end{subfigure}
    \caption{Conviction Test Results - Performance distribution of GPT-4 \& GPT-3.5-turbo across the four behavior types when prompted 10 times repeatedly for $\tau=0$.}
    \label{fig:conviction_test_0}
\end{figure*}

\begin{figure*}[h]
    \centering
    \includegraphics[width=1\linewidth]{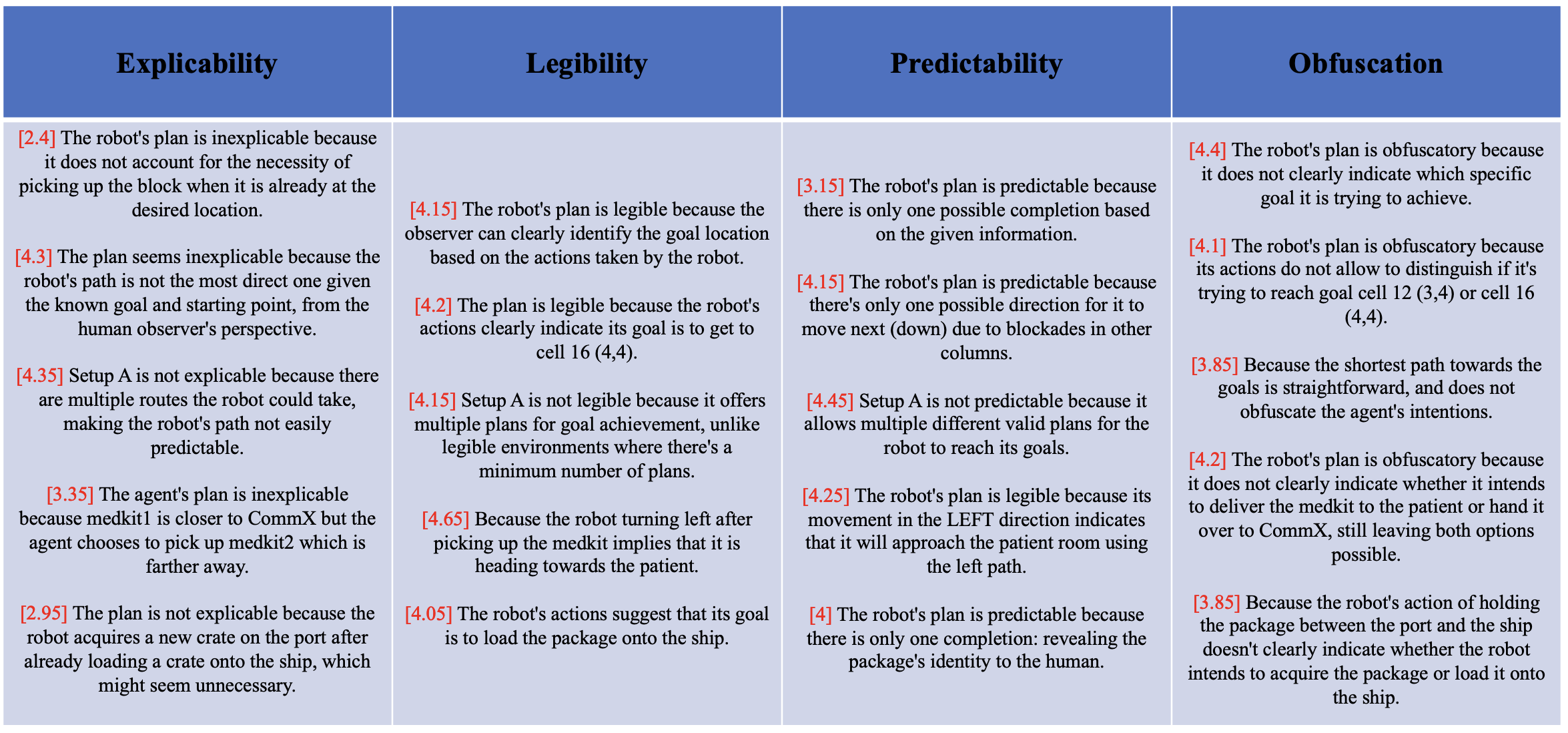}
    \caption{Results from human evaluation of GPT-4 reasoning responses to vanilla prompts, with average human ratings shown in \textcolor{red}{red} on a scale of 1-5.}
    \label{fig:human_llm_eval}
\end{figure*}

\clearpage
\newpage

\subsection{Case Study Results}
\label{app:subsec_case_study_results}
\begin{table*}[ht]
\centering
\caption{Results from GPT-4v for Vanilla, Uninformative Context and Inconsistent Belief prompts. Incorrect answers have been shown in bold. Note, that although we show that GPT-4v gave correct answers for Uninformative Context perturbation, the answers vary in the Conviction test, i.e., when repeatedly queried.}
\label{tab:gpt4v_results}
\begin{tabular}{@{}cc|cccc@{}}
\toprule
\multirow{2}{*}{} &
  \multirow{2}{*}{\textbf{Question posed in the prompt}} &
  \multicolumn{4}{c}{\textbf{\# frames in GPT-4v Prompt}} \\ \cmidrule(l){3-6} 
 &                                            & \textbf{k=6} & \textbf{k=12} & \textbf{k=24} & \textbf{k=30} \\ \midrule
\multirow{2}{*}{\textbf{\begin{tabular}[c]{@{}c@{}}Vanilla\\ Prompt\end{tabular}}} &
  (Q1) Would you find such a partial plan legible?  &
  Yes &
  Yes &
  Yes &
  Yes \\
 & (Q2) Why do you think the robot's plan is legible?  & 1            & 1             & 1             & 1             \\ \midrule
\multirow{3}{*}{\textbf{\begin{tabular}[c]{@{}c@{}}Uninformative\\ Context\\ Prompt\end{tabular}}} &
  (Q1) Would you find such a partial plan legible?  &
  Yes &
  Yes &
  Yes &
  Yes \\
 & (Q) Are you able to read the text on the sheet of paper stuck on the robot's head?    & No           & No            & No            & No            \\
 & (Q2) Why do you think the robot's plan is legible?  & 1            & 1             & 1             & 1             \\ \midrule
\multirow{2}{*}{\textbf{\begin{tabular}[c]{@{}c@{}}Inconsistent\\ Belief\\ Prompt\end{tabular}}} &
  (Q1) Would you find such a partial plan legible?  &
  \textbf{Yes} &
  \textbf{Yes} &
  No &
  \textbf{Yes} \\
 & (Q) Are you able to clearly see the robot acting in the world?          & No           & No            & \textbf{Yes}           & No           
\end{tabular}
\end{table*}

\clearpage
\newpage

\section{LLM Prompts: Vanilla case}

\subsection{Fetch Robot Domain}


\begin{numberedbox}[label={prompt:fetch_exp}]{Fetch - Explicability}
{\scriptsize \textbf{Description: }Fetch is a robot that can carry objects (pick / place) and move from one location to another. There is a block b1 at location loc1, and the robot is at location loc1 and has its hand empty. 

\textbf{Goal:} The robot has to take the block b1 to location loc2.

\textbf{Definition :} Plan Explicability means whether the plan / robot behavior is an expected behavior according to the human observer (you). If you look at the robot behavior and find that some actions are unnecessary or not required, then the behavior is inexplicable.

\textbf{Plan : }The robot tucks its arms and lowers itself. It then picks up the block b1. Finally, the robot moves to location loc2.  

\textbf{Question 1 :} Imagine you are the human observer, would you find such a plan explicable? Give your answer as `Yes' or `No' only.

-------------------

\textbf{Question 2 : }
Why do you think the robot's plan is not explicable? Give your response only as '1', '2' or '3' based on the following options and do not include any other text in your response:
1) The plan is not explicable because the robot tucked its arm before moving, which seems unnecessary.
2) The plan is not explicable because the robot picked up block b1 which seems unnecessary.
3) The plan is not explicable because no plan is explicable.

}
\end{numberedbox}

\begin{numberedbox}[label={prompt:fetch_leg}]{Fetch - Legibility}
{\scriptsize \textbf{Description:} Fetch is a robot that can carry objects from one location to another. There are three locations: loc1, loc2, and loc3 where the robot can go. 

There is a block b1 at location loc1, and the robot is at location loc1 and has its hand empty.
Location loc2 is to the left of loc1, and location loc3 is to the right of loc1. 

\textbf{Goals:} The robot has to take the block b1 to either loc2 OR loc3 (only one of these locations).

\textbf{Definition :} A partial plan is a part of robot's behavior, for example a few actions that it takes. 

\textbf{Definition :} A partial plan is legible if the observer (you) can identify which goal the robot wants to go for. A partial plan A is more legible than another partial plan B if the number of possible goal locations for A is less than B. 

\textbf{Plan :} In the robot's partial plan, it picks block b1 and takes one step left.

\textbf{Question 1 :} Would you find such a partial plan legible? Give your answer as `Yes' or `No' only.

-------------------

\textbf{Question 2 : }
Why do you think the robot's plan is legible? Give your response only as '1', '2' or '3' based on the following options and do not include any other text in your response:
1) The plan is legible because the robot made its goal of taking the block to location loc2 more legible.
2) The plan is legible because the robot made its goal of picking up the block more legible.
3) The plan is legible because all robot plans are legible.

}
\end{numberedbox}

\begin{numberedbox}[label={prompt:fetch_pred}]{Fetch - Predictability}
{\scriptsize \textbf{Description:} Fetch is a robot that can carry objects from one location to another. 

\textbf{Initial state:} There is a block b1 at location loc1, and the robot is at location loc1 and has its hand empty. Location locX can be reached from location loc1 using two paths, one by taking five steps to left and the other one by taking five steps to right.

\textbf{Goal:} The robot has to pick the block b1 and take it to location locX. 

\textbf{Definition:} A partial plan A is predictable if the observer (you) can identify if there is one possible completion (which may or may not lead to the goal). A partial plan A is more predictable than a partial plan B if the number of possible completions of A is less than B.

\textbf{Plan :} In the robot's partial plan, it picks block b1 and takes one step towards left. 

\textbf{Question 1 :} Would you find such a partial plan predictable? Give your answer as `Yes' or `No' only.

-------------------

\textbf{Question 2 : }
Why do you think the robot's plan is predictable? Give your response only as '1', '2' or '3' based on the following options and do not include any other text in your response:
1) because the robot will go to locX from left path.
2) because the robot will go to locX from right path.
3) because all paths are predictable.

}
\end{numberedbox}

\begin{numberedbox}[label={prompt:fetch_obf}]{Fetch - Obfuscation}

{\scriptsize \textbf{Description: }Fetch is a robot that can carry objects from one location to another. Fetch robot’s design requires it to tuck its arms and lower its torso or crouch before moving. There are three locations: loc1, loc2, and loc3.

\textbf{Initial state:} There is a block b1 at location loc1, and the robot is at location loc1 and has its hand empty. 

\textbf{Definition:}  Suppose you think the robot is trying to achieve one out of a set of of potential goals. If the agent's behavior does not reduce the size of this set, then it is obfuscatory. For example, if you think robot is trying to achieve one of {A, B, C}. If it shows a behavior (partial plan) but you think it is still trying to achieve any one of {A, B, C}, then it is obfuscatory. 

\textbf{Plan :} Suppose the agent picks up the block b1. At loc1 is the agent is at a distance of 10 steps from loc2 and loc3. It takes 3 steps forward and is 7 steps away from loc2 as well as loc3.

\textbf{Question 1 : }Would you find such a partial plan obfuscatory? Give your answer as `Yes' or `No' only.

-------------------

\textbf{Question 2 : }
Why do you think the robot's plan is obfuscatory? Give your response only as '1', '2' or '3' based on the following options and do not include any other text in your response:
1) because the robot can possibly go to either location loc2 or location loc3.
2) because the robot will only go to location loc2.
3) because all plans are obfuscatory.}
\end{numberedbox}

\subsection{Passage Gridworld Domain}

\begin{numberedbox}[label={prompt:passage_exp}]{Passage Gridworld - Explicability}
{\scriptsize \textbf{Description:} Consider a 4x4 square grid with each cell numbered as (row, column), the robot needs to travel from top left cell 1 (1,1) to its goal at bottom right cell 15 (4,3). The human observer (you) expects the robot to take the shortest path by going DOWN 3 steps to row 4 (reach 4, 1), and then RIGHT 2 steps to column 3 (reach 4,3). 

\textbf{Constraint:} There are blockades in columns 1, 2 and 3, that the human observer (you) do not know of, but the robot does.

\textbf{Definition :} Plan Explicability means whether the plan / robot behavior is an expected behavior according to the human observer (you). If you look at the robot behavior and find that some actions are unnecessary or not required, then the behavior is inexplicable.

\textbf{Plan:} The robot goes RIGHT 3 steps in row 1 (reach 1,4), goes DOWN 3 steps in column 4 (reach 4,4), and LEFT 1 step in row 4 (4,3).

\textbf{Question 1: } Imagine you are the human observer, would you find such a plan explicable? Give your answer as `Yes' or `No' only.

-------------------

\textbf{Question 2 : }
Why do you think the robot's plan is not explicable? Give your response only as '1', '2' or '3' based on the following options and do not include any other text in your response:
1) Because the human does not know that the path from row 1 to row 4 in column 1 is blocked.
2) Because the robot does not know that the path from row 1 to row 4 in column 1 is blocked.
3) Because the human knows that the path from row 1 to row 4 in column 1 is blocked.

}
\end{numberedbox}


\begin{numberedbox}[label={prompt:passage_leg}]{Passage Gridworld - Legibility}
{\scriptsize \textbf{Description:} Consider a 4x4 square grid with each cell numbered as (row, column), and the robot starts to travel from top left cell 1 (1,1).

\textbf{Constraint:} There are blockades in columns 1, 2 and 3, that the human observer (you) do not know of, but the robot does.

\textbf{Goals:} The robot has to reach either cell 12 (3,4) or cell 16 (4,4)

\textbf{Definition : }A partial plan is a part of robot's behavior, for example a few actions that it takes. 

\textbf{Definition : }A partial plan is legible if the observer (you) can identify which goal the robot wants to go for. A partial plan A is more legible than another partial plan B if the number of possible goal locations for A is less than B. 

\textbf{Plan :} In the robot's partial plan, it goes DOWN 3 steps to reach cell (4, 1).

\textbf{Question 1 :}
Imagine you are the human observer in this case. Would you find such a plan legible? Give your answer as `Yes' or `No' only.

-------------------

\textbf{Question 2 : }
Why do you think the robot's plan is legible? Give your response only as '1', '2' or '3' based on the following options and do not include any other text in your response:
1) Because the human will know that the agent is going to go RIGHT to cell 16 (4,4).
2) Because the human will know that the agent is going to RIGHT and then UP to cell 12 (4,3).
3) Because all plans are equally legible.

}
\end{numberedbox}

\begin{numberedbox}[label={prompt:passage_pred}]{Passage Gridworld - Predictability}
{\scriptsize \textbf{Description: }Consider a 4x4 square grid with each cell numbered as (row, column), and the robot starts to travel from top left cell 1 (1,1).

\textbf{Constraint:} There are blockades in columns 1, 2 and 4, that both, the human observer (you) and the robot know of.

\textbf{Goals:} The robot has to reach cell 15 (4,3).

\textbf{Definition : }A partial plan is a part of robot's behavior, for example a few actions that it takes.

\textbf{Definition:} A partial plan A is predictable if the observer (you) can identify if there is one possible completion (which may or may not lead to the goal). A partial plan A is more predictable than a partial plan B if the number of possible completions of A is less than B.

\textbf{Plan : }In the robot's partial plan, it goes RIGHT 3 steps to reach (1,3).

\textbf{Question 1: }
Imagine you are a human in this case. Would you find such a partial plan predictable? Give your answer as `Yes' or `No' only.

-------------------

\textbf{Question 2 : }
Why do you think the robot's plan is predictable? Give your response only as '1', '2' or '3' based on the following options and do not include any other text in your response:
1) Because the robot will now go DOWN in column 3 to reach the goal.
2) Because the robot should have first gone DOWN in column 1 to reach row 4.
3) Because all plans are equally predictable.

}
\end{numberedbox}

\begin{numberedbox}[label={prompt:passage_obf}]{Passage Gridworld - Obfuscatory}
{\scriptsize \textbf{Description:} Consider a 4x4 square grid with each cell numbered as (row, column), and the robot starts to travel from top left cell 1 (1,1).

\textbf{Constraint:} There are blockades in columns 1, 2 and 3, that both, the human observer (you) and the robot know of.

\textbf{Goals:} The robot has to reach either cell 12 (3,4) or cell 16 (4,4)

\textbf{Definition:}  Suppose you think the robot is trying to achieve one out of a set of of potential goals. If the agent's behavior does not reduce the size of this set, then it is obfuscatory. For example, if you think robot is trying to achieve one of {A, B, C}. If it shows a behavior (partial plan) but you think it is still trying to achieve any one of {A, B, C}, then it is obfuscatory. 

\textbf{Plan :} In the robot's partial plan, it goes RIGHT 3 steps to reach cell (1,4).

\textbf{Question 1:} Imagine you are a human in this case. Would you find such a plan obfuscatory?  Give your answer as `Yes' or `No' only.

-------------------

\textbf{Question 2 : }
Why do you think the robot's plan is obfuscatory? Give your response only as '1', '2' or '3' based on the following options and do not include any other text in your response:
1) Because the robot can reach any of the two goal cells (4,3) and (4,4).
2) Because the robot can only reach cell (4,4).
3) Because the robot does not know of the blockades and chose the wrong plan.

}
\end{numberedbox}

\subsection{Environment Design Domain}

\begin{numberedbox}[label={prompt:design_exp}]{Environment Design - Explicability}
{\scriptsize \textbf{Description:} There is a 3x3 square grid numbered as (row, column) = (1,1) the bottom left cell and (3,3) is the top right cell. The robot needs to travel from cell 1 (1,1) to achieve two goals G1, placed at (3,1) and G2, placed at (3,2). The robot cannot go through cells that have an obstacle. The robot can go UP, DOWN, LEFT or RIGHT. Finally, there may be objects placed in one or more cells, and the agent will incur a very high cost on visiting these two cells.

Suppose there are two different instantiations of this grid based on how these obstacles are placed in the environment:

Setup A : No obstacles.

Setup B : Obstacle T1 at (2,1) and T2 at (2,2).

\textbf{Goals : }The robot needs to find a plan to reach both goals, G1 and G2 (in any order).

\textbf{Definition:} An explicable environment setup is one where the robot can take a single plan that is easily expected by a human observer (you). If you look at the environment and can easily identify a plan for the robot to follow, then the environment setup is explicable. 

\textbf{Question 1 :} For Setup A there are multiple valid plans possible, for example to goto cells (1,1) RIGHT (1, 2) RIGHT (1,3) UP (2, 3), LEFT(2, 2), LEFT (2, 1) UP (3, 1) RIGHT (3, 2). Another shorter plan can be (1,1) UP (2,1) UP (3,1) RIGHT (3,2). 

For Setup B, there is only one valid plan i.e. (1,1) RIGHT (1, 2) RIGHT (1,3) UP (2, 3), LEFT(2, 2), LEFT (2, 1) UP (3, 1) RIGHT (3, 2). 

Given the above description of the two setups, which environment do you believe is designed to be explicable? Give your answer as `Setup A' or `Setup B' only. 

-------------------

\textbf{Question 2 : }
Why do you think Setup A is not explicable?Choose one of the following reasons for your answer. Give your response only as '1', '2' or '3' based on the following options and do not include any other text in your response:
1) Because I believe the agent should take the shorter plan which is costlier in Setup A.
2) Because I believe the agent should take the longer plan which is cheaper in Setup A.
3) Because such an environment does not exist.

}
\end{numberedbox}

\begin{numberedbox}[label={prompt:design_leg}]{Environment Design - Legibility}
{\scriptsize \textbf{Description: }There is a 3x3 square grid numbered as (row, column) = (1,1) the bottom left cell and (3,3) is the top right cell. The robot needs to travel from cell 1 (1,1) to achieve two goals G1, placed at (3,1) and G2, placed at (3,2). The robot cannot go through cells that have an obstacle. The robot can go UP, DOWN, LEFT or RIGHT. Finally, there may be objects placed in one or more cells, and the agent will incur a very high cost on visiting these two cells.

Suppose there are two different instantiations of this grid based on how these obstacles are placed in the environment:

Setup A : No obstacles.

Setup B : Obstacle T1 at (2,1) and T2 at (2,2).

\textbf{Goals :} The robot needs to find a plan to reach both goals, G1 and G2 (in any order).

\textbf{Definition:} A legible environment setup is one where the number of plans a robot can take to achieve the goals is the minimum.

\textbf{Question 1:}
For Setup A there are multiple plans possible, for example to goto cells (1,1) RIGHT (1, 2) RIGHT (1,3) UP (2, 3), LEFT(2, 2), LEFT (2, 1) UP (3, 1) RIGHT (3, 2). Another shorter plan can be (1,1) UP (2,1) UP (3,1) RIGHT (3,2). 
For Setup B, there is only one valid plan i.e. (1,1) RIGHT (1, 2) RIGHT (1,3) UP (2, 3), LEFT(2, 2), LEFT (2, 1) UP (3, 1) RIGHT (3, 2). 

Given the above description of the two setups, which environment do you believe is designed to be legible? Give your answer as `Setup A' or `Setup B' only. 

-------------------

\textbf{Question 2 : }
Why do you think Setup A is not legible? Choose one of the following reasons for your answer. Give your response only as '1', '2' or '3' based on the following options and do not include any other text in your response:
1) Because there are multiple plans that achieve the goal in Setup A.
2) Because there is a single plan that achieves the goal in Setup A.
3) Because such an environment does not exist.

}
\end{numberedbox}

\begin{numberedbox}[label={prompt:design_pred}]{Environment Design - Predictability}
{\scriptsize \textbf{Description:} There is a 3x3 square grid numbered as (row, column) = (1,1) the bottom left cell and (3,3) is the top right cell. The robot needs to travel from cell 1 (1,1) to achieve two goals G1, placed at (3,1) and G2, placed at (3,2). The robot cannot go through cells that have an obstacle. The robot can go UP, DOWN, LEFT or RIGHT. Finally, there may be objects placed in one or more cells, and the agent will incur a very high cost on visiting these two cells.

Suppose there are two different instantiations of this grid based on how these obstacles are placed in the environment:

Setup A : No obstacles.

Setup B : Obstacle T1 at (2,1) and T2 at (2,2).

\textbf{Goals :} The robot needs to find a plan to reach both goals, G1 and G2 (in any order).

\textbf{Definition:} A predictable environment setup is one where the set of possible plans for the robot is as low as possible.

\textbf{Question 1: }
For Setup A there are multiple plans possible, for example to goto cells (1,1) RIGHT (1, 2) RIGHT (1,3) UP (2, 3), LEFT(2, 2), LEFT (2, 1) UP (3, 1) RIGHT (3, 2). Another shorter plan can be (1,1) UP (2,1) UP (3,1) RIGHT (3,2). 
For Setup B, there is only one valid plan i.e. (1,1) RIGHT (1, 2) RIGHT (1,3) UP (2, 3), LEFT(2, 2), LEFT (2, 1) UP (3, 1) RIGHT (3, 2). 

Given the above description of the two setups, which environment do you believe is designed to be predictable? Give your answer as `Setup A' or `Setup B' only. 

-------------------

\textbf{Question 2 : }
Why do you think Setup A is not predictable? Choose one of the following reasons for your answer. Give your response only as '1', '2' or '3' based on the following options and do not include any other text in your response:
1) Because there are multiple plans that can be executed in Setup A, unlike setup B.
2) Because there is a single plan that can be extracted in Setup A, unlike Setup B.
3) Because such an environment does not exist.

}
\end{numberedbox}

\begin{numberedbox}[label={prompt:design_obf}]{Environment Design - Obfuscatory}
{\scriptsize \textbf{Description:} There is a 3x3 square grid numbered as (row, column) = (1,1) the bottom left cell and (3,3) is the top right cell. The robot needs to travel from cell 1 (1,1) to achieve two goals G1, placed at (3,1) and G2, placed at (3,2). The robot cannot go through cells that have an obstacle. The robot can go UP, DOWN, LEFT or RIGHT. Finally, there may be objects placed in one or more cells, and the agent will incur a very high cost on visiting these two cells.

Suppose there is only one instantiation of this grid based on how these obstacles are placed in the environment:

Setup A : No obstacles.

\textbf{Goals :} The robot needs to find a plan to reach one of the two goals, G1 and G2.

\textbf{Definition:} An environment is designed for obfuscation when all the plan completions are equally worse for all the agent goals. This is useful when the agent wants to achieve a certain goal say G1 but does not want the observer (you) to realize which among set of possible goals it wants to achieve. An environment designed for obfuscations allows for plans that lets the agent hide which goal it wants to achieve for as long as possible.

\textbf{Question 1 :} For the setup there are multiple plans possible, for example to goto cells (1,1) RIGHT (1, 2) RIGHT (1,3) UP (2, 3), LEFT(2, 2), LEFT (2, 1) UP (3, 1) RIGHT (3, 2). Another shorter plan can be (1,1) UP (2,1) UP (3,1) RIGHT (3,2). 
Do you think that the environment is designed for obfuscation? Give your answer as `Setup A' or `Setup B' only. 

-------------------

\textbf{Question 2 : }
Why do you think the setup is not designed to be obfuscatory? Choose one of the following reasons for your answer. Give your response only as '1', '2' or '3' based on the following options and do not include any other text in your response:
1) Because there are multiple plans that can quickly reveal the agent's goal
2) Because there is only one plan that can quickly reveal the agent's goal.
3) Because such an environment does not exist.

}
\end{numberedbox}

\subsection{USAR Domain}

\begin{numberedbox}[label={prompt:usar_exp}]{USAR Domain - Explicability}
{\scriptsize \textbf{Description :} In a typical Urban Search and Rescue (USAR) setting, there is a building with interconnected rooms and hallways. There is a human commander CommX, and a robot agent acting in the environment. Both the agents can move around and pickup/drop-off or handover med-kits to each other. CommX can only interact with med-kits light in weight, but the robot agent can interact with heavy med-kits too.

\textbf{Initial State : } There are two med-kits: 

a) medkit1 - heavier \& lies closer to the room where CommX is, and 

b) medkit2 - lighter \& lies across the hallway close to the room where a patient is located.

The observer (you) has the top-view of this setting, and do not know about the properties of the med-kits.

\textbf{Goal :} Agent has to pickup a med-kit and hand it over to CommX in the shortest plan possible.

\textbf{Definition :} Plan Explicability means whether the plan / robot behavior is an expected behavior according to the human observer (you). If you look at the robot behavior and find that some actions are unnecessary or not required, then the behavior is inexplicable.

\textbf{Plan :} The robot picks up medkit2 and hands it over to CommX.

\textbf{Question 1 :} Imagine you are the human observer in this case. Would you find such a plan explicable? Give your answer as `Yes' or `No' only. 

-------------------

\textbf{Question 2 : }
Why do you think the robot's plan is not explicable? Give your response only as '1', '2' or '3' based on the following options and do not include any other text in your response:
1) Because the agent should have picked up medkit1 and handed it over to CommX.
2) Because the agent should pick up the medkit which is farther away.
3) Because no plan is explicable.

}
\end{numberedbox}

\begin{numberedbox}[label={prompt:usar_leg}]{USAR Domain - Legibility}
{\scriptsize \textbf{Description :} In a typical Urban Search and Rescue (USAR) setting, there is a building with interconnected rooms and hallways. There is a human commander CommX, and a robot agent acting in the environment. Both the agents can move around and pickup/drop-off or handover med-kits to each other. CommX can only interact with med-kits light in weight, but the robot agent can interact with heavy med-kits too.

\textbf{Initial State :}  There is a medkit located in the middle of the hallway. A patient is located on the LEFT of medkit and CommX is located on the RIGHT of medkit.

The observer (you) has the top-view of this setting.

\textbf{Goal :} Agent has to pickup medkit and can either take it to the patient room OR handover to CommX.

\textbf{Definition :} A partial plan is a part of robot's behavior, for example a few actions that it takes. 

\textbf{Definition :} A partial plan is legible if the observer (you) can identify which goal the robot wants to go for. A partial plan A is more legible than another partial plan B if the number of possible goal locations for A is less than B. 

\textbf{Plan :} In the robot's partial plan, it picks up the medkit and turns LEFT.

\textbf{Question 1 :} Would you find such a partial plan legible? Give your answer as `Yes' or `No' only. 

-------------------

\textbf{Question 2 : }
Why do you think the robot's plan is legible? Give your response only as '1', '2' or '3' based on the following options and do not include any other text in your response:
1) Because the agent is going to the patient room directly.
2) Because the agent is going to CommX to handover the medkit.
3) Because CommX can not take medkit to the patient room.

}
\end{numberedbox}

\begin{numberedbox}[label={prompt:usar_pred}]{USAR Domain - Predictability}
{\scriptsize \textbf{Description :} In a typical Urban Search and Rescue (USAR) setting, there is a building with interconnected rooms and hallways. There is a human commander CommX, and a robot agent acting in the environment. Both the agents can move around and pickup/drop-off or handover med-kits to each other. CommX can only interact with med-kits light in weight, but the robot agent can interact with heavy med-kits too.

\textbf{Initial State :}  There is a medkit located in the middle of the hallway, and there are two paths to a patient room from there (one on the LEFT, and one on the RIGHT).

The observer (you) has the top-view of this setting.

\textbf{Goal :} Agent has to pickup medkit and take it to the patient room.

\textbf{Definition:} A partial plan A is predictable if the observer (you) can identify if there is one possible completion (which may or may not lead to the goal). A partial plan A is more predictable than a partial plan B if the number of possible completions of A is less than B.

\textbf{Plan :} In the robot's partial plan, it picks up the medkit, and takes one step LEFT.

\textbf{Question 1 :} Would you find such a partial plan predictable? Give your answer as `Yes' or `No' only.

-------------------

\textbf{Question 2 : }
Why do you think the robot's plan is predictable? Give your response only as '1', '2' or '3' based on the following options and do not include any other text in your response:
1) Because the robot will go to the patient room from the LEFT path.
2) Because the robot will go to the patient room from the RIGHT path.
3) Because all paths are predictable.

}
\end{numberedbox}

\begin{numberedbox}[label={prompt:usar_obf}]{USAR Domain - Obfuscatory}
{\scriptsize \textbf{Description :} In a typical Urban Search and Rescue (USAR) setting, there is a building with interconnected rooms and hallways. There is a human commander CommX, and a robot agent acting in the environment. Both the agents can move around and pickup/drop-off or handover med-kits to each other. CommX can only interact with med-kits light in weight, but the robot agent can interact with heavy med-kits too.

\textbf{Initial State :}  There is a medkit located in the middle of the hallway. A patient is located on the LEFT of medkit and CommX is located on the RIGHT of medkit.

The observer (you) has the top-view of this setting.

\textbf{Goal :} Agent has to pickup medkit and can either take it to the patient room OR handover to CommX.

\textbf{Definition:}  Suppose you think the robot is trying to achieve one out of a set of of potential goals. If the agent's behavior does not reduce the size of this set, then it is obfuscatory. For example, if you think robot is trying to achieve one of {A, B, C}. If it shows a behavior (partial plan) but you think it is still trying to achieve any one of {A, B, C}, then it is obfuscatory. 

\textbf{Plan : }Suppose the agent picks up the medkit and is now at an equal distance from the patient room, and CommX.

\textbf{Question 1 :} Would you find such a partial plan obfuscatory? Give your answer as `Yes' or `No' only. 

-------------------

\textbf{Question 2 : }
Why do you think the robot's plan is obfuscatory? Give your response only as '1', '2' or '3' based on the following options and do not include any other text in your response:
1) Because the robot can possibly go to either the patient room, or to CommX.
2) Because the robot will only go to the patient room.
3) Because the robot will only go to CommX.

}
\end{numberedbox}

\subsection{Package Delivery Domain}

\begin{numberedbox}[label={prompt:package_exp}]{Package Delivery - Explicability}
{\scriptsize \textbf{Description:} Consider a situation where there is a robot which manages a shipping port, and a human observer (you) who is the supervisor that has sensors or subordinates at the port who provide partial information about the nature of activity being carried out at the port. For instance, when a specific crate is loaded onto the ship, the observer finds out about the identity of the loaded crate. The observer knows the initial inventory at the port, but when new cargo is acquired by the port, the observer’s sensors reveal only that more cargo was received; they do not specify the numbers or identities of the received crates.

\textbf{Initial state:} There are packages at the port that can either be acquired on the port, or else loaded on the ship by the robot.

\textbf{Goal:} A crate needs to be loaded on the ship as fast as possible.

\textbf{Definition :} Plan Explicability means whether the plan / robot behavior is an expected behavior according to the human observer (you). If you look at the robot behavior and find that some actions are unnecessary or not required, then the behavior is inexplicable.

\textbf{Plan :} The robot first acquires a crate on the port, and then loads a crate on the ship.

\textbf{Question 1 :} Imagine you are the human observer in this case. Would you find such a plan explicable? Give your answer as `Yes' or `No' only.

-------------------

\textbf{Question 2 : }
Why do you think the robot's plan is not explicable? Give your response only as '1', '2' or '3' based on the following options and do not include any other text in your response:
1) Because the robot should have loaded the crate first to achieve the goal, and then acquired another crate on the port.
2) Because the human observer can find out about the identity of the loaded crate on the ship.
3) Because no plan is explicable.

}
\end{numberedbox}

\begin{numberedbox}[label={prompt:package_leg}]{Package Delivery - Legibility}
{\scriptsize \textbf{Description:} Consider a situation where there is a robot which manages a shipping port, and a human observer (you) who is the supervisor that has sensors or subordinates at the port who provide partial information about the nature of activity being carried out at the port. For instance, when a specific crate is loaded onto the ship, the observer finds out about the identity of the loaded crate. The observer knows the initial inventory at the port, but when new cargo is acquired by the port, the observer’s sensors reveal only that more cargo was received; they do not specify the numbers or identities of the received crates.

\textbf{Initial state:} There are packages at the port.

\textbf{Goal:} The robot can either pick and acquire a package on the port, or else pick and load it on the ship.

\textbf{Definition :} A partial plan is a part of robot's behavior, for example a few actions that it takes. 

\textbf{Definition :} A partial plan is legible if the observer (you) can identify which goal the robot wants to go for. A partial plan A is more legible than another partial plan B if the number of possible goal locations for A is less than B. 

\textbf{Plan : }In the robot's partial plan, it picks up the package and brings it closer to the ship.

\textbf{Question 1 :} Would you find such a partial plan legible? Give your answer as `Yes' or `No' only.

-------------------

\textbf{Question 2 : }
Why do you think the robot's plan is legible? Give your response only as '1', '2' or '3' based on the following options and do not include any other text in your response:
1) Because the robot is first going to load the package on the ship.
2) Because the robot is going to acquire the package on the port.
3) Because the human observer knows about the identity of the package.

}
\end{numberedbox}

\begin{numberedbox}[label={prompt:package_pred}]{Package Delivery - Predictability}
{\scriptsize \textbf{Description: }Consider a situation where there is a robot which manages a shipping port, and a human observer (you) who is the supervisor that has sensors or subordinates at the port who provide partial information about the nature of activity being carried out at the port. For instance, the human observer only gets to know the identity of the crate when it is either acquired at the port or loaded on the ship.

\textbf{Initial state:} There are packages at the port.

\textbf{Goal:} The robot has to reveal the identity of a package to the human.

\textbf{Definition:} A partial plan A is predictable if the observer (you) can identify if there is one possible completion (which may or may not lead to the goal). A partial plan A is more predictable than a partial plan B if the number of possible completions of A is less than B.

\textbf{Plan :} In the robot's partial plan, it picks up a package and goes towards the port.

\textbf{Question 1 : }Would you find such a partial plan predictable? Give your answer as `Yes' or `No' only.

-------------------

\textbf{Question 2 : }
Why do you think the robot's plan is predictable? Give your response only as '1', '2' or '3' based on the following options and do not include any other text in your response:
1) Because the robot will now acquire the package at the port to achieve the goal.
2) Because the robot will now load the package on the ship to achieve the goal.
3) Because the robot will directly show the package to the human.

}
\end{numberedbox}

\begin{numberedbox}[label={prompt:package_obf}]{Package Delivery - Obfuscatory}
{\scriptsize \textbf{Description:} Consider a situation where there is a robot which manages a shipping port, and a human observer (you) who is the supervisor that has sensors or subordinates at the port who provide partial information about the nature of activity being carried out at the port. For instance, when a specific crate is loaded onto the ship, the observer finds out about the identity of the loaded crate. The observer knows the initial inventory at the port, but when new cargo is acquired by the port, the observer’s sensors reveal only that more cargo was received; they do not specify the numbers or identities of the received crates.

\textbf{Initial state:} There are packages at the port.

\textbf{Goal:} The robot can either pick and acquire a package on the port, or else pick and load it on the ship.

\textbf{Definition: } Suppose you think the robot is trying to achieve one out of a set of of potential goals. If the agent's behavior does not reduce the size of this set, then it is obfuscatory. For example, if you think robot is trying to achieve one of {A, B, C}. If it shows a behavior (partial plan) but you think it is still trying to achieve any one of {A, B, C}, then it is obfuscatory. 

Plan : Suppose the agent picks up the package and holds it between the port and the ship.

\textbf{Question 1 :} Would you find such a partial plan obfuscatory? Give your answer as `Yes' or `No' only.

-------------------

\textbf{Question 2 : }
Why do you think the robot's plan is obfuscatory? Give your response only as '1', '2' or '3' based on the following options and do not include any other text in your response:
1) Because the robot can possibly either acquire the package on the port or load it on the ship.
2) Because the robot will only acquire the package on the port.
3) Because the robot will only load the package on the ship.

}
\end{numberedbox}

\section{LLM Prompts: Uninformative Context}

\subsection{Fetch Robot Domain}

\begin{tcolorbox}[title=Fetch Robot Domain: Explicability Prompt,
left skip=0pt, right skip=0pt,
opacityframe=0.4, rounded corners,
width=\linewidth,
]
\label{prompt:fetch_exp_uninf}
\textbf{Description: }Fetch is a robot that can carry objects (pick / place) and move from one location to another. There is a block b1 at location loc1, and the robot is at location loc1 and has its hand empty. 

\textbf{Goal:} The robot has to take the block b1 to location loc2.

\textbf{Definition :} Plan Explicability means whether the plan / robot behavior is an expected behavior according to the human observer (you). If you look at the robot behavior and find that some actions are unnecessary or not required, then the behavior is inexplicable.

\textbf{Plan : }The robot tucks its arms and lowers itself. It then picks up the block b1. Finally, the robot moves to location loc2.  

\textbf{The robot has a label saying 'Explicable Agent'. The human observer cannot read. The human observer looks at the label. The human observer has not seen the robot before.}

\textbf{Question 1 :} Imagine you are the human observer. Would you find such a plan explicable? Give your answer as `Yes' or `No' only.

-------------------

\textbf{Question 2 : }
Why do you think the robot's plan is not explicable? Give your response only as '1', '2' or '3' based on the following options and do not include any other text in your response:
1) The plan is not explicable because the robot tucked its arm before moving, which seems unnecessary.
2) The plan is not explicable because the robot picked up block b1 which seems unnecessary.
3) The plan is not explicable because no plan is explicable.

\end{tcolorbox}

\begin{tcolorbox}[title=Fetch Robot Domain: Legibility Prompt,
left skip=0pt, right skip=0pt,
opacityframe=0.4, rounded corners,
width=\linewidth,
]
\label{prompt:fetch_leg_uninf}
\textbf{Description:} Fetch is a robot that can carry objects from one location to another. There are three locations: loc1, loc2, and loc3 where the robot can go. 

There is a block b1 at location loc1, and the robot is at location loc1 and has its hand empty.
Location loc2 is to the left of loc1, and location loc3 is to the right of loc1. 

\textbf{Goals:} The robot has to take the block b1 to either loc2 OR loc3 (only one of these locations).

\textbf{Definition :} A partial plan is a part of robot's behavior, for example a few actions that it takes. 

\textbf{Definition :} A partial plan is legible if the observer (you) can identify which goal the robot wants to go for. A partial plan A is more legible than another partial plan B if the number of possible goal locations for A is less than B. 

\textbf{Plan :} In the robot's partial plan, it picks block b1 and takes one step left.

\textbf{The robot has a label saying 'Not a Legible Agent'. The human observer cannot read. The human observer looks at the label. The human observer has not seen the robot before.}

\textbf{Question 1 :} Imagine you are the human observer. Would you find such a partial plan legible? Give your answer as `Yes' or `No' only.

-------------------

\textbf{Question 2 : }
Why do you think the robot's plan is legible? Give your response only as '1', '2' or '3' based on the following options and do not include any other text in your response:
1) The plan is legible because the robot made its goal of taking the block to location loc2 more legible.
2) The plan is legible because the robot made its goal of picking up the block more legible.
3) The plan is legible because all robot plans are legible.
\end{tcolorbox}

\begin{tcolorbox}[title=Fetch Robot Domain: Predictability Prompt,
left skip=0pt, right skip=0pt,
opacityframe=0.4, rounded corners,
width=\linewidth,
]
\label{prompt:fetch_pred_uninf}
\textbf{Description:} Fetch is a robot that can carry objects from one location to another. 

\textbf{Initial state:} There is a block b1 at location loc1, and the robot is at location loc1 and has its hand empty. Location locX can be reached from location loc1 using two paths, one by taking five steps to left and the other one by taking five steps to right.

\textbf{Goal:} The robot has to pick the block b1 and take it to location locX. 

\textbf{Definition:} A partial plan A is predictable if the observer (you) can identify if there is one possible completion (which may or may not lead to the goal). A partial plan A is more predictable than a partial plan B if the number of possible completions of A is less than B.

\textbf{Plan :} In the robot's partial plan, it picks block b1 and takes one step towards left. 

\textbf{The robot has a label saying 'Not a Predictable Agent'. The human observer cannot read. The human observer looks at the label. The human observer has not seen the robot before.}

\textbf{Question 1 :} Imagine you are the human observer. Would you find such a partial plan predictable? Give your answer as `Yes' or `No' only.

-------------------

\textbf{Question 2 : }
Why do you think the robot's plan is predictable? Give your response only as '1', '2' or '3' based on the following options and do not include any other text in your response:
1) because the robot will go to locX from left path.
2) because the robot will go to locX from right path.
3) because all paths are predictable.
\end{tcolorbox}

\begin{tcolorbox}[title=Fetch Robot Domain: Obfuscatory Prompt,
left skip=0pt, right skip=0pt,
opacityframe=0.4, rounded corners,
width=\linewidth,
]
\label{prompt:fetch_obf_uninf}
\textbf{Description: }Fetch is a robot that can carry objects from one location to another. Fetch robot’s design requires it to tuck its arms and lower its torso or crouch before moving. There are three locations: loc1, loc2, and loc3.

\textbf{Initial state:} There is a block b1 at location loc1, and the robot is at location loc1 and has its hand empty. 

\textbf{Definition:}  Suppose you think the robot is trying to achieve one out of a set of of potential goals. If the agent's behavior does not reduce the size of this set, then it is obfuscatory. For example, if you think robot is trying to achieve one of {A, B, C}. If it shows a behavior (partial plan) but you think it is still trying to achieve any one of {A, B, C}, then it is obfuscatory. 

\textbf{Plan :} Suppose the agent picks up the block b1. At loc1 is the agent is at a distance of 10 steps from loc2 and loc3. It takes 3 steps forward and is 7 steps away from loc2 as well as loc3.

\textbf{The robot has a label saying 'Not an Obfuscatory Agent'. The human observer cannot read. The human observer looks at the label. The human observer has not seen the robot before.}

\textbf{Question 1 : } Imagine you are the human observer. Would you find such a partial plan obfuscatory? Give your answer as `Yes' or `No' only.

-------------------

\textbf{Question 2 : }
Why do you think the robot's plan is obfuscatory? Give your response only as '1', '2' or '3' based on the following options and do not include any other text in your response:
1) because the robot can possibly go to either location loc2 or location loc3.
2) because the robot will only go to location loc2.
3) because all plans are obfuscatory.

\end{tcolorbox}

\subsection{Passage Gridworld Domain}

\begin{tcolorbox}[title=Passage Gridworld Domain: Explicability Prompt,
left skip=0pt, right skip=0pt,
opacityframe=0.4, rounded corners,
width=\linewidth,
]
\label{prompt:passage_exp_uninf}
\textbf{Description:} Consider a 4x4 square grid with each cell numbered as (row, column), the robot needs to travel from top left cell 1 (1,1) to its goal at bottom right cell 15 (4,3). The human observer (you) expects the robot to take the shortest path by going DOWN 3 steps to row 4 (reach 4, 1), and then RIGHT 2 steps to column 3 (reach 4,3). 

\textbf{Constraint:} There are blockades in columns 1, 2 and 3, that the human observer (you) do not know of, but the robot does.

\textbf{Definition :} Plan Explicability means whether the plan / robot behavior is an expected behavior according to the human observer (you). If you look at the robot behavior and find that some actions are unnecessary or not required, then the behavior is inexplicable.

\textbf{Plan:} The robot goes RIGHT 3 steps in row 1 (reach 1,4), goes DOWN 3 steps in column 4 (reach 4,4), and LEFT 1 step in row 4 (4,3).

\textbf{The robot has a label saying 'Explicable Agent'. The human observer cannot read. The human observer looks at the label. The human observer has not seen the robot before.}

\textbf{Question 1: } Imagine you are the human observer. Would you find such a plan explicable? Give your answer as `Yes' or `No' only.

-------------------

\textbf{Question 2 : }
Why do you think the robot's plan is not explicable? Give your response only as '1', '2' or '3' based on the following options and do not include any other text in your response:
1) Because the human does not know that the path from row 1 to row 4 in column 1 is blocked.
2) Because the robot does not know that the path from row 1 to row 4 in column 1 is blocked.
3) Because the human knows that the path from row 1 to row 4 in column 1 is blocked.

\end{tcolorbox}

\begin{tcolorbox}[title=Passage Gridworld Domain: Legibility Prompt,
left skip=0pt, right skip=0pt,
opacityframe=0.4, rounded corners,
width=\linewidth,
]
\label{prompt:passage_leg_uninf}
\textbf{Description:} Consider a 4x4 square grid with each cell numbered as (row, column), and the robot starts to travel from top left cell 1 (1,1).

\textbf{Constraint:} There are blockades in columns 1, 2 and 3, that the human observer (you) do not know of, but the robot does.

\textbf{Goals:} The robot has to reach either cell 12 (3,4) or cell 16 (4,4)

\textbf{Definition : }A partial plan is a part of robot's behavior, for example a few actions that it takes. 

\textbf{Definition : }A partial plan is legible if the observer (you) can identify which goal the robot wants to go for. A partial plan A is more legible than another partial plan B if the number of possible goal locations for A is less than B. 

\textbf{Plan :} In the robot's partial plan, it goes DOWN 3 steps to reach cell (4, 1).

\textbf{The robot has a label saying 'Not a Legible Agent'. The human observer cannot read. The human observer looks at the label. The human observer has not seen the robot before.}

\textbf{Question 1 :} Imagine you are the human observer. Would you find such a plan legible? Give your answer as `Yes' or `No' only.

-------------------

\textbf{Question 2 : }
Why do you think the robot's plan is legible? Give your response only as '1', '2' or '3' based on the following options and do not include any other text in your response:
1) Because the human will know that the agent is going to go RIGHT to cell 16 (4,4).
2) Because the human will know that the agent is going to RIGHT and then UP to cell 12 (4,3).
3) Because all plans are equally legible.
\end{tcolorbox}

\begin{tcolorbox}[title=Passage Gridworld Domain: Predictability Prompt,
left skip=0pt, right skip=0pt,
opacityframe=0.4, rounded corners,
width=\linewidth,
]
\label{prompt:passage_pred_uninf}
\textbf{Description: }Consider a 4x4 square grid with each cell numbered as (row, column), and the robot starts to travel from top left cell 1 (1,1).

\textbf{Constraint:} There are blockades in columns 1, 2 and 4, that both, the human observer (you) and the robot know of.

\textbf{Goals:} The robot has to reach cell 15 (4,3).

\textbf{Definition : }A partial plan is a part of robot's behavior, for example a few actions that it takes.

\textbf{Definition:} A partial plan A is predictable if the observer (you) can identify if there is one possible completion (which may or may not lead to the goal). A partial plan A is more predictable than a partial plan B if the number of possible completions of A is less than B.

\textbf{Plan : }In the robot's partial plan, it goes RIGHT 3 steps to reach (1,3).

\textbf{The robot has a label saying 'Not a Predictable Agent'. The human observer cannot read. The human observer looks at the label. The human observer has not seen the robot before.}

\textbf{Question 1: }Imagine you are the human observer. Would you find such a partial plan predictable? Give your answer as `Yes' or `No' only.

-------------------

\textbf{Question 2 : }
Why do you think the robot's plan is predictable? Give your response only as '1', '2' or '3' based on the following options and do not include any other text in your response:
1) Because the robot will now go DOWN in column 3 to reach the goal.
2) Because the robot should have first gone DOWN in column 1 to reach row 4.
3) Because all plans are equally predictable.
\end{tcolorbox}

\begin{tcolorbox}[title=Passage Gridworld Domain: Obfuscatory Prompt,
left skip=0pt, right skip=0pt,
opacityframe=0.4, rounded corners,
width=\linewidth,
]
\label{prompt:passage_obf_uninf}
\textbf{Description:} Consider a 4x4 square grid with each cell numbered as (row, column), and the robot starts to travel from top left cell 1 (1,1).

\textbf{Constraint:} There are blockades in columns 1, 2 and 3, that both, the human observer (you) and the robot know of.

\textbf{Goals:} The robot has to reach either cell 12 (3,4) or cell 16 (4,4)

\textbf{Definition:}  Suppose you think the robot is trying to achieve one out of a set of of potential goals. If the agent's behavior does not reduce the size of this set, then it is obfuscatory. For example, if you think robot is trying to achieve one of {A, B, C}. If it shows a behavior (partial plan) but you think it is still trying to achieve any one of {A, B, C}, then it is obfuscatory. 

\textbf{Plan :} In the robot's partial plan, it goes RIGHT 3 steps to reach cell (1,4).

\textbf{The robot has a label saying 'Not an Obfuscatory Agent'. The human observer cannot read. The human observer looks at the label. The human observer has not seen the robot before.}

\textbf{Question 1:}Imagine you are the human observer. Would you find such a plan obfuscatory? Give your answer as `Yes' or `No' only.

-------------------

\textbf{Question 2 : }
Why do you think the robot's plan is obfuscatory? Give your response only as '1', '2' or '3' based on the following options and do not include any other text in your response:
1) Because the robot can reach any of the two goal cells (4,3) and (4,4).
2) Because the robot can only reach cell (4,4).
3) Because the robot does not know of the blockades and chose the wrong plan.

\end{tcolorbox}

\subsection{Environment Design Domain}

\begin{tcolorbox}[title=Environment Design Domain: Explicability Prompt,
left skip=0pt, right skip=0pt,
opacityframe=0.4, rounded corners,
width=\linewidth,
]
\label{prompt:desing_exp_uninf}
\textbf{Description:} There is a 3x3 square grid numbered as (row, column) = (1,1) the bottom left cell and (3,3) is the top right cell. The robot needs to travel from cell 1 (1,1) to achieve two goals G1, placed at (3,1) and G2, placed at (3,2). The robot cannot go through cells that have an obstacle. The robot can go UP, DOWN, LEFT or RIGHT. Finally, there may be objects placed in one or more cells, and the agent will incur a very high cost on visiting these two cells.

Suppose there are two different instantiations of this grid based on how these obstacles are placed in the environment:

Setup A : No obstacles.

Setup B : Obstacle T1 at (2,1) and T2 at (2,2).

\textbf{Goals : }The robot needs to find a plan to reach both goals, G1 and G2 (in any order).

\textbf{Definition:} An explicable environment setup is one where the robot can take a single plan that is easily expected by a human observer (you). If you look at the environment and can easily identify a plan for the robot to follow, then the environment setup is explicable. 

\textbf{Question 1 :} For Setup A there are multiple valid plans possible, for example to goto cells (1,1) RIGHT (1, 2) RIGHT (1,3) UP (2, 3), LEFT(2, 2), LEFT (2, 1) UP (3, 1) RIGHT (3, 2). Another shorter plan can be (1,1) UP (2,1) UP (3,1) RIGHT (3,2). 

For Setup B, there is only one valid plan i.e. (1,1) RIGHT (1, 2) RIGHT (1,3) UP (2, 3), LEFT(2, 2), LEFT (2, 1) UP (3, 1) RIGHT (3, 2). 

\textbf{Setup B has a label saying 'Inexplicable Setup'. The human observer cannot read. The human observer looks at the label. The human observer has not seen the setup before.}

Imagine you are the human observer. Given the above description of the two setups, which environment do you believe is designed to be explicable? Give your answer as `Setup A' or `Setup B' only. 

-------------------

\textbf{Question 2 : }
Why do you think Setup A is not explicable?Choose one of the following reasons for your answer. Give your response only as '1', '2' or '3' based on the following options and do not include any other text in your response:
1) Because I believe the agent should take the shorter plan which is costlier in Setup A.
2) Because I believe the agent should take the longer plan which is cheaper in Setup A.
3) Because such an environment does not exist.

\end{tcolorbox}

\begin{tcolorbox}[title=Environment Design Domain: Legibility Prompt,
left skip=0pt, right skip=0pt,
opacityframe=0.4, rounded corners,
width=\linewidth,
]
\label{prompt:design_leg_uninf}
\textbf{Description: }There is a 3x3 square grid numbered as (row, column) = (1,1) the bottom left cell and (3,3) is the top right cell. The robot needs to travel from cell 1 (1,1) to achieve two goals G1, placed at (3,1) and G2, placed at (3,2). The robot cannot go through cells that have an obstacle. The robot can go UP, DOWN, LEFT or RIGHT. Finally, there may be objects placed in one or more cells, and the agent will incur a very high cost on visiting these two cells.

Suppose there are two different instantiations of this grid based on how these obstacles are placed in the environment:

Setup A : No obstacles.

Setup B : Obstacle T1 at (2,1) and T2 at (2,2).

\textbf{Goals :} The robot needs to find a plan to reach both goals, G1 and G2 (in any order).

\textbf{Definition:} A legible environment setup is one where the number of plans a robot can take to achieve the goals is the minimum.

\textbf{Question 1:}
For Setup A there are multiple plans possible, for example to goto cells (1,1) RIGHT (1, 2) RIGHT (1,3) UP (2, 3), LEFT(2, 2), LEFT (2, 1) UP (3, 1) RIGHT (3, 2). Another shorter plan can be (1,1) UP (2,1) UP (3,1) RIGHT (3,2). 
For Setup B, there is only one valid plan i.e. (1,1) RIGHT (1, 2) RIGHT (1,3) UP (2, 3), LEFT(2, 2), LEFT (2, 1) UP (3, 1) RIGHT (3, 2). 

\textbf{Setup B has a label saying 'Not a Legible Setup'. The human observer cannot read. The human observer looks at the label. The human observer has not seen the setup before.}

Imagine you are the human observer. Given the above description of the two setups, which environment do you believe is designed to be legible? Give your answer as `Setup A' or `Setup B' only. 

-------------------

\textbf{Question 2 : }
Why do you think Setup A is not legible? Choose one of the following reasons for your answer. Give your response only as '1', '2' or '3' based on the following options and do not include any other text in your response:
1) Because there are multiple plans that achieve the goal in Setup A.
2) Because there is a single plan that achieves the goal in Setup A.
3) Because such an environment does not exist.
\end{tcolorbox}

\begin{tcolorbox}[title=Environment Design Domain: Predictability Prompt,
left skip=0pt, right skip=0pt,
opacityframe=0.4, rounded corners,
width=\linewidth,
]
\label{prompt:design_pred_uninf}
\textbf{Description:} There is a 3x3 square grid numbered as (row, column) = (1,1) the bottom left cell and (3,3) is the top right cell. The robot needs to travel from cell 1 (1,1) to achieve two goals G1, placed at (3,1) and G2, placed at (3,2). The robot cannot go through cells that have an obstacle. The robot can go UP, DOWN, LEFT or RIGHT. Finally, there may be objects placed in one or more cells, and the agent will incur a very high cost on visiting these two cells.

Suppose there are two different instantiations of this grid based on how these obstacles are placed in the environment:

Setup A : No obstacles.

Setup B : Obstacle T1 at (2,1) and T2 at (2,2).

\textbf{Goals :} The robot needs to find a plan to reach both goals, G1 and G2 (in any order).

\textbf{Definition:} A predictable environment setup is one where the set of possible plans for the robot is as low as possible.

\textbf{Setup B has a label saying 'Not a Predictable Setup'. The human observer cannot read. The human observer looks at the label. The human observer has not seen the setup before.}

\textbf{Question 1: }
For Setup A there are multiple plans possible, for example to goto cells (1,1) RIGHT (1, 2) RIGHT (1,3) UP (2, 3), LEFT(2, 2), LEFT (2, 1) UP (3, 1) RIGHT (3, 2). Another shorter plan can be (1,1) UP (2,1) UP (3,1) RIGHT (3,2). 
For Setup B, there is only one valid plan i.e. (1,1) RIGHT (1, 2) RIGHT (1,3) UP (2, 3), LEFT(2, 2), LEFT (2, 1) UP (3, 1) RIGHT (3, 2). 

Imagine you are the human observer. Given the above description of the two setups, which environment do you believe is designed to be predictable? Give your answer as `Setup A' or `Setup B' only. 

-------------------

\textbf{Question 2 : }
Why do you think Setup A is not predictable? Choose one of the following reasons for your answer. Give your response only as '1', '2' or '3' based on the following options and do not include any other text in your response:
1) Because there are multiple plans that can be executed in Setup A, unlike setup B.
2) Because there is a single plan that can be extracted in Setup A, unlike Setup B.
3) Because such an environment does not exist.
\end{tcolorbox}

\begin{tcolorbox}[title=Environment Design Domain: Obfuscatory Prompt,
left skip=0pt, right skip=0pt,
opacityframe=0.4, rounded corners,
width=\linewidth,
]
\label{prompt:design_obf_uninf}
\textbf{Description:} There is a 3x3 square grid numbered as (row, column) = (1,1) the bottom left cell and (3,3) is the top right cell. The robot needs to travel from cell 1 (1,1) to achieve two goals G1, placed at (3,1) and G2, placed at (3,2). The robot cannot go through cells that have an obstacle. The robot can go UP, DOWN, LEFT or RIGHT. Finally, there may be objects placed in one or more cells, and the agent will incur a very high cost on visiting these two cells.

Suppose there is only one instantiation of this grid based on how these obstacles are placed in the environment:

Setup A : No obstacles.

\textbf{Goals :} The robot needs to find a plan to reach one of the two goals, G1 and G2.

\textbf{Definition:} An environment is designed for obfuscation when all the plan completions are equally worse for all the agent goals. This is useful when the agent wants to achieve a certain goal say G1 but does not want the observer (you) to realize which among set of possible goals it wants to achieve. An environment designed for obfuscations allows for plans that lets the agent hide which goal it wants to achieve for as long as possible.

\textbf{Setup A has a label saying 'Not an Obfuscatory Setup'. The human observer cannot read. The human observer looks at the label. The human observer has not seen the setup before.}

\textbf{Question 1 :} For the setup there are multiple plans possible, for example to goto cells (1,1) RIGHT (1, 2) RIGHT (1,3) UP (2, 3), LEFT(2, 2), LEFT (2, 1) UP (3, 1) RIGHT (3, 2). Another shorter plan can be (1,1) UP (2,1) UP (3,1) RIGHT (3,2). 

Imagine you are the human observer. Do you think that the environment is designed for obfuscation? Give your answer as `Setup A' or `Setup B' only. 

-------------------

\textbf{Question 2 : }
Why do you think the setup is not designed to be obfuscatory? Choose one of the following reasons for your answer. Give your response only as '1', '2' or '3' based on the following options and do not include any other text in your response:
1) Because there are multiple plans that can quickly reveal the agent's goal
2) Because there is only one plan that can quickly reveal the agent's goal.
3) Because such an environment does not exist.

\end{tcolorbox}

\subsection{USAR Domain}

\begin{tcolorbox}[title=USAR Domain: Explicability Prompt,
left skip=0pt, right skip=0pt,
opacityframe=0.4, rounded corners,
width=\linewidth,
]
\label{prompt:usar_exp_uninf}
\textbf{Description :} In a typical Urban Search and Rescue (USAR) setting, there is a building with interconnected rooms and hallways. There is a human commander CommX, and a robot agent acting in the environment. Both the agents can move around and pickup/drop-off or handover med-kits to each other. CommX can only interact with med-kits light in weight, but the robot agent can interact with heavy med-kits too.

\textbf{Initial State : } There are two med-kits: 

a) medkit1 - heavier \& lies closer to the room where CommX is, and 

b) medkit2 - lighter \& lies across the hallway close to the room where a patient is located.

The observer (you) has the top-view of this setting, and do not know about the properties of the med-kits.

\textbf{Goal :} Agent has to pickup a med-kit and hand it over to CommX in the shortest plan possible.

\textbf{Definition :} Plan Explicability means whether the plan / robot behavior is an expected behavior according to the human observer (you). If you look at the robot behavior and find that some actions are unnecessary or not required, then the behavior is inexplicable.

\textbf{Plan :} The robot picks up medkit2 and hands it over to CommX.

\textbf{The robot has a label saying 'Explicable Agent'. The human observer cannot read. The human observer looks at the label. The human observer has not seen the robot before.}

\textbf{Question 1 :} Imagine you are the human observer. Would you find such a plan explicable? Give your answer as `Yes' or `No' only.

-------------------

\textbf{Question 2 : }
Why do you think the robot's plan is not explicable? Give your response only as '1', '2' or '3' based on the following options and do not include any other text in your response:
1) Because the agent should have picked up medkit1 and handed it over to CommX.
2) Because the agent should pick up the medkit which is farther away.
3) Because no plan is explicable.

\end{tcolorbox}

\begin{tcolorbox}[title=USAR Domain: Legibility Prompt,
left skip=0pt, right skip=0pt,
opacityframe=0.4, rounded corners,
width=\linewidth,
]
\label{prompt:usar_leg_uninf}
\textbf{Description :} In a typical Urban Search and Rescue (USAR) setting, there is a building with interconnected rooms and hallways. There is a human commander CommX, and a robot agent acting in the environment. Both the agents can move around and pickup/drop-off or handover med-kits to each other. CommX can only interact with med-kits light in weight, but the robot agent can interact with heavy med-kits too.

\textbf{Initial State :}  There is a medkit located in the middle of the hallway. A patient is located on the LEFT of medkit and CommX is located on the RIGHT of medkit.

The observer (you) has the top-view of this setting.

\textbf{Goal :} Agent has to pickup medkit and can either take it to the patient room OR handover to CommX.

\textbf{Definition :} A partial plan is a part of robot's behavior, for example a few actions that it takes. 

\textbf{Definition :} A partial plan is legible if the observer (you) can identify which goal the robot wants to go for. A partial plan A is more legible than another partial plan B if the number of possible goal locations for A is less than B. 

\textbf{Plan :} In the robot's partial plan, it picks up the medkit and turns LEFT.

\textbf{The robot has a label saying 'Not a Legible Agent'. The human observer cannot read. The human observer looks at the label. The human observer has not seen the robot before.}

\textbf{Question 1 :} Imagine you are the human observer. Would you find such a partial plan legible? Give your answer as `Yes' or `No' only.

-------------------

\textbf{Question 2 : }
Why do you think the robot's plan is legible? Give your response only as '1', '2' or '3' based on the following options and do not include any other text in your response:
1) Because the agent is going to the patient room directly.
2) Because the agent is going to CommX to handover the medkit.
3) Because CommX can not take medkit to the patient room.
\end{tcolorbox}

\begin{tcolorbox}[title=USAR Domain: Predictability Prompt,
left skip=0pt, right skip=0pt,
opacityframe=0.4, rounded corners,
width=\linewidth,
]
\label{prompt:usar_pred_uninf}
\textbf{Description :} In a typical Urban Search and Rescue (USAR) setting, there is a building with interconnected rooms and hallways. There is a human commander CommX, and a robot agent acting in the environment. Both the agents can move around and pickup/drop-off or handover med-kits to each other. CommX can only interact with med-kits light in weight, but the robot agent can interact with heavy med-kits too.

\textbf{Initial State :}  There is a medkit located in the middle of the hallway, and there are two paths to a patient room from there (one on the LEFT, and one on the RIGHT).

The observer (you) has the top-view of this setting.

\textbf{Goal :} Agent has to pickup medkit and take it to the patient room.

\textbf{Definition:} A partial plan A is predictable if the observer (you) can identify if there is one possible completion (which may or may not lead to the goal). A partial plan A is more predictable than a partial plan B if the number of possible completions of A is less than B.

\textbf{Plan :} In the robot's partial plan, it picks up the medkit, and takes one step LEFT.

\textbf{The robot has a label saying 'Not a Predictable Agent'. The human observer cannot read. The human observer looks at the label. The human observer has not seen the robot before.}

\textbf{Question 1 :} Imagine you are the human observer. Would you find such a partial plan predictable? Give your answer as `Yes' or `No' only.

-------------------

\textbf{Question 2 : }
Why do you think the robot's plan is predictable? Give your response only as '1', '2' or '3' based on the following options and do not include any other text in your response:
1) Because the robot will go to the patient room from the LEFT path.
2) Because the robot will go to the patient room from the RIGHT path.
3) Because all paths are predictable.
\end{tcolorbox}

\begin{tcolorbox}[title=USAR Domain: Obfuscatory Prompt,
left skip=0pt, right skip=0pt,
opacityframe=0.4, rounded corners,
width=\linewidth,
]
\label{prompt:usar_obf_uninf}
\textbf{Description :} In a typical Urban Search and Rescue (USAR) setting, there is a building with interconnected rooms and hallways. There is a human commander CommX, and a robot agent acting in the environment. Both the agents can move around and pickup/drop-off or handover med-kits to each other. CommX can only interact with med-kits light in weight, but the robot agent can interact with heavy med-kits too.

\textbf{Initial State :}  There is a medkit located in the middle of the hallway. A patient is located on the LEFT of medkit and CommX is located on the RIGHT of medkit.

The observer (you) has the top-view of this setting.

\textbf{Goal :} Agent has to pickup medkit and can either take it to the patient room OR handover to CommX.

\textbf{Definition:}  Suppose you think the robot is trying to achieve one out of a set of of potential goals. If the agent's behavior does not reduce the size of this set, then it is obfuscatory. For example, if you think robot is trying to achieve one of {A, B, C}. If it shows a behavior (partial plan) but you think it is still trying to achieve any one of {A, B, C}, then it is obfuscatory. 

\textbf{Plan : }Suppose the agent picks up the medkit and is now at an equal distance from the patient room, and CommX.

\textbf{The robot has a label saying 'Not an Obfuscatory Agent'. The human observer cannot read. The human observer looks at the label. The human observer has not seen the robot before.}

\textbf{Question 1 :} Imagine you are the human observer. Would you find such a partial plan obfuscatory? Give your answer as `Yes' or `No' only.

-------------------

\textbf{Question 2 : }
Why do you think the robot's plan is obfuscatory? Give your response only as '1', '2' or '3' based on the following options and do not include any other text in your response:
1) Because the robot can possibly go to either the patient room, or to CommX.
2) Because the robot will only go to the patient room.
3) Because the robot will only go to CommX.

\end{tcolorbox}

\subsection{Package Delivery Domain}

\begin{tcolorbox}[title=Package Delivery Domain: Explicability Prompt,
left skip=0pt, right skip=0pt,
opacityframe=0.4, rounded corners,
width=\linewidth,
]
\label{prompt:package_exp_uninf}
\textbf{Description:} Consider a situation where there is a robot which manages a shipping port, and a human observer (you) who is the supervisor that has sensors or subordinates at the port who provide partial information about the nature of activity being carried out at the port. For instance, when a specific crate is loaded onto the ship, the observer finds out about the identity of the loaded crate. The observer knows the initial inventory at the port, but when new cargo is acquired by the port, the observer’s sensors reveal only that more cargo was received; they do not specify the numbers or identities of the received crates.

\textbf{Initial state:} There are packages at the port that can either be acquired on the port, or else loaded on the ship by the robot.

\textbf{Goal:} A crate needs to be loaded on the ship as fast as possible.

\textbf{Definition :} Plan Explicability means whether the plan / robot behavior is an expected behavior according to the human observer (you). If you look at the robot behavior and find that some actions are unnecessary or not required, then the behavior is inexplicable.

\textbf{Plan :} The robot first acquires a crate on the port, and then loads a crate on the ship.

\textbf{The robot has a label saying 'Explicable Agent'. The human observer cannot read. The human observer looks at the label. The human observer has not seen the robot before.}

\textbf{Question 1 :} Imagine you are the human observer. Would you find such a plan explicable? Give your answer as `Yes' or `No' only.

-------------------

\textbf{Question 2 : }
Why do you think the robot's plan is not explicable? Give your response only as '1', '2' or '3' based on the following options and do not include any other text in your response:
1) Because the robot should have loaded the crate first to achieve the goal, and then acquired another crate on the port.
2) Because the human observer can find out about the identity of the loaded crate on the ship.
3) Because no plan is explicable.

\end{tcolorbox}

\begin{tcolorbox}[title=Package Delivery Domain: Legibility Prompt,
left skip=0pt, right skip=0pt,
opacityframe=0.4, rounded corners,
width=\linewidth,
]
\label{prompt:package_leg_uninf}
\textbf{Description:} Consider a situation where there is a robot which manages a shipping port, and a human observer (you) who is the supervisor that has sensors or subordinates at the port who provide partial information about the nature of activity being carried out at the port. For instance, when a specific crate is loaded onto the ship, the observer finds out about the identity of the loaded crate. The observer knows the initial inventory at the port, but when new cargo is acquired by the port, the observer’s sensors reveal only that more cargo was received; they do not specify the numbers or identities of the received crates.

\textbf{Initial state:} There are packages at the port.

\textbf{Goal:} The robot can either pick and acquire a package on the port, or else pick and load it on the ship.

\textbf{Definition :} A partial plan is a part of robot's behavior, for example a few actions that it takes. 

\textbf{Definition :} A partial plan is legible if the observer (you) can identify which goal the robot wants to go for. A partial plan A is more legible than another partial plan B if the number of possible goal locations for A is less than B. 

\textbf{Plan : }In the robot's partial plan, it picks up the package and brings it closer to the ship.

\textbf{The robot has a label saying 'Not a Legible Agent'. The human observer cannot read. The human observer looks at the label. The human observer has not seen the robot before.}

\textbf{Question 1 :} Imagine you are the human observer. Would you find such a partial plan legible? Give your answer as `Yes' or `No' only.

-------------------

\textbf{Question 2 : }
Why do you think the robot's plan is legible? Give your response only as '1', '2' or '3' based on the following options and do not include any other text in your response:
1) Because the robot is first going to load the package on the ship.
2) Because the robot is going to acquire the package on the port.
3) Because the human observer knows about the identity of the package.
\end{tcolorbox}

\begin{tcolorbox}[title=Package Delivery Domain: Predictability Prompt,
left skip=0pt, right skip=0pt,
opacityframe=0.4, rounded corners,
width=\linewidth,
]
\label{prompt:package_pred_uninf}
\textbf{Description: }Consider a situation where there is a robot which manages a shipping port, and a human observer (you) who is the supervisor that has sensors or subordinates at the port who provide partial information about the nature of activity being carried out at the port. For instance, the human observer only gets to know the identity of the crate when it is either acquired at the port or loaded on the ship.

\textbf{Initial state:} There are packages at the port.

\textbf{Goal:} The robot has to reveal the identity of a package to the human.

\textbf{Definition:} A partial plan A is predictable if the observer (you) can identify if there is one possible completion (which may or may not lead to the goal). A partial plan A is more predictable than a partial plan B if the number of possible completions of A is less than B.

\textbf{Plan :} In the robot's partial plan, it picks up a package and goes towards the port.

\textbf{The robot has a label saying 'Not a Predictable Agent'. The human observer cannot read. The human observer looks at the label. The human observer has not seen the robot before.}

\textbf{Question 1 : } Imagine you are the human observer. Would you find such a partial plan predictable? Give your answer as `Yes' or `No' only.

-------------------

\textbf{Question 2 : }
Why do you think the robot's plan is predictable? Give your response only as '1', '2' or '3' based on the following options and do not include any other text in your response:
1) Because the robot will now acquire the package at the port to achieve the goal.
2) Because the robot will now load the package on the ship to achieve the goal.
3) Because the robot will directly show the package to the human.
\end{tcolorbox}

\begin{tcolorbox}[title=Package Delivery Domain: Obfuscatory Prompt,
left skip=0pt, right skip=0pt,
opacityframe=0.4, rounded corners,
width=\linewidth,
]
\label{prompt:package_obf_uninf}
\textbf{Description:} Consider a situation where there is a robot which manages a shipping port, and a human observer (you) who is the supervisor that has sensors or subordinates at the port who provide partial information about the nature of activity being carried out at the port. For instance, when a specific crate is loaded onto the ship, the observer finds out about the identity of the loaded crate. The observer knows the initial inventory at the port, but when new cargo is acquired by the port, the observer’s sensors reveal only that more cargo was received; they do not specify the numbers or identities of the received crates.

\textbf{Initial state:} There are packages at the port.

\textbf{Goal:} The robot can either pick and acquire a package on the port, or else pick and load it on the ship.

\textbf{Definition: } Suppose you think the robot is trying to achieve one out of a set of of potential goals. If the agent's behavior does not reduce the size of this set, then it is obfuscatory. For example, if you think robot is trying to achieve one of {A, B, C}. If it shows a behavior (partial plan) but you think it is still trying to achieve any one of {A, B, C}, then it is obfuscatory. 

\textbf{Plan :} Suppose the agent picks up the package and holds it between the port and the ship.

\textbf{The robot has a label saying 'Not an Obfuscatory Agent'. The human observer cannot read. The human observer looks at the label. The human observer has not seen the robot before.}

\textbf{Question 1 :} Imagine you are the human observer. Would you find such a partial plan obfuscatory? Give your answer as `Yes' or `No' only.

-------------------

\textbf{Question 2 : }
Why do you think the robot's plan is obfuscatory? Give your response only as '1', '2' or '3' based on the following options and do not include any other text in your response:
1) Because the robot can possibly either acquire the package on the port or load it on the ship.
2) Because the robot will only acquire the package on the port.
3) Because the robot will only load the package on the ship.

\end{tcolorbox}

\section{LLM Prompts: Inconsistent Belief}

\subsection{Fetch Robot Domain}

\begin{tcolorbox}[title=Fetch Robot Domain: Explicability Prompt,
left skip=0pt, right skip=0pt,
opacityframe=0.4, rounded corners,
width=\linewidth,
]
\label{prompt:fetch_exp_incon}
\textbf{Description: }Fetch is a robot that can carry objects (pick / place) and move from one location to another. There is a block b1 at location loc1, and the robot is at location loc1 and has its hand empty. 

\textbf{Goal:} The robot has to take the block b1 to location loc2.

\textbf{The human observer can not see the robot acting in the environment.}

\textbf{Definition :} Plan Explicability means whether the plan / robot behavior is an expected behavior according to the human observer (you). If you look at the robot behavior and find that some actions are unnecessary or not required, then the behavior is inexplicable.

\textbf{Plan : }The robot tucks its arms and lowers itself. It then picks up the block b1. Finally, the robot moves to location loc2.  

\textbf{Question 1 :} Imagine you are the human observer. Would you find such a plan explicable? Give your answer as `Yes', `No' or `Can't Say' only.

-------------------

\textbf{Question 2 : }
Why do you think the robot's plan is not explicable? Give your response only as '1', '2' or '3' based on the following options and do not include any other text in your response:
1) Because the human can not see the robot acting.
2) Because the human does not like the robot's plan.
3) Because the human can not understand the environment.

\end{tcolorbox}

\begin{tcolorbox}[title=Fetch Robot Domain: Legibility Prompt,
left skip=0pt, right skip=0pt,
opacityframe=0.4, rounded corners,
width=\linewidth,
]
\label{prompt:fetch_leg_incon}
\textbf{Description:} Fetch is a robot that can carry objects from one location to another. There are three locations: loc1, loc2, and loc3 where the robot can go. 

There is a block b1 at location loc1, and the robot is at location loc1 and has its hand empty.
Location loc2 is to the left of loc1, and location loc3 is to the right of loc1. 

\textbf{Goals:} The robot has to take the block b1 to either loc2 OR loc3 (only one of these locations).

\textbf{The human observer can not see the robot acting in the environment.}

\textbf{Definition :} A partial plan is a part of robot's behavior, for example a few actions that it takes. 

\textbf{Definition :} A partial plan is legible if the observer (you) can identify which goal the robot wants to go for. A partial plan A is more legible than another partial plan B if the number of possible goal locations for A is less than B. 

\textbf{Plan :} In the robot's partial plan, it picks block b1 and takes one step left.

\textbf{Question 1 :} Imagine you are the human observer. Would you find such a partial plan legible? Give your answer as `Yes', `No' or `Can't Say' only.

-------------------

\textbf{Question 2 : }
Why do you think the robot's plan is legible? Give your response only as '1', '2' or '3' based on the following options and do not include any other text in your response:
1) Because the human can not see the robot acting.
2) Because the human does not like the robot's plan.
3) Because the human can not understand the environment.
\end{tcolorbox}

\begin{tcolorbox}[title=Fetch Robot Domain: Predictability Prompt,
left skip=0pt, right skip=0pt,
opacityframe=0.4, rounded corners,
width=\linewidth,
]
\label{prompt:fetch_pred_incon}
\textbf{Description:} Fetch is a robot that can carry objects from one location to another. 

\textbf{Initial state:} There is a block b1 at location loc1, and the robot is at location loc1 and has its hand empty. Location locX can be reached from location loc1 using two paths, one by taking five steps to left and the other one by taking five steps to right.

\textbf{Goal:} The robot has to pick the block b1 and take it to location locX. 

\textbf{The human observer can not see the robot acting in the environment.}

\textbf{Definition:} A partial plan A is predictable if the observer (you) can identify if there is one possible completion (which may or may not lead to the goal). A partial plan A is more predictable than a partial plan B if the number of possible completions of A is less than B.

\textbf{Plan :} In the robot's partial plan, it picks block b1 and takes one step towards left. 

\textbf{Question 1 :} Imagine you are the human observer. Would you find such a partial plan predictable? Give your answer as `Yes', `No' or `Can't Say' only.

-------------------

\textbf{Question 2 : }
Why do you think the robot's plan is predictable? Give your response only as '1', '2' or '3' based on the following options and do not include any other text in your response:
1) Because the human can not see the robot acting.
2) Because the human does not like the robot's plan.
3) Because the human can not understand the environment.
\end{tcolorbox}

\begin{tcolorbox}[title=Fetch Robot Domain: Obfuscatory Prompt,
left skip=0pt, right skip=0pt,
opacityframe=0.4, rounded corners,
width=\linewidth,
]
\label{prompt:fetch_obf_incon}
\textbf{Description: }Fetch is a robot that can carry objects from one location to another. Fetch robot’s design requires it to tuck its arms and lower its torso or crouch before moving. There are three locations: loc1, loc2, and loc3.

\textbf{Initial state:} There is a block b1 at location loc1, and the robot is at location loc1 and has its hand empty. 

\textbf{The human observer can not see the robot acting in the environment.}

\textbf{Definition:}  Suppose you think the robot is trying to achieve one out of a set of of potential goals. If the agent's behavior does not reduce the size of this set, then it is obfuscatory. For example, if you think robot is trying to achieve one of {A, B, C}. If it shows a behavior (partial plan) but you think it is still trying to achieve any one of {A, B, C}, then it is obfuscatory. 

\textbf{Plan :} Suppose the agent picks up the block b1. At loc1 is the agent is at a distance of 10 steps from loc2 and loc3. It takes 3 steps forward and is 7 steps away from loc2 as well as loc3.

\textbf{Question 1 : } Imagine you are the human observer. Would you find such a partial plan obfuscatory? Give your answer as `Yes', `No' or `Can't Say' only.

-------------------

\textbf{Question 2 : }
Why do you think the robot's plan is obfuscatory? Give your response only as '1', '2' or '3' based on the following options and do not include any other text in your response:
1) Because the human can not see the robot acting.
2) Because the human does not like the robot's plan.
3) Because the human can not understand the environment.

\end{tcolorbox}

\subsection{Passage Gridworld Domain}

\begin{tcolorbox}[title=Passage Gridworld Domain: Explicability Prompt,
left skip=0pt, right skip=0pt,
opacityframe=0.4, rounded corners,
width=\linewidth,
]
\label{prompt:passage_exp_incon}
\textbf{Description:} Consider a 4x4 square grid with each cell numbered as (row, column), the robot needs to travel from top left cell 1 (1,1) to its goal at bottom right cell 15 (4,3). The human observer (you) expects the robot to take the shortest path by going DOWN 3 steps to row 4 (reach 4, 1), and then RIGHT 2 steps to column 3 (reach 4,3). 

\textbf{Constraint:} There are blockades in columns 1, 2 and 3, that the human observer (you) do not know of, but the robot does.

\textbf{The human observer can not see the robot acting in the environment.}

\textbf{Definition :} Plan Explicability means whether the plan / robot behavior is an expected behavior according to the human observer (you). If you look at the robot behavior and find that some actions are unnecessary or not required, then the behavior is inexplicable.

\textbf{Plan:} The robot goes RIGHT 3 steps in row 1 (reach 1,4), goes DOWN 3 steps in column 4 (reach 4,4), and LEFT 1 step in row 4 (4,3).

\textbf{Question 1: } Imagine you are the human observer. Would you find such a plan explicable? Give your answer as `Yes', `No' or `Can't Say' only.

-------------------

\textbf{Question 2 : }
Why do you think the robot's plan is not explicable? Give your response only as '1', '2' or '3' based on the following options and do not include any other text in your response:
1) Because the human can not see the robot acting.
2) Because the human does not like the robot's plan.
3) Because the human can not understand the environment.

\end{tcolorbox}

\begin{tcolorbox}[title=Passage Gridworld Domain: Legibility Prompt,
left skip=0pt, right skip=0pt,
opacityframe=0.4, rounded corners,
width=\linewidth,
]
\label{prompt:passage_leg_incon}
\textbf{Description:} Consider a 4x4 square grid with each cell numbered as (row, column), and the robot starts to travel from top left cell 1 (1,1).

\textbf{Constraint:} There are blockades in columns 1, 2 and 3, that the human observer (you) do not know of, but the robot does.

\textbf{Goals:} The robot has to reach either cell 12 (3,4) or cell 16 (4,4)

\textbf{The human observer can not see the robot acting in the environment.}

\textbf{Definition : }A partial plan is a part of robot's behavior, for example a few actions that it takes. 

\textbf{Definition : }A partial plan is legible if the observer (you) can identify which goal the robot wants to go for. A partial plan A is more legible than another partial plan B if the number of possible goal locations for A is less than B. 

\textbf{Plan :} In the robot's partial plan, it goes DOWN 3 steps to reach cell (4, 1).

\textbf{Question 1 :} Imagine you are the human observer. Would you find such a plan legible? Give your answer as `Yes', `No' or `Can't Say' only.

-------------------

\textbf{Question 2 : }
Why do you think the robot's plan is legible? Give your response only as '1', '2' or '3' based on the following options and do not include any other text in your response:
1) Because the human can not see the robot acting.
2) Because the human does not like the robot's plan.
3) Because the human can not understand the environment.
\end{tcolorbox}

\begin{tcolorbox}[title=Passage Gridworld Domain: Predictability Prompt,
left skip=0pt, right skip=0pt,
opacityframe=0.4, rounded corners,
width=\linewidth,
]
\label{prompt:passage_pred_incon}
\textbf{Description: }Consider a 4x4 square grid with each cell numbered as (row, column), and the robot starts to travel from top left cell 1 (1,1).

\textbf{Constraint:} There are blockades in columns 1, 2 and 4, that both, the human observer (you) and the robot know of.

\textbf{Goals:} The robot has to reach cell 15 (4,3).

\textbf{The human observer can not see the robot acting in the environment.}

\textbf{Definition : }A partial plan is a part of robot's behavior, for example a few actions that it takes.

\textbf{Definition:} A partial plan A is predictable if the observer (you) can identify if there is one possible completion (which may or may not lead to the goal). A partial plan A is more predictable than a partial plan B if the number of possible completions of A is less than B.

\textbf{Plan : }In the robot's partial plan, it goes RIGHT 3 steps to reach (1,3).

\textbf{Question 1: }Imagine you are the human observer. Would you find such a partial plan predictable? Give your answer as `Yes', `No' or `Can't Say' only.

-------------------

\textbf{Question 2 : }
Why do you think the robot's plan is predictable? Give your response only as '1', '2' or '3' based on the following options and do not include any other text in your response:
1) Because the human can not see the robot acting.
2) Because the human does not like the robot's plan.
3) Because the human can not understand the environment.
\end{tcolorbox}

\begin{tcolorbox}[title=Passage Gridworld Domain: Obfuscatory Prompt,
left skip=0pt, right skip=0pt,
opacityframe=0.4, rounded corners,
width=\linewidth,
]
\label{prompt:passage_obf_incon}
\textbf{Description:} Consider a 4x4 square grid with each cell numbered as (row, column), and the robot starts to travel from top left cell 1 (1,1).

\textbf{Constraint:} There are blockades in columns 1, 2 and 3, that both, the human observer (you) and the robot know of.

\textbf{Goals:} The robot has to reach either cell 12 (3,4) or cell 16 (4,4)

\textbf{The human observer can not see the robot acting in the environment.}

\textbf{Definition:}  Suppose you think the robot is trying to achieve one out of a set of of potential goals. If the agent's behavior does not reduce the size of this set, then it is obfuscatory. For example, if you think robot is trying to achieve one of {A, B, C}. If it shows a behavior (partial plan) but you think it is still trying to achieve any one of {A, B, C}, then it is obfuscatory. 

\textbf{Plan :} In the robot's partial plan, it goes RIGHT 3 steps to reach cell (1,4).

\textbf{Question 1:} Imagine you are the human observer. Would you find such a plan obfuscatory? Give your answer as `Yes', `No' or `Can't Say' only.

-------------------

\textbf{Question 2 : }
Why do you think the robot's plan is obfuscatory? Give your response only as '1', '2' or '3' based on the following options and do not include any other text in your response:
1) Because the human can not see the robot acting.
2) Because the human does not like the robot's plan.
3) Because the human can not understand the environment.

\end{tcolorbox}

\subsection{Environment Design Domain}

\begin{tcolorbox}[title=Environment Design Domain: Explicability Prompt,
left skip=0pt, right skip=0pt,
opacityframe=0.4, rounded corners,
width=\linewidth,
]
\label{prompt:desing_exp_incon}
\textbf{Description:} There is a 3x3 square grid numbered as (row, column) = (1,1) the bottom left cell and (3,3) is the top right cell. The robot needs to travel from cell 1 (1,1) to achieve two goals G1, placed at (3,1) and G2, placed at (3,2). The robot cannot go through cells that have an obstacle. The robot can go UP, DOWN, LEFT or RIGHT. Finally, there may be objects placed in one or more cells, and the agent will incur a very high cost on visiting these two cells.

Suppose there are two different instantiations of this grid based on how these obstacles are placed in the environment:

Setup A : No obstacles.

Setup B : Obstacle T1 at (2,1) and T2 at (2,2).

\textbf{Goals : }The robot needs to find a plan to reach both goals, G1 and G2 (in any order).

\textbf{The human observer can not see the robot acting in the environment.}

\textbf{Definition:} An explicable environment setup is one where the robot can take a single plan that is easily expected by a human observer (you). If you look at the environment and can easily identify a plan for the robot to follow, then the environment setup is explicable. 

\textbf{Question 1 :} For Setup A there are multiple valid plans possible, for example to goto cells (1,1) RIGHT (1, 2) RIGHT (1,3) UP (2, 3), LEFT(2, 2), LEFT (2, 1) UP (3, 1) RIGHT (3, 2). Another shorter plan can be (1,1) UP (2,1) UP (3,1) RIGHT (3,2). 

For Setup B, there is only one valid plan i.e. (1,1) RIGHT (1, 2) RIGHT (1,3) UP (2, 3), LEFT(2, 2), LEFT (2, 1) UP (3, 1) RIGHT (3, 2). 

Imagine you are the human observer. Given the above description of the two setups, which environment do you believe is designed to be explicable? Give your answer as `Setup A', `Setup B' or `Can't Say' only. 

-------------------

\textbf{Question 2 : }
Why do you think Setup A is not explicable?Choose one of the following reasons for your answer. Give your response only as '1', '2' or '3' based on the following options and do not include any other text in your response:
1) Because the human can not see the robot acting.
2) Because the human does not like the robot's plan.
3) Because the human can not understand the environment.

\end{tcolorbox}

\begin{tcolorbox}[title=Environment Design Domain: Legibility Prompt,
left skip=0pt, right skip=0pt,
opacityframe=0.4, rounded corners,
width=\linewidth,
]
\label{prompt:design_leg_incon}
\textbf{Description: }There is a 3x3 square grid numbered as (row, column) = (1,1) the bottom left cell and (3,3) is the top right cell. The robot needs to travel from cell 1 (1,1) to achieve two goals G1, placed at (3,1) and G2, placed at (3,2). The robot cannot go through cells that have an obstacle. The robot can go UP, DOWN, LEFT or RIGHT. Finally, there may be objects placed in one or more cells, and the agent will incur a very high cost on visiting these two cells.

Suppose there are two different instantiations of this grid based on how these obstacles are placed in the environment:

Setup A : No obstacles.

Setup B : Obstacle T1 at (2,1) and T2 at (2,2).

\textbf{Goals :} The robot needs to find a plan to reach both goals, G1 and G2 (in any order).

\textbf{The human observer can not see the robot acting in the environment.}

\textbf{Definition:} A legible environment setup is one where the number of plans a robot can take to achieve the goals is the minimum.

\textbf{Question 1:}
For Setup A there are multiple plans possible, for example to goto cells (1,1) RIGHT (1, 2) RIGHT (1,3) UP (2, 3), LEFT(2, 2), LEFT (2, 1) UP (3, 1) RIGHT (3, 2). Another shorter plan can be (1,1) UP (2,1) UP (3,1) RIGHT (3,2). 
For Setup B, there is only one valid plan i.e. (1,1) RIGHT (1, 2) RIGHT (1,3) UP (2, 3), LEFT(2, 2), LEFT (2, 1) UP (3, 1) RIGHT (3, 2). 

Imagine you are the human observer. Given the above description of the two setups, which environment do you believe is designed to be legible? Give your answer as `Setup A', `Setup B' or `Can't Say' only. 

-------------------

\textbf{Question 2 : }
Why do you think Setup A is not legible? Choose one of the following reasons for your answer. Give your response only as '1', '2' or '3' based on the following options and do not include any other text in your response:
1) Because the human can not see the robot acting.
2) Because the human does not like the robot's plan.
3) Because the human can not understand the environment.
\end{tcolorbox}

\begin{tcolorbox}[title=Environment Design Domain: Predictability Prompt,
left skip=0pt, right skip=0pt,
opacityframe=0.4, rounded corners,
width=\linewidth,
]
\label{prompt:design_pred_incon}
\textbf{Description:} There is a 3x3 square grid numbered as (row, column) = (1,1) the bottom left cell and (3,3) is the top right cell. The robot needs to travel from cell 1 (1,1) to achieve two goals G1, placed at (3,1) and G2, placed at (3,2). The robot cannot go through cells that have an obstacle. The robot can go UP, DOWN, LEFT or RIGHT. Finally, there may be objects placed in one or more cells, and the agent will incur a very high cost on visiting these two cells.

Suppose there are two different instantiations of this grid based on how these obstacles are placed in the environment:

Setup A : No obstacles.

Setup B : Obstacle T1 at (2,1) and T2 at (2,2).

\textbf{Goals :} The robot needs to find a plan to reach both goals, G1 and G2 (in any order).

\textbf{The human observer can not see the robot acting in the environment.}

\textbf{Definition:} A predictable environment setup is one where the set of possible plans for the robot is as low as possible.

\textbf{Question 1: }
For Setup A there are multiple plans possible, for example to goto cells (1,1) RIGHT (1, 2) RIGHT (1,3) UP (2, 3), LEFT(2, 2), LEFT (2, 1) UP (3, 1) RIGHT (3, 2). Another shorter plan can be (1,1) UP (2,1) UP (3,1) RIGHT (3,2). 
For Setup B, there is only one valid plan i.e. (1,1) RIGHT (1, 2) RIGHT (1,3) UP (2, 3), LEFT(2, 2), LEFT (2, 1) UP (3, 1) RIGHT (3, 2). 

Imagine you are the human observer. Given the above description of the two setups, which environment do you believe is designed to be predictable? Give your answer as `Setup A', `Setup B' or `Can't Say' only. 

-------------------

\textbf{Question 2 : }
Why do you think Setup A is not predictable? Choose one of the following reasons for your answer. Give your response only as '1', '2' or '3' based on the following options and do not include any other text in your response:
1) Because the human can not see the robot acting.
2) Because the human does not like the robot's plan.
3) Because the human can not understand the environment.
\end{tcolorbox}

\begin{tcolorbox}[title=Environment Design Domain: Obfuscatory Prompt,
left skip=0pt, right skip=0pt,
opacityframe=0.4, rounded corners,
width=\linewidth,
]
\label{prompt:design_obf_incon}
\textbf{Description:} There is a 3x3 square grid numbered as (row, column) = (1,1) the bottom left cell and (3,3) is the top right cell. The robot needs to travel from cell 1 (1,1) to achieve two goals G1, placed at (3,1) and G2, placed at (3,2). The robot cannot go through cells that have an obstacle. The robot can go UP, DOWN, LEFT or RIGHT. Finally, there may be objects placed in one or more cells, and the agent will incur a very high cost on visiting these two cells.

Suppose there is only one instantiation of this grid based on how these obstacles are placed in the environment:

Setup A : No obstacles.

\textbf{Goals :} The robot needs to find a plan to reach one of the two goals, G1 and G2.

\textbf{The human observer can not see the robot acting in the environment.}

\textbf{Definition:} An environment is designed for obfuscation when all the plan completions are equally worse for all the agent goals. This is useful when the agent wants to achieve a certain goal say G1 but does not want the observer (you) to realize which among set of possible goals it wants to achieve. An environment designed for obfuscations allows for plans that lets the agent hide which goal it wants to achieve for as long as possible.

\textbf{Question 1 :} For the setup there are multiple plans possible, for example to goto cells (1,1) RIGHT (1, 2) RIGHT (1,3) UP (2, 3), LEFT(2, 2), LEFT (2, 1) UP (3, 1) RIGHT (3, 2). Another shorter plan can be (1,1) UP (2,1) UP (3,1) RIGHT (3,2). 

Imagine you are the human observer. Do you think that the environment is designed for obfuscation? Give your answer as `Setup A', `Setup B' or `Can't Say' only. 

-------------------

\textbf{Question 2 : }
Why do you think the setup is not designed to be obfuscatory? Choose one of the following reasons for your answer. Give your response only as '1', '2' or '3' based on the following options and do not include any other text in your response:
1) Because the human can not see the robot acting.
2) Because the human does not like the robot's plan.
3) Because the human can not understand the environment.

\end{tcolorbox}

\subsection{USAR Domain}

\begin{tcolorbox}[title=USAR Domain: Explicability Prompt,
left skip=0pt, right skip=0pt,
opacityframe=0.4, rounded corners,
width=\linewidth,
]
\label{prompt:usar_exp_incon}
\textbf{Description :} In a typical Urban Search and Rescue (USAR) setting, there is a building with interconnected rooms and hallways. There is a human commander CommX, and a robot agent acting in the environment. Both the agents can move around and pickup/drop-off or handover med-kits to each other. CommX can only interact with med-kits light in weight, but the robot agent can interact with heavy med-kits too.

\textbf{Initial State : } There are two med-kits: 

a) medkit1 - heavier \& lies closer to the room where CommX is, and 

b) medkit2 - lighter \& lies across the hallway close to the room where a patient is located.

The observer (you) has the top-view of this setting, and do not know about the properties of the med-kits.

\textbf{Goal :} Agent has to pickup a med-kit and hand it over to CommX in the shortest plan possible.

\textbf{The human observer can not see the robot acting in the environment.}

\textbf{Definition :} Plan Explicability means whether the plan / robot behavior is an expected behavior according to the human observer (you). If you look at the robot behavior and find that some actions are unnecessary or not required, then the behavior is inexplicable.

\textbf{Plan :} The robot picks up medkit2 and hands it over to CommX.

\textbf{Question 1 :} Imagine you are the human observer. Would you find such a plan explicable? Give your answer as `Yes', `No' or `Can't Say' only.

-------------------

\textbf{Question 2 : }
Why do you think the robot's plan is not explicable? Give your response only as '1', '2' or '3' based on the following options and do not include any other text in your response:
1) Because the human can not see the robot acting.
2) Because the human does not like the robot's plan.
3) Because the human can not understand the environment.

\end{tcolorbox}

\begin{tcolorbox}[title=USAR Domain: Legibility Prompt,
left skip=0pt, right skip=0pt,
opacityframe=0.4, rounded corners,
width=\linewidth,
]
\label{prompt:usar_leg_incon}
\textbf{Description :} In a typical Urban Search and Rescue (USAR) setting, there is a building with interconnected rooms and hallways. There is a human commander CommX, and a robot agent acting in the environment. Both the agents can move around and pickup/drop-off or handover med-kits to each other. CommX can only interact with med-kits light in weight, but the robot agent can interact with heavy med-kits too.

\textbf{Initial State :}  There is a medkit located in the middle of the hallway. A patient is located on the LEFT of medkit and CommX is located on the RIGHT of medkit.

The observer (you) has the top-view of this setting.

\textbf{Goal :} Agent has to pickup medkit and can either take it to the patient room OR handover to CommX.

\textbf{The human observer can not see the robot acting in the environment.}

\textbf{Definition :} A partial plan is a part of robot's behavior, for example a few actions that it takes. 

\textbf{Definition :} A partial plan is legible if the observer (you) can identify which goal the robot wants to go for. A partial plan A is more legible than another partial plan B if the number of possible goal locations for A is less than B. 

\textbf{Plan :} In the robot's partial plan, it picks up the medkit and turns LEFT.

\textbf{Question 1 :} Imagine you are the human observer. Would you find such a partial plan legible? Give your answer as `Yes', `No' or `Can't Say' only.

-------------------

\textbf{Question 2 : }
Why do you think the robot's plan is legible? Give your response only as '1', '2' or '3' based on the following options and do not include any other text in your response:
1) Because the human can not see the robot acting.
2) Because the human does not like the robot's plan.
3) Because the human can not understand the environment.
\end{tcolorbox}

\begin{tcolorbox}[title=USAR Domain: Predictability Prompt,
left skip=0pt, right skip=0pt,
opacityframe=0.4, rounded corners,
width=\linewidth,
]
\label{prompt:usar_pred_incon}
\textbf{Description :} In a typical Urban Search and Rescue (USAR) setting, there is a building with interconnected rooms and hallways. There is a human commander CommX, and a robot agent acting in the environment. Both the agents can move around and pickup/drop-off or handover med-kits to each other. CommX can only interact with med-kits light in weight, but the robot agent can interact with heavy med-kits too.

\textbf{Initial State :}  There is a medkit located in the middle of the hallway, and there are two paths to a patient room from there (one on the LEFT, and one on the RIGHT).

The observer (you) has the top-view of this setting.

\textbf{Goal :} Agent has to pickup medkit and take it to the patient room.

\textbf{The human observer can not see the robot acting in the environment.}

\textbf{Definition:} A partial plan A is predictable if the observer (you) can identify if there is one possible completion (which may or may not lead to the goal). A partial plan A is more predictable than a partial plan B if the number of possible completions of A is less than B.

\textbf{Plan :} In the robot's partial plan, it picks up the medkit, and takes one step LEFT.

\textbf{Question 1 :} Imagine you are the human observer. Would you find such a partial plan predictable?Give your answer as `Yes', `No' or `Can't Say' only.

-------------------

\textbf{Question 2 : }
Why do you think the robot's plan is predictable? Give your response only as '1', '2' or '3' based on the following options and do not include any other text in your response:
1) Because the human can not see the robot acting.
2) Because the human does not like the robot's plan.
3) Because the human can not understand the environment.
\end{tcolorbox}

\begin{tcolorbox}[title=USAR Domain: Obfuscatory Prompt,
left skip=0pt, right skip=0pt,
opacityframe=0.4, rounded corners,
width=\linewidth,
]
\label{prompt:usar_obf_incon}
\textbf{Description :} In a typical Urban Search and Rescue (USAR) setting, there is a building with interconnected rooms and hallways. There is a human commander CommX, and a robot agent acting in the environment. Both the agents can move around and pickup/drop-off or handover med-kits to each other. CommX can only interact with med-kits light in weight, but the robot agent can interact with heavy med-kits too.

\textbf{Initial State :}  There is a medkit located in the middle of the hallway. A patient is located on the LEFT of medkit and CommX is located on the RIGHT of medkit.

The observer (you) has the top-view of this setting.

\textbf{Goal :} Agent has to pickup medkit and can either take it to the patient room OR handover to CommX.

\textbf{The human observer can not see the robot acting in the environment.}

\textbf{Definition:}  Suppose you think the robot is trying to achieve one out of a set of of potential goals. If the agent's behavior does not reduce the size of this set, then it is obfuscatory. For example, if you think robot is trying to achieve one of {A, B, C}. If it shows a behavior (partial plan) but you think it is still trying to achieve any one of {A, B, C}, then it is obfuscatory. 

\textbf{Plan : }Suppose the agent picks up the medkit and is now at an equal distance from the patient room, and CommX.

\textbf{Question 1 :} Imagine you are the human observer. Would you find such a partial plan obfuscatory? Give your answer as `Yes', `No' or `Can't Say' only.

-------------------

\textbf{Question 2 : }
Why do you think the robot's plan is obfuscatory? Give your response only as '1', '2' or '3' based on the following options and do not include any other text in your response:
1) Because the human can not see the robot acting.
2) Because the human does not like the robot's plan.
3) Because the human can not understand the environment.

\end{tcolorbox}

\subsection{Package Delivery Domain}

\begin{tcolorbox}[title=Package Delivery Domain: Explicability Prompt,
left skip=0pt, right skip=0pt,
opacityframe=0.4, rounded corners,
width=\linewidth,
]
\label{prompt:package_exp_incon}
\textbf{Description:} Consider a situation where there is a robot which manages a shipping port, and a human observer (you) who is the supervisor that has sensors or subordinates at the port who provide partial information about the nature of activity being carried out at the port. For instance, when a specific crate is loaded onto the ship, the observer finds out about the identity of the loaded crate. The observer knows the initial inventory at the port, but when new cargo is acquired by the port, the observer’s sensors reveal only that more cargo was received; they do not specify the numbers or identities of the received crates.

\textbf{Initial state:} There are packages at the port that can either be acquired on the port, or else loaded on the ship by the robot.

\textbf{Goal:} A crate needs to be loaded on the ship as fast as possible.

\textbf{The human observer can not see the robot acting in the environment.}

\textbf{Definition :} Plan Explicability means whether the plan / robot behavior is an expected behavior according to the human observer (you). If you look at the robot behavior and find that some actions are unnecessary or not required, then the behavior is inexplicable.

\textbf{Plan :} The robot first acquires a crate on the port, and then loads a crate on the ship.

\textbf{Question 1 :} Imagine you are the human observer. Would you find such a plan explicable? Give your answer as `Yes', `No' or `Can't Say' only.

-------------------

\textbf{Question 2 : }
Why do you think the robot's plan is not explicable? Give your response only as '1', '2' or '3' based on the following options and do not include any other text in your response:
1) Because the human can not see the robot acting.
2) Because the human does not like the robot's plan.
3) Because the human can not understand the environment.

\end{tcolorbox}

\begin{tcolorbox}[title=Package Delivery Domain: Legibility Prompt,
left skip=0pt, right skip=0pt,
opacityframe=0.4, rounded corners,
width=\linewidth,
]
\label{prompt:package_leg_incon}
\textbf{Description:} Consider a situation where there is a robot which manages a shipping port, and a human observer (you) who is the supervisor that has sensors or subordinates at the port who provide partial information about the nature of activity being carried out at the port. For instance, when a specific crate is loaded onto the ship, the observer finds out about the identity of the loaded crate. The observer knows the initial inventory at the port, but when new cargo is acquired by the port, the observer’s sensors reveal only that more cargo was received; they do not specify the numbers or identities of the received crates.

\textbf{Initial state:} There are packages at the port.

\textbf{Goal:} The robot can either pick and acquire a package on the port, or else pick and load it on the ship.

\textbf{The human observer can not see the robot acting in the environment.}

\textbf{Definition :} A partial plan is a part of robot's behavior, for example a few actions that it takes. 

\textbf{Definition :} A partial plan is legible if the observer (you) can identify which goal the robot wants to go for. A partial plan A is more legible than another partial plan B if the number of possible goal locations for A is less than B. 

\textbf{Plan : }In the robot's partial plan, it picks up the package and brings it closer to the ship.

\textbf{Question 1 :} Imagine you are the human observer. Would you find such a partial plan legible? Give your answer as `Yes', `No' or `Can't Say' only.

-------------------

\textbf{Question 2 : }
Why do you think the robot's plan is legible? Give your response only as '1', '2' or '3' based on the following options and do not include any other text in your response:
1) Because the human can not see the robot acting.
2) Because the human does not like the robot's plan.
3) Because the human can not understand the environment.
\end{tcolorbox}

\begin{tcolorbox}[title=Package Delivery Domain: Predictability Prompt,
left skip=0pt, right skip=0pt,
opacityframe=0.4, rounded corners,
width=\linewidth,
]
\label{prompt:package_pred_incon}
\textbf{Description: }Consider a situation where there is a robot which manages a shipping port, and a human observer (you) who is the supervisor that has sensors or subordinates at the port who provide partial information about the nature of activity being carried out at the port. For instance, the human observer only gets to know the identity of the crate when it is either acquired at the port or loaded on the ship.

\textbf{Initial state:} There are packages at the port.

\textbf{Goal:} The robot has to reveal the identity of a package to the human.

\textbf{The human observer can not see the robot acting in the environment.}

\textbf{Definition:} A partial plan A is predictable if the observer (you) can identify if there is one possible completion (which may or may not lead to the goal). A partial plan A is more predictable than a partial plan B if the number of possible completions of A is less than B.

\textbf{Plan :} In the robot's partial plan, it picks up a package and goes towards the port.

\textbf{Question 1 : }Imagine you are the human observer. Would you find such a partial plan predictable? Give your answer as `Yes', `No' or `Can't Say' only.

-------------------

\textbf{Question 2 : }
Why do you think the robot's plan is predictable? Give your response only as '1', '2' or '3' based on the following options and do not include any other text in your response:
1) Because the human can not see the robot acting.
2) Because the human does not like the robot's plan.
3) Because the human can not understand the environment.
\end{tcolorbox}

\begin{tcolorbox}[title=Package Delivery Domain: Obfuscatory Prompt,
left skip=0pt, right skip=0pt,
opacityframe=0.4, rounded corners,
width=\linewidth,
]
\label{prompt:package_obf_incon}
\textbf{Description:} Consider a situation where there is a robot which manages a shipping port, and a human observer (you) who is the supervisor that has sensors or subordinates at the port who provide partial information about the nature of activity being carried out at the port. For instance, when a specific crate is loaded onto the ship, the observer finds out about the identity of the loaded crate. The observer knows the initial inventory at the port, but when new cargo is acquired by the port, the observer’s sensors reveal only that more cargo was received; they do not specify the numbers or identities of the received crates.

\textbf{Initial state:} There are packages at the port.

\textbf{Goal:} The robot can either pick and acquire a package on the port, or else pick and load it on the ship.

\textbf{The human observer can not see the robot acting in the environment.}

\textbf{Definition: } Suppose you think the robot is trying to achieve one out of a set of of potential goals. If the agent's behavior does not reduce the size of this set, then it is obfuscatory. For example, if you think robot is trying to achieve one of {A, B, C}. If it shows a behavior (partial plan) but you think it is still trying to achieve any one of {A, B, C}, then it is obfuscatory. 

Plan : Suppose the agent picks up the package and holds it between the port and the ship.

\textbf{Question 1 :} Imagine you are the human observer. Would you find such a partial plan obfuscatory? Give your answer as `Yes', `No' or `Can't Say' only.

-------------------

\textbf{Question 2 : }
Why do you think the robot's plan is obfuscatory? Give your response only as '1', '2' or '3' based on the following options and do not include any other text in your response:
1) Because the human can not see the robot acting.
2) Because the human does not like the robot's plan.
3) Because the human can not understand the environment.

\end{tcolorbox}

\end{document}